\documentclass[%
preprint,
amsmath,
amssymb,
aps,
physrev,
]{revtex4-2}

\usepackage{graphicx}
\usepackage{dcolumn}
\usepackage{bm}
\usepackage{caption, subcaption}
\usepackage{booktabs}

\begin{document}

\title{\textbf{Predicting Onflow Parameters Using Transfer Learning for Domain and Task Adaptation} 
}%

\author{Emre Y{\i}lmaz}
\email{Contact author: emre.yilmaz@dlr.de}
\affiliation{%
  German Aerospace Center (DLR) - Institute of Aerodynamics and Flow Technology, Center for Computer Applications in AeroSpace Science and Engineering
}

\author{Philipp Bekemeyer}%
\affiliation{%
  German Aerospace Center (DLR) - Institute of Aerodynamics and Flow Technology, Center for Computer Applications in AeroSpace Science and Engineering
}
 
\date{\today}

\maketitle

Determining onflow parameters is crucial from the perspectives of wind tunnel testing and regular flight and wind turbine operations. These parameters have traditionally been predicted via direct measurements which might lead to challenges in case of sensor faults. Alternatively, a data-driven prediction model based on surface pressure data can be used to determine these parameters. It is essential that such predictors achieve close to real-time learning as dictated by practical applications such as monitoring wind tunnel operations or learning the variations in aerodynamic performance of aerospace and wind energy systems. To overcome the challenges caused by changes in the data distribution as well as in adapting to a new prediction task, we propose a transfer learning methodology to predict the onflow parameters, specifically angle of attack, $\alpha$, and onflow speed, $V_{\infty}$. It requires first training a convolutional neural network (ConvNet) model offline for the core prediction task, then freezing the weights of this model except the selected layers preceding the output node, and finally executing transfer learning by retraining these layers. A demonstration of this approach is provided using steady CFD analysis data for an airfoil for i) domain adaptation where transfer learning is performed with data from a target domain having different data distribution than the source domain and ii) task adaptation where the prediction task is changed. Further exploration on the influence of noisy data, performance on an extended domain, and trade studies varying sampling sizes and architectures are provided. Results successfully demonstrate the potential of the approach for adaptation to changing data distribution, domain extension, and task update while the application for noisy data is concluded to be not as effective. 

\clearpage
\section*{Nomenclature}
\subsection*{Abbreviations}
\begin{tabular}{cl}
  CFD & Computation fluid dynamics\\ 
  ConvNet & Convolutional neural network \\ 
  DoE  & Design of experiments \\
  FCNN & Fully connected neural network \\
  INS  & Inertial navigation system \\
  MAE & Mean absolute error \\
  MSE & Mean squared error \\
  NN & Neural network \\
  OL  & Offline learning \\
  RANS  & Reynolds-averaged Navier Stokes \\
  RSM & Reynolds stress model \\
  SA & Spalart-Allmaras one-equation turbulence model \\
  TL & Transfer learning 
\end{tabular}

\newpage 
\subsection*{List of Symbols}
\begin{tabular}{cl}
$\alpha$ & Angle of attack [deg or rad] \\
$\mathcal{D}_e$ & Extended data domain \\
$\mathcal{D}_i$ & Initial data domain \\
$\mathcal{D}_R$ & Data domain created using RANS+RSM \\
$\mathcal{D}_S$ & Data domain generated using RANS+SA \\
$\mathcal{D}_s$ & Source domain \\
$\mathcal{D}_t$ & Target domain \\
$\mathcal{T}_{\alpha}$  & Task of predicting $\alpha$ \\
$\mathcal{T}_{V_{\infty}}$ & Task of predicting $V_{\infty}$ \\
$\mu$ & Mean \\
$\sigma$ & Standard deviation \\
$n_{d}$ & Dataset size \\
$n_{in}$ & Input dimension for FCNN \\
$N_{OL}$ & NN trained during the offline learning phase \\
$n_{s}$ & Number of surface data points \\
$N_{TL}$ & NN trained during the transfer learning phase \\
$P$ & Surface pressure [Pa] \\
$t_{OL}$ & Training time for offline learning \\
$t_{TL}$ & Training time for transfer learning \\
$V_{\infty}$ & Onflow speed [m/s]
\end{tabular}	

\clearpage
\section{Introduction} \label{section_introduction} 
Estimating critical onflow parameters such as angle of attack, $\alpha$, and onflow speed, $V_{\infty}$, with high accuracy is essential for reliably assessing model characteristics in wind tunnel tests and ensuring safe and efficient flight and wind turbine operations. These parameters have traditionally been determined by either direct measurements or statistical inference based on sensor data. The conventional measurement approaches for angles of attack include inertial and optical techniques \cite{finley92, lee92, bogue04, crawford07} as described in \cite{toro18}. Additionally, $\alpha$ is estimated by model-based \cite{valasek20} or data-driven \cite{lerro21} inference schemes using the received state data via instruments such as Inertial Navigation system (INS) \cite{zeis88}. Similarly, flow velocity is typically determined using pressure tubes, anemometry, probes, and PIV as well as optical methods such as the Schlieren technique \cite{gracey58, selig04, bridges11, narbuntas20, smith17}. Furthermore, neural networks have also been employed as a data-driven technique to estimate the components of the flow velocity vector based on surface pressure data \cite{rohloff99, quindlen13, topac23}.

Aircraft manufacturers rely on redundant systems for online monitoring of such critical parameters and leverage consistency tests and voting mechanisms \cite{goupil11}. Examples of the use of redundant systems in modern fly-by-wire aircraft include signal consolidation and monitoring which are executed using multiple $\alpha$ sensors, generally as many as three, before passing the data to the flight control system \cite{ossmann17, vitale21}. As an alternative, the incorporation of data-driven estimation models based on state information into voting systems has been proposed to address the risk of having two faulty sensors in agreement \cite{mersha23}. Inspired by these works, we further investigate the potential of data-driven techniques, specifically neural networks, for predicting onflow parameters based on surface pressure data.

Recent trends in data-driven modeling related to aerodynamic applications are vastly influenced by emerging deep learning techniques which achieved remarkable accomplishments in many fields such as decision making \cite{schrittwieser20}, natural language processing \cite{brown20}, and computational biology \cite{jumper21}. Neural networks with fully connected layers have been employed for predicting aerodynamic coefficients before the breakthroughs in deep learning concepts that occurred over the previous decade \cite{ross97, lo00, rajkumar02}. With the introduction of deep architectures, there has been an expanding use of these techniques promising data-driven models that can extract physical correlations from flow field data. These efforts contain flow field estimation \cite{guo16, yilmaz17, lee17, nagawkar22, morimoto22, hines23, wassing24}, generative design and shape optimization \cite{yan19, chen20, yilmaz20, nagawkar22}, and turbulence modeling \cite{ling16}. Motivated by these advancements, we explore deep convolutional neural network (ConvNet)-based architectures to learn the mapping from sensor pressure readings to onflow condition.

The lifelong variations in system behavior or changes in the environment can also lead to further challenges when determining onflow parameters based on sensor data. The environmental impacts present during operations or tests may alter the distribution of physical data and require the data-driven model to adapt to this new domain via further updates. In this context, transfer learning, a technique effectively used in computer vision \cite{oquab14}, has been applied to force identification problems during wind tunnel testing \cite{sun23}, predicting flow fields around airfoils \cite{krishnan22, wang23, li23}, inverse airfoil design \cite{li23}, performance change due to damage \cite{cappugi21}, aerodynamic load estimation \cite{vaiuso25}, and anomaly detection in wind turbines \cite{roelofs24}. In all studies, knowledge transfer from the domains with abundant data to the domains with sparse data is investigated. With the transfer learning approach, the training optimization can be conducted at reduced computational cost and requires lower training times.

With the objective of overcoming challenges related to domain and task adaptation when predicting onflow parameters based on surface pressure data, a deep learning-based framework is created incorporating transfer learning. The specific onflow parameters to be predicted are selected as i) onflow speed, $V_{\infty}$, and ii) angle of attack, $\alpha$. As a demonstration case, 2-D flow around an airfoil is considered, and the representative datasets are generated using a computational fluid dynamics (CFD) code. The learning architectures evaluated in the framework are neural networks with fully connected or convolutional layers. 

The contributions of the proposed methodology can be summarized as follows:
\begin{enumerate}
  \item A ConvNet based data-driven model to predict onflow parameters, specifically $\alpha$ and $V_{\infty}$, based on surface pressure data
  \item A transfer learning framework that enables updating the network models with new data for domain and task adaptation
  \item Demonstration cases for adaptation to changing data distribution, domain extension, and task change on the same domain as well as robustness to noise for 2-D flow around an airfoil
\end{enumerate}

The remainder of this paper is organized as follows. First, we introduce the transfer learning approach to predicting the onflow parameters providing further theoretical details. Then, we explain the procedures of how to generate the representative aerodynamics data using a CFD solver for an airfoil geometry. Next, the offline learning is demonstrated for the task of predicting $\alpha$ and $V_{\infty}$ with discussions about the selected parameters and the influence of the number of CFD runs, the number of points at which surface data is available, and the architecture depth. Finally, we present the transfer learning results for the scenarios of changing data distribution, domain extension, learning in a noisy domain, and task adaptation.

\section{Methodology \label{section_methodology}} 
\begin{figure}[!t] \centering
\includegraphics[width=.65\columnwidth]{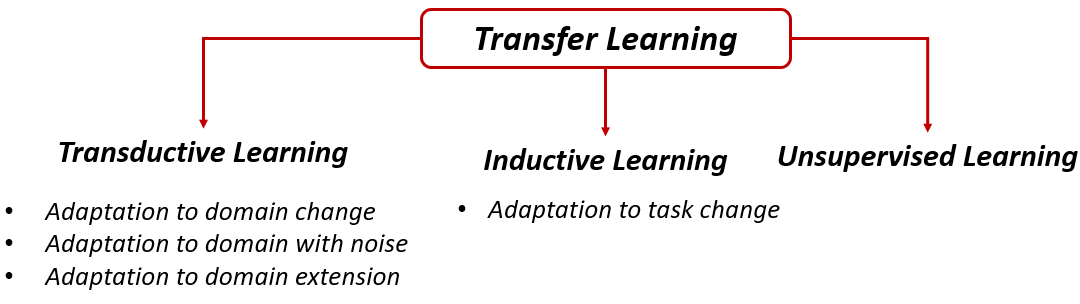}
\caption{Categorization of Transfer Learning}
\label{fig_tl_types}
\end{figure}

\begin{figure} \centering
\begin{subfigure}[b]{0.48\textwidth} \centering
\includegraphics[width=.9\columnwidth]{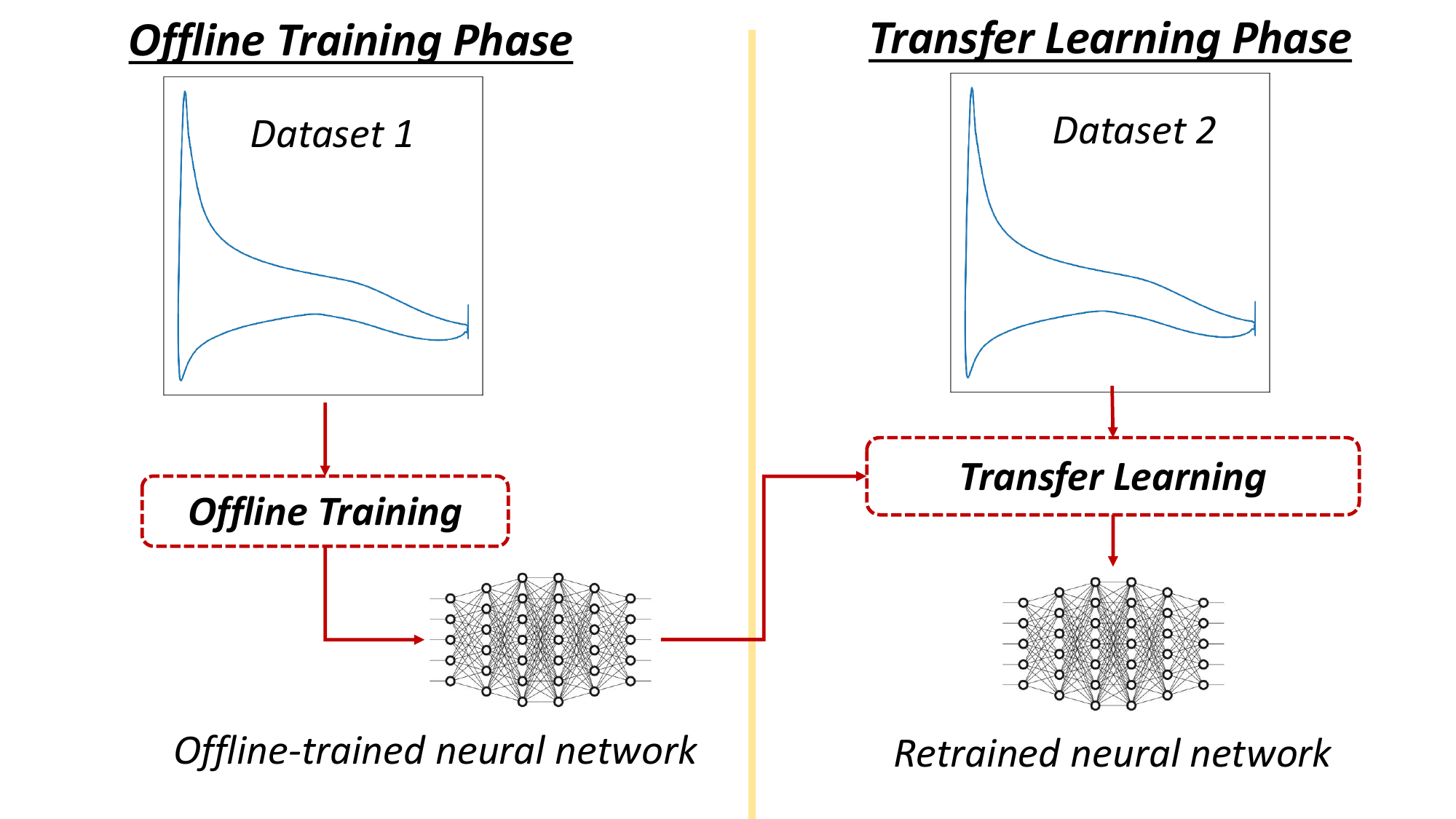}
\caption{Domain Adaptation} \label{fig_domain_adaptation}
\end{subfigure} \hfill
\begin{subfigure}[b]{0.48\textwidth} \centering
\includegraphics[width=.9\columnwidth]{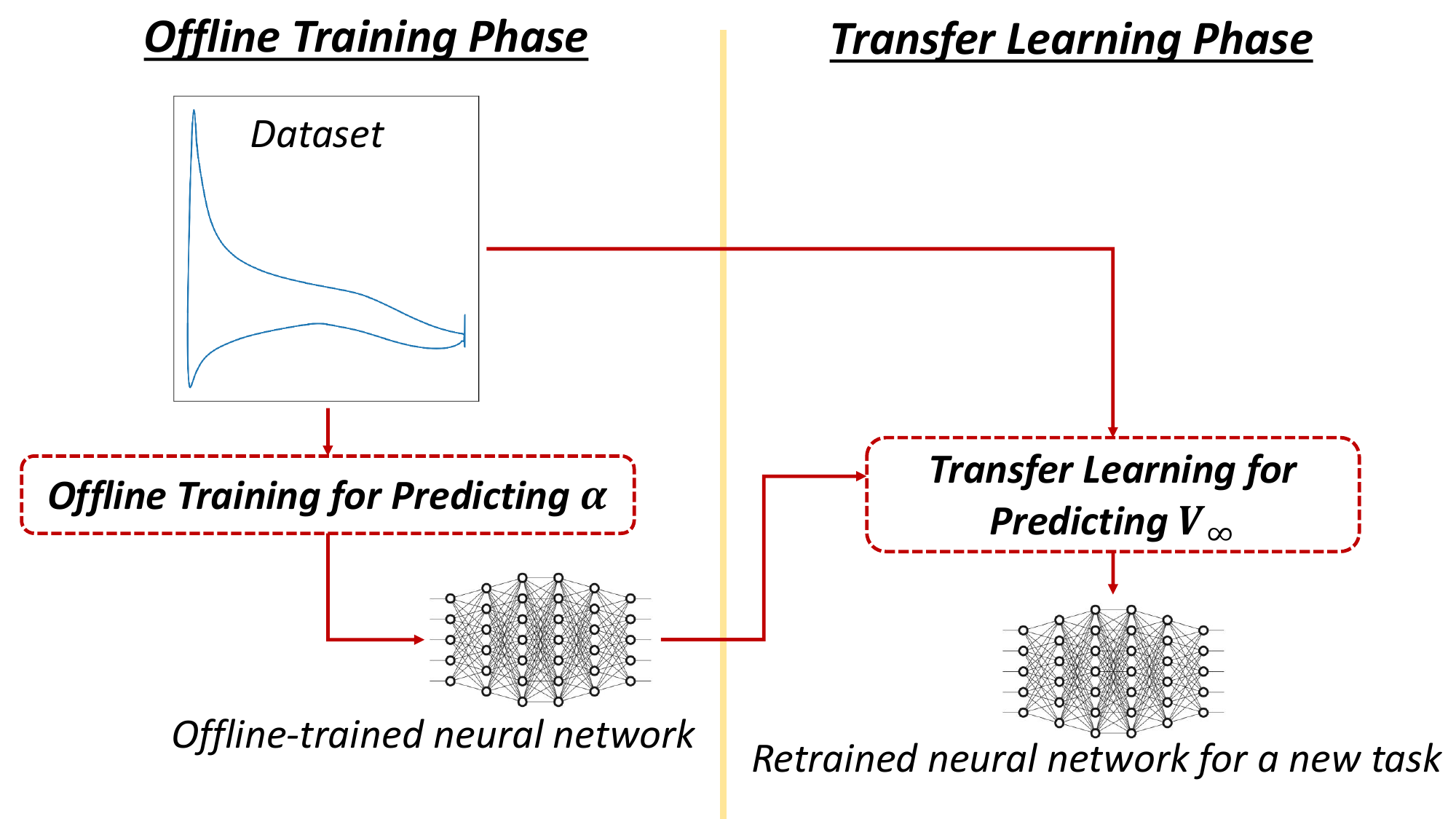}
\caption{Task Adaptation} \label{fig_task_adaptation}
\end{subfigure}
\hfill \vspace{4mm} 
\caption{Transfer Learning Framework}
\label{fig_transfer_learning_framework}  
\end{figure} 

In this section, we introduce how transfer learning can be employed to predict the onflow parameters, $\alpha$ and $V_{\infty}$, based on surface pressure data when either the underlying data distribution of the domain or the type of task changes. This approach can pave the way to achieve online training with streaming data in wind tunnel experiments and during operations and thus enable predictions taking changes in the domain of operation into account. In addition, the knowledge obtained from the offline learning phase can be transferred to perform related tasks such as predicting a different onflow parameter than the one for which the source model was trained offline. Furthermore, we give a brief summary about the deep learning concepts such as convolutional neural networks (ConvNets) that are used to demonstrate the transfer learning framework.

Early literature related to transfer learning includes the discussions about concepts such as knowledge transfer from one task to another, life-long learning, and adaptation based on prior learning \cite{pratt91, naik93, mitchell94}. In a more recent survey \cite{pan10}, tasks are divided into the categories of inductive, transductive, and unsupervised transfer learning. Following the notation and categorization in \cite{pan10}, the types of transfer learning are presented with several examples in Fig. \ref{fig_tl_types} and the formal definitions are provided as follows. Given a domain, $\mathcal{D}$, with a feature space, $\mathcal{X}$, and a marginal probability distribution, $P(X)$, where $X \in \mathcal{X}$, a task, $\tau$, consists of learning a function that can make predictions on an output space, $\mathcal{Y}$. The objective of transfer learning is to improve the learning process for a target task, $\tau_t$ on a target domain, $\mathcal{D}_t$ using the knowledge obtained executing the learning task, $\tau_s$, on the source domain,$\mathcal{D}_s$. In transductive transfer learning, learning experience is transferred from a source domain, $\mathcal{D}_s$, to a target domain, $\mathcal{D}_t$, with different but similar characteristics for the same task. For inductive transfer learning, source task, $\tau_s$, and target task, $\tau_t$, are different but related to each other while source and target domains have the same data distribution. Although both domains and tasks differ from each other in the case of unsupervised learning, they are still sufficiently similar to each other. A significant advancement related to the utilization of transfer learning is the pretraining and task specific tuning of large language models (LLMs) such as BERT \cite{devlin19} and GPT-3 \cite{brown20}.

For the task of predicting onflow parameters, we propose a transfer learning framework for domain and task adaptation. The transfer learning framework used for domain adaptation in this paper is described in Fig. \ref{fig_domain_adaptation}. In this case, for the same prediction task (i.e. either predicting $\alpha$ or $V_{\infty}$), the neural networks, $N_{OL}$, trained offline using data from the source domain can be leveraged for the training optimization of another neural network, $N_{TL}$, on the target domain via transfer learning. Conceptually, the source and target data can be composed of CFD runs or data from wind tunnel experiments or real life operations. Moreover, the transfer learning approach can be used for task adaptation as described by Figure \ref{fig_task_adaptation} when transferring the knowledge learned for the task of predicting $\alpha$ to the task of predicting $V_{\infty}$. In this case, the domains of source and target have the same data distribution whereas the prediction task changes.

As the first step of transfer learning, we train a neural network offline for a given prediction task using available data which corresponds to executing the source task on the source domain. Then, another neural network with the same architecture is initialized using the weights of the pretrained neural network. In a ConvNet based architecture, the weights of the convolutional layers and several fully connected layers following it are generally frozen while the remaining layers preceding the output node are retrained. This results from the assumption that the features extracted on source targets using convolutional operations are generic and can be reused for target tasks \cite{oquab14}. Furthermore, the ability of convolutional layers to generalize to different datasets are investigated, and it is concluded that more discriminative features are learned as the feature hierarchies become deeper \cite{zeiler14}. As a result of the training session with the unfrozen weights, the knowledge extracted on the source domain for the source task can be transferred to execute the target task on the target domain.

At the core of our transfer learning framework, the main prediction task is executed via ConvNet based architectures. Originally inspired by the receptive fields in visual cognition systems, ConvNets are a class of neural networks with at least one layer that contains so-called ``convolutional'' filters to generate feature maps \cite{lecun98}. These filters are slid throughout inputs to apply convolution operations. ConvNets are particularly efficient in detecting local features of neighboring elements by applying a series of convolutional filters to the input images to extract various features from data.

ConvNets have resulted in extraordinary achievements in a vast variety of computer vision applications \cite{mnih15, ronneberger15, redmon16}. At the basis of this idea lies the assumption that local features in an input to the network (e.g. an array representing an input image) remain stationary, and therefore, ConvNets are effective in extracting these local features by executing computations on localized input subregions \cite{lecun98}. Additionally, the concept of weight sharing is exploited with ConvNets where the weights of each filter are shared throughout the spatial locations. Since the same weights are used for the entire input array, extracting local features invariant to translation is possible. In addition to the convolutional operation, a ConvNet-based architecture can include pooling and dropout operations, fully connected linear layers, batch normalization, and activation via function such as Rectified Linear Units (ReLU) \cite{goodfellow16}.

\begin{figure*}[!t] \centering
 \includegraphics[width=1.\columnwidth]{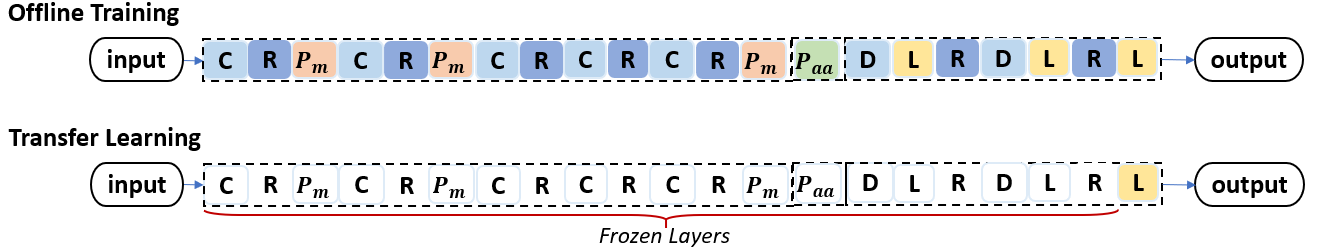}
 \caption{ConvNet-D (C: convolutional, R: ReLU, P: pooling, m: max, aa: adaptive averaging, D: dropout, L: linear)}
 \label{fig_convnet1}
\end{figure*}

\begin{figure*}[!t] \centering
 \includegraphics[width=0.65\columnwidth]{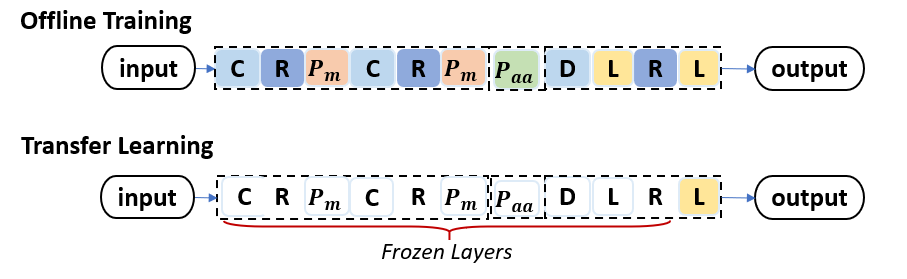}
 \caption{ConvNet-S (C: convolutional, R: ReLU, P: pooling, m: max, aa: adaptive averaging, D: dropout, L: linear)}
 \label{fig_convnet2}
\end{figure*}

The two ConvNet architectures selected for the learning tasks in this paper are shown in Figures \ref{fig_convnet1} and \ref{fig_convnet2}. They can be seen as representatives for a relatively deep and shallow network, respectively. The deeper architecture labeled as ``ConvNet-D'' is selected by converting the AlexNet architecture \cite{krizhevsky12} implemented in the torchvision models \cite{torchvision16} into a one dimensional (1-D) form. This 1-D form allows inputting 1-D arrays to networks and then applying 1-D filters throughout these input arrays. The shallower architecture labeled as ``ConvNet-S'' is created by removing several of these layers. Figures \ref{fig_convnet1} and \ref{fig_convnet2} describe how the weights of the layers of these network architectures are frozen during the transfer learning phase when training only the last layer. For comparison purposes, a fully connected neural network (FCNN) labeled as ``FCNN'' is also formed based on the linear layers of the shallow architecture.

\section{Data Generation \label{section_data_generation}} \vspace{2mm}
This section presents how the CFD analyses are executed to generate the representative data for aerodynamic performance of an airfoil. The generated data is assumed to represent the quasi-steady data obtained during i) an ordinary measurement task in a wind tunnel experiments, where tests are conducted based on a set of prefixed $\alpha$ and $V_{\infty}$ values, and ii) steady flight or wind turbine operations within the operational envelope where the different combinations of $V_{\infty}$ and $\alpha$ are held constant. 

\begin{figure}[!t] \centering
\includegraphics[width=.45\columnwidth]{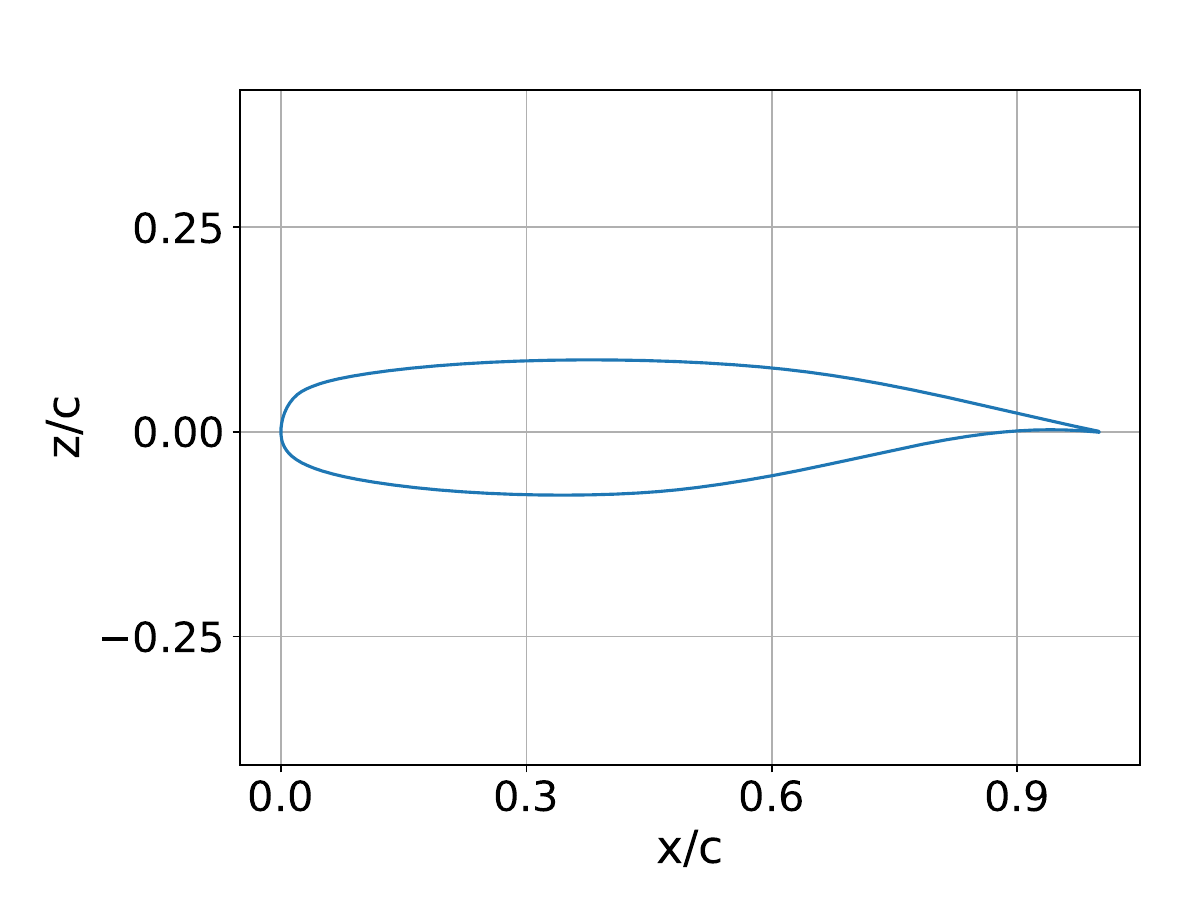}
\caption{NLR 7301}
\label{fig_airfoil}
\end{figure}

The steady CFD analyses are performed for the supercritical NLR 7301 airfoil \cite{zwaan82} shown in Fig \ref{fig_airfoil} using the DLR-TAU Code \cite{schwamborn06} as flow solver. For all analyses, the Reynolds-averaged Navier-Stokes (RANS) equations are solved in conjunction with two different turbulence models, namely the Spalart Allmaras one-equation turbulence model (SA) \cite{allmaras12} and the Reynolds stress model (RSM), SSG/LRR-ln$\omega$, \cite{eisfeld16}, in order to generate datasets with relevant but different distributions.

\begin{align}  \begin{split} 
\hspace{35mm} V_{\infty} &\in [40, 70] \hspace{2mm} \text{m/s} \hspace{20mm} \alpha \in [-15, 17] \hspace{2mm} \text{deg} 
\end{split} \label{eqn_doe_boundaries}
\end{align}

\begin{figure}[!t]
	\centering
	\includegraphics[width=.45\columnwidth]{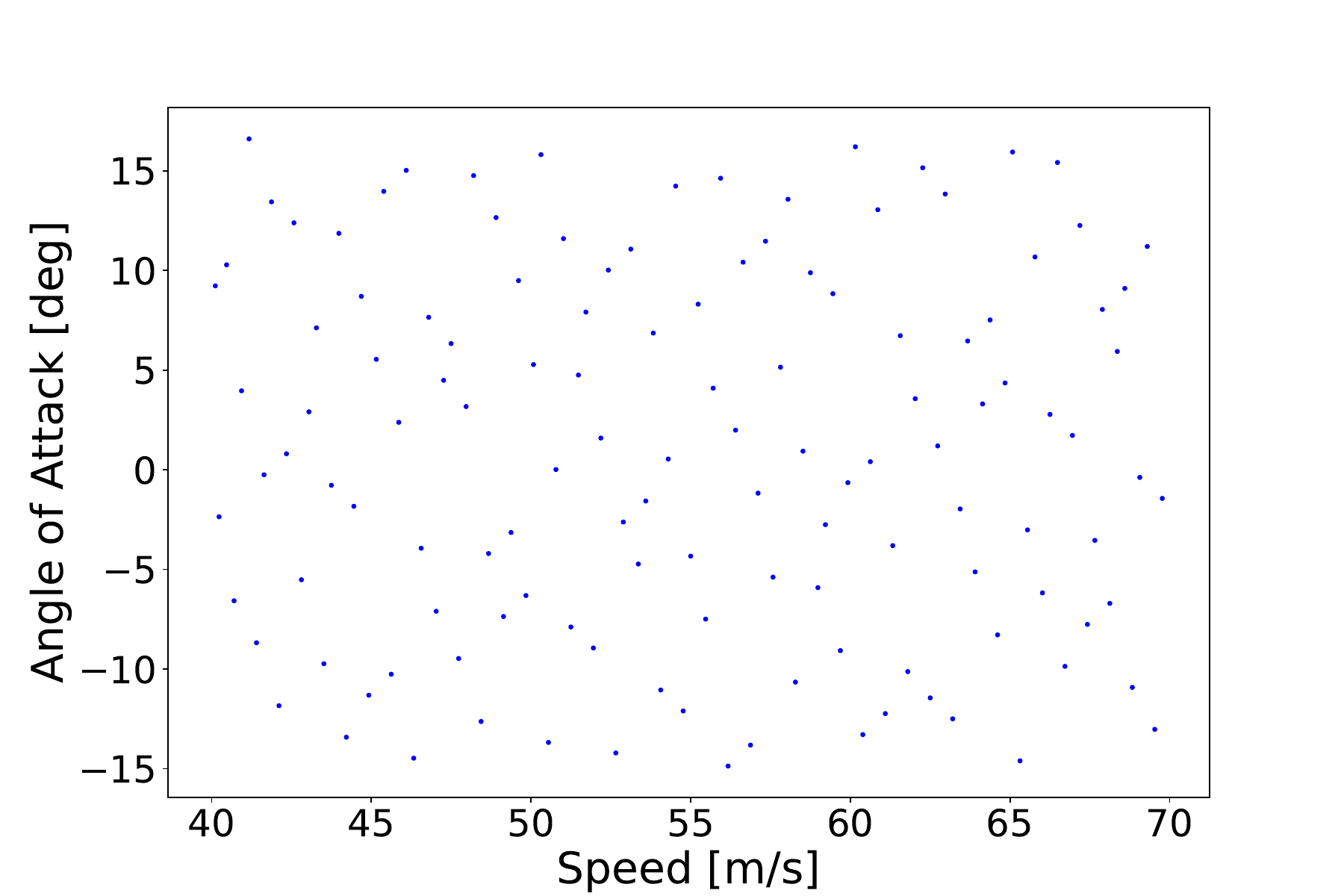}
	\caption{Design of experiments for 128 CFD runs}
	\label{fig_doe_cfd}
\end{figure}

To create the design of experiments (DoE) cases, $\alpha$ and $V_{\infty}$ are chosen as factors within the boundaries shown in Eqn.~\ref{eqn_doe_boundaries}. The set of DoE cases are created using Halton's method available in the DLR Surrogate Modeling for AeRo data Toolbox Python Package (SMARTy)~\cite{bekemeyer22}. The main advantages of the Halton's method are the low discrepancy for lower dimensions and the flexibility to add further points to the generated datasets without changing the sampling characteristics due to its deterministic nature~\cite{halton1964}. A subset of the generated DoE cases are shown in Fig.~\ref{fig_doe_cfd} by including the first 128 points for visualization purposes.

\section{Results and Discussion} \label{section_results_and_discussion}
In this section, we present the offline and transfer learning results for the problems of predicting $\alpha$ and $V_{\infty}$ based on sparse surface pressure data. First, we provide an explanation of the steps for i) the parameter selection for the training optimization and demonstration cases, ii) how to preprocess the CFD data before the optimization phase, iii) how to split the data sets into training, validation, and test sets, and iv) the assessment of the performance of the trained models using evaluation metrics. Next, we discuss the prediction accuracy observed using the ConvNet-based architectures during the offline learning phase and provide several comparison cases in terms of the size of the entire dataset including training, validation, and test sets and the number of sampled surface measurement points. In this context, the dataset size corresponds to the number of samples selected from the entire CFD data set for the learning phases. Then, we discuss the transfer learning results regarding the demonstration cases for domain and task adaptation. The demonstration cases presented for domain adaptation include i) adaptation to a domain with relevant but different data distribution, ii) adaptation to a domain extension, and iii) adaptation to a noisy data domain. The feasibility of applying the transfer learning approach for task adaptation is also discussed on a demonstration problem where the source and target tasks are chosen as predicting $\alpha$ and $V_{\infty}$, respectively. For all demonstration cases, the variation of the results with the architecture type and dataset size are included. Finally, an investigation of the required training time for the shallow and dense networks as well as the dataset sizes during the offline and transfer learning phases are provided.

\subsection{Data preprocessing and splitting, parameter selection, and evaluation metrics \label{subsec_data_preprocess}}
In this subsection, we explain the procedures for i) preprocessing and splitting the data and ii) selecting the parameters for the training optimization and the metrics that are used to evaluate the trained models. The pressure data on the airfoil surface obtained from the CFD analyses is cast into 1-D arrays and used as inputs to the neural network architectures. In this setting, convolution operations are executed with 1-D filters along these 1-D arrays. To investigate the variation of the results with the dimension of the input data, the surface data values shown in Table~\ref{table_param} are considered, and the input arrays are extracted from the CFD data accordingly. For each of these cases, surface pressure data points are selected equidistantly. Since the learning problem is essentially a regression problem, the output is a real-valued scalar corresponding to the onflow parameter of interest. Before the training optimization, inputs and outputs are normalized via min-max normalization such that their values remain between 0 and 1. In addition, the sensitivity of the training optimization to the size of the datasets used for both offline and transfer learning is investigated in terms of the test set prediction accuracy in the succeeding subsections. Therefore, the number of CFD runs expressed in Table~\ref{table_param} are used to create the datasets of different sizes. These datasets are then split into training, validation, and test sets and used for the training optimization of neural networks and the assessment of prediction accuracy.

\begin{table}[!t] 
\centering  \caption{Parameter sets for investigating prediction performance}
\begin{tabular}{cc}
 Parameter & Set  \\
 \cmidrule(l r ){1-1} \cmidrule(l r ){2-2} 
 Surface data points & \{38, 75, 150, 299, 597\}
 \vspace{2mm} \\
 Number of CFD runs & \{128, 256, 512, 1024\} \\
\end{tabular}  
\label{table_param}
\end{table}

As mentioned in Section~\ref{section_data_generation}, CFD runs are executed based on a set of DoE cases created by employing Halton sequences. Since this approach ensures low discrepancy as we add more points sequentially, it is agnostic to the number of samples in the set. Therefore, the generated DoE set is divided into training, validation, and test sets successively retaining the order of its elements. The splitting percentages of training, validation, and test sets are selected as 72\%, 18\%, and 10\%, respectively. 

Regarding the demonstration cases for adaptation to a domain extension, the datasets are first sorted in ascending order with respect to their output values and split from the median value into two subsets with equal numbers of elements. Then, these two subsets are sorted again based on the indices of their elements in the entire dataset before the first sorting operation. The resorting operation is executed to restore the Halton's sequence due to the reasons mentioned above. Next, the aforementioned procedure for creating training, validation, and test sets are followed for both subsets. Finally, offline training is performed on one half of the dataset, and transfer learning is employed by merging the training, validation, and test sets of both halves. For the remaining domain adaptation cases, the procedure for offline training is repeated with the new dataset.

We also investigate the influence of retraining only the last layer or the last two layers, which are fully-connected and linear, during the transfer learning phase. In case of one layer, we retrain 0.086\% of the entire architecture for the shallow network and 0.017\% for the dense one. Alternatively, if we choose to retrain two layers for both architectures, then the percentage of unfrozen weights are 98.697\% for ConvNet-S and 70.359\% for ConvNet-D. For these ConvNet architectures introduced in Section~\ref{section_methodology}, the sequences and sizes of layers and convolution and pooling parameters are included in Tables~\ref{table_convnetD} and ~\ref{table_convnetS} in the Appendices, respectively. For comparison purposes, the results obtained employing only a fully connected neural network architecture labeled as FCNN are also provided for the offline training step. The parameters of the fully connected neural network are given in Table~\ref{table_fcnn}. 

The tasks of predicting $\alpha$ and $V_{\infty}$ are formulated as regression problems. The mean squared error (MSE) is selected as the loss function for the training optimization since it has a differentiable form, which is critical for gradient-based optimization and backpropagation. The contributions from outliers to MSE can result in misinterpretation of prediction errors. Alternatively, mean absolute error (MAE) is robust to outliers, and, the metric for assessing the accuracy of predictions is, therefore, defined as mean absolute error (MAE). The relationships for these metrics are expressed in Eqn. \ref{eqn_metrics} where $N_{\text{test}}$ is the size of the test set, $y_i$ is the output of the $i^{\text{th}}$ sample from the test set, which is used as the ground truth, and $\hat{y}_i$ is the predicted value corresponding to the $i^{\text{th}}$ sample.

\begin{align} \begin{split}
 \text{MAE} &= \frac{1}{N_{\text{test}}} \sum_{i=1}^{N_{\text{test}}} |y_i - \hat{y}_i|  \\
 \text{MSE} &= \frac{1}{N_{\text{test}}} \sum_{i=1}^{N_{\text{test}}} (y_i - \hat{y}_i)^2 \label{eqn_metrics}
\end{split} \end{align}

\subsection{Predicting onflow parameters via ConvNet architectures}
\begin{figure}[!t] \centering
\begin{subfigure}[b]{0.48\textwidth}
\includegraphics[width=.99\columnwidth]{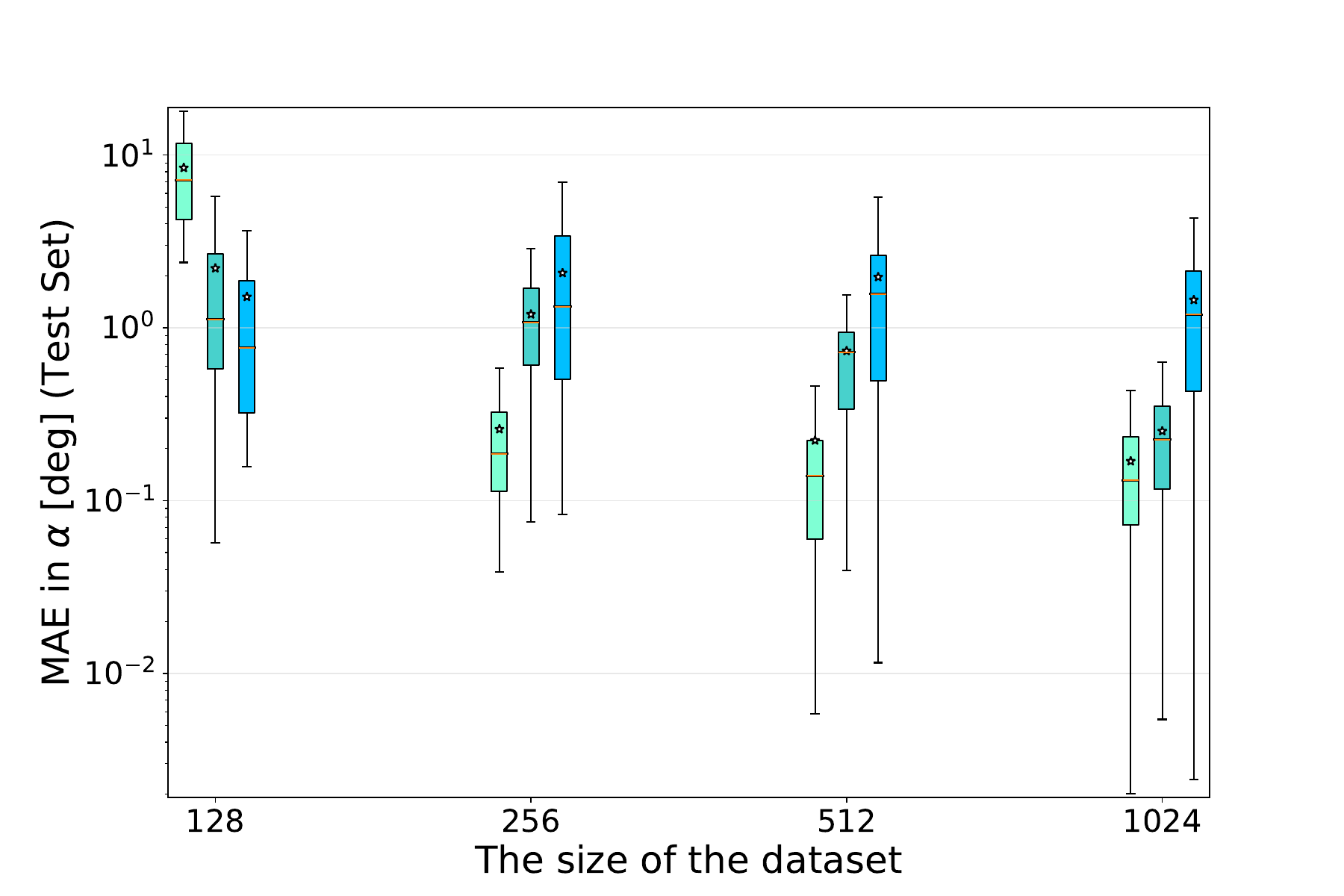}
\caption{Task: Predicting $\alpha$} \label{fig_mae_aoa_nn_ol}
\end{subfigure} \hfill
\begin{subfigure}[b]{0.48\textwidth}
\includegraphics[width=.99\columnwidth]{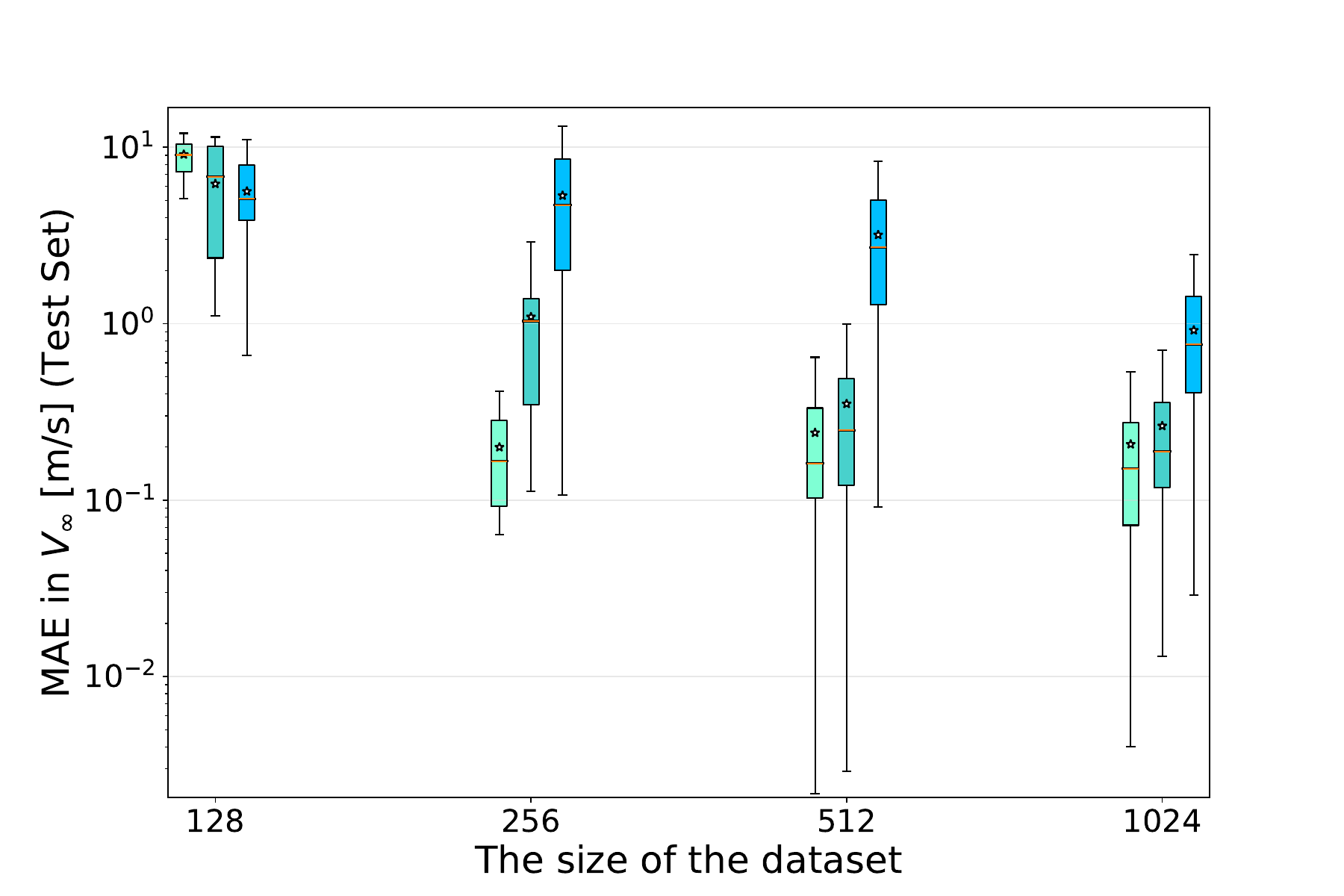} 	
\caption{Task: Predicting $V_{\infty}$} \label{fig_mae_vinf_nn_ol}
\end{subfigure} \hfill \vspace{4mm} \break 
\begin{subfigure}[b]{0.48\textwidth}
  \includegraphics[width=.99\columnwidth]{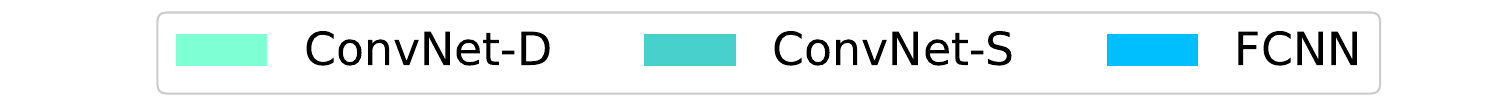}
  \label{fig_mae_nn_ol_leg}
\end{subfigure}  
\caption{The variation of the test set MAE values with the architecture type and dataset size for 75 surface data points} \label{fig_mae_nn_ol}  
\end{figure} 

\begin{figure}[!t] \centering
\begin{subfigure}[b]{0.48\textwidth}
  \includegraphics[width=.99\columnwidth]{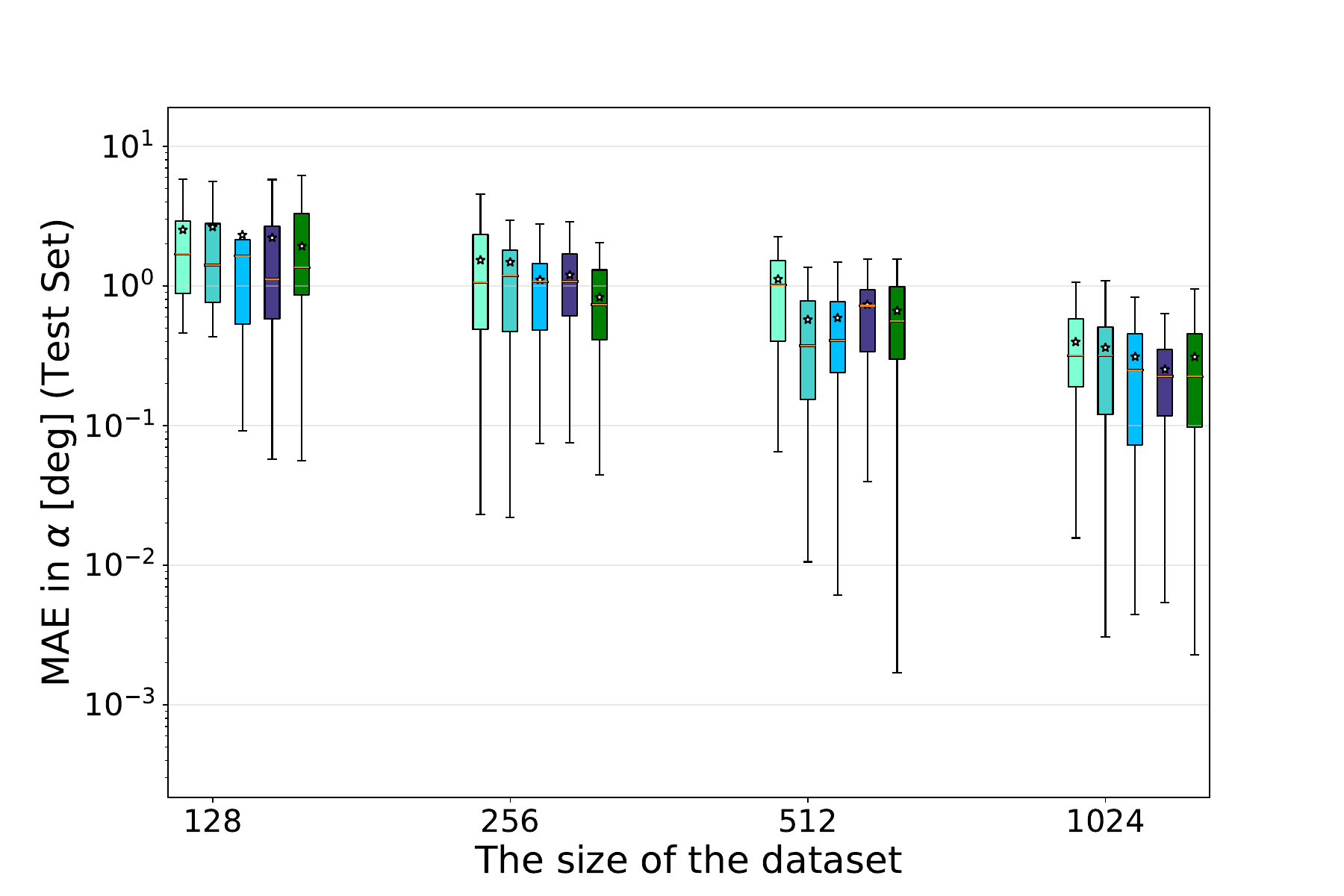} 	
  \caption{ConvNet-S} \label{fig_mae_a_ol_convS}
\end{subfigure} \hfill
\begin{subfigure}[b]{0.48\textwidth}
  \includegraphics[width=.99\columnwidth]{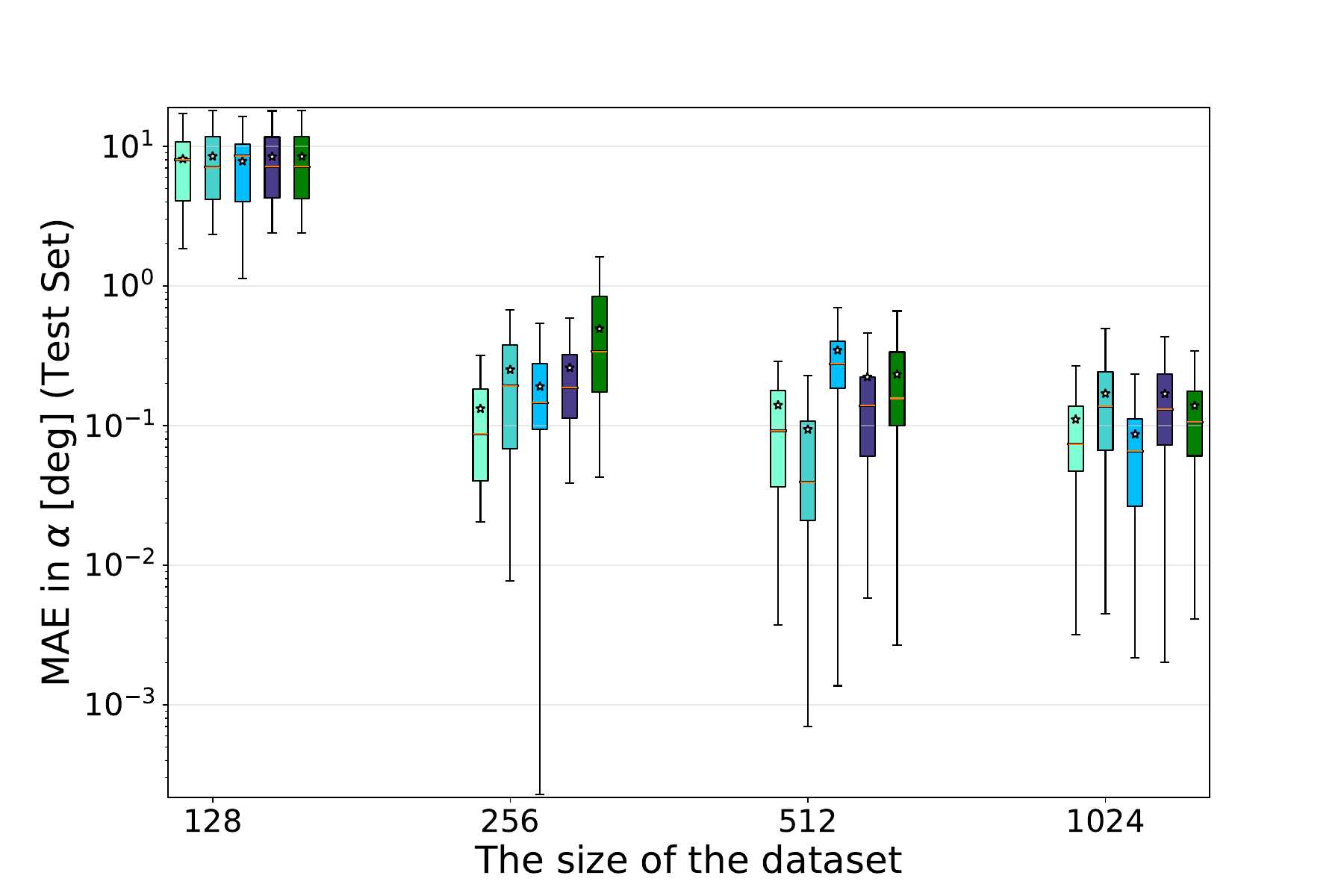} 	
  \caption{ConvNet-D} \label{fig_mae_a_ol_convD}
\end{subfigure} \hfill \vspace{4mm} \break 
\begin{subfigure}[b]{0.48\textwidth}
  \includegraphics[width=.99\columnwidth]{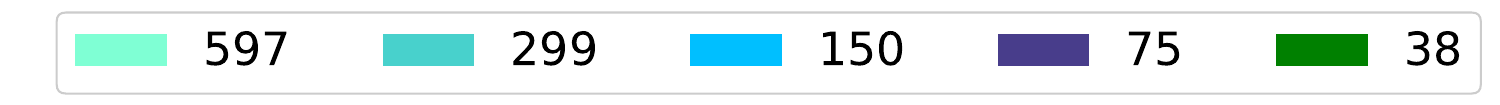}
  \caption*{$n_s$ (the number of surface data points)}
  \label{fig_mae_a_ol_leg}
\end{subfigure}  
\caption{The variation of the test set MAE values with the number of surface data points and dataset size (Task: Predicting $\alpha$)} \label{fig_mae_aoa_ol}
\end{figure} 

\begin{figure}[!t] \centering
\begin{subfigure}[b]{0.48\textwidth}
\includegraphics[width=.99\columnwidth]{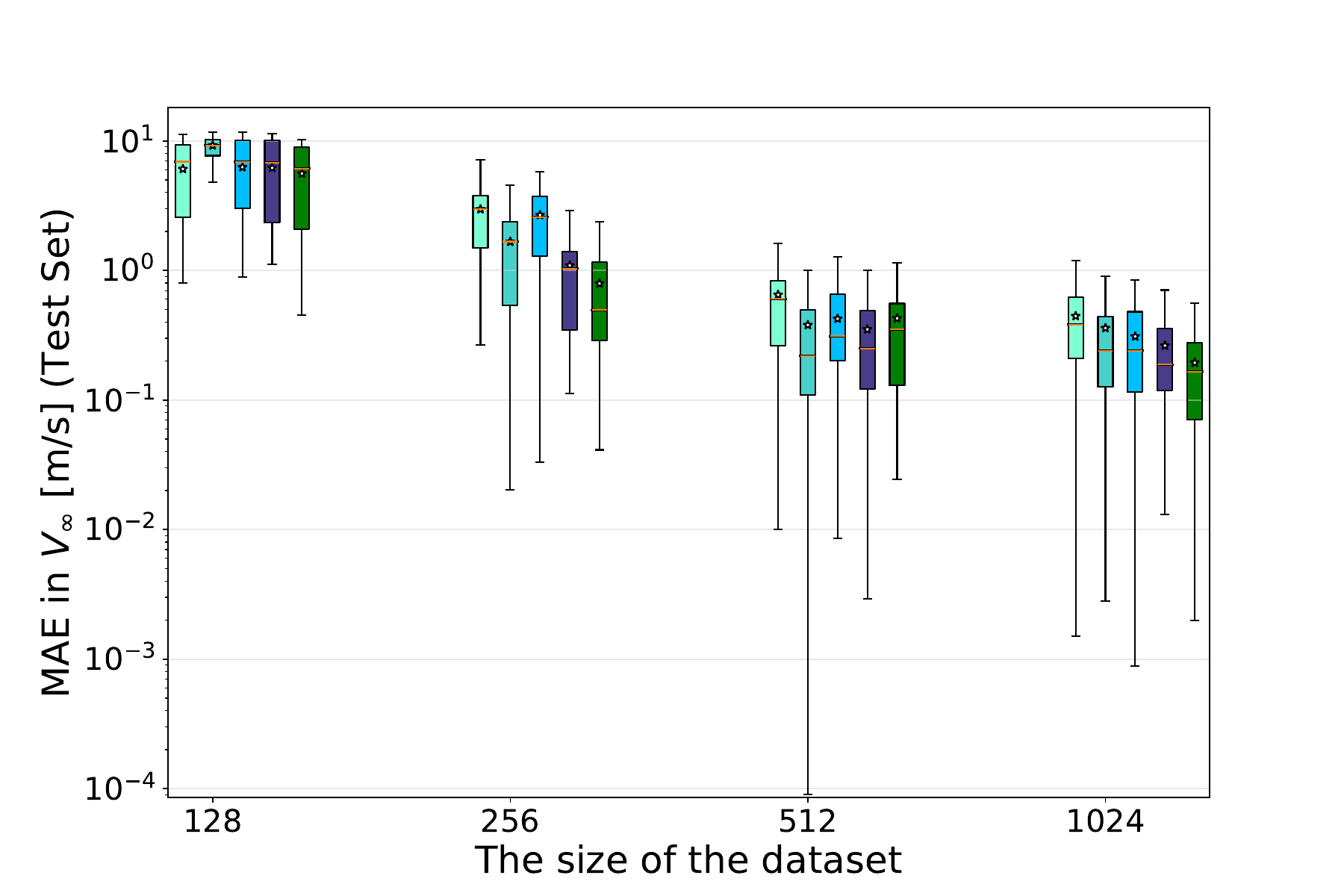} 	
\caption{ConvNet-S} \label{fig_mae_vinf_ol_convS}
\end{subfigure} \hfill
\begin{subfigure}[b]{0.48\textwidth}
\includegraphics[width=.99\columnwidth]{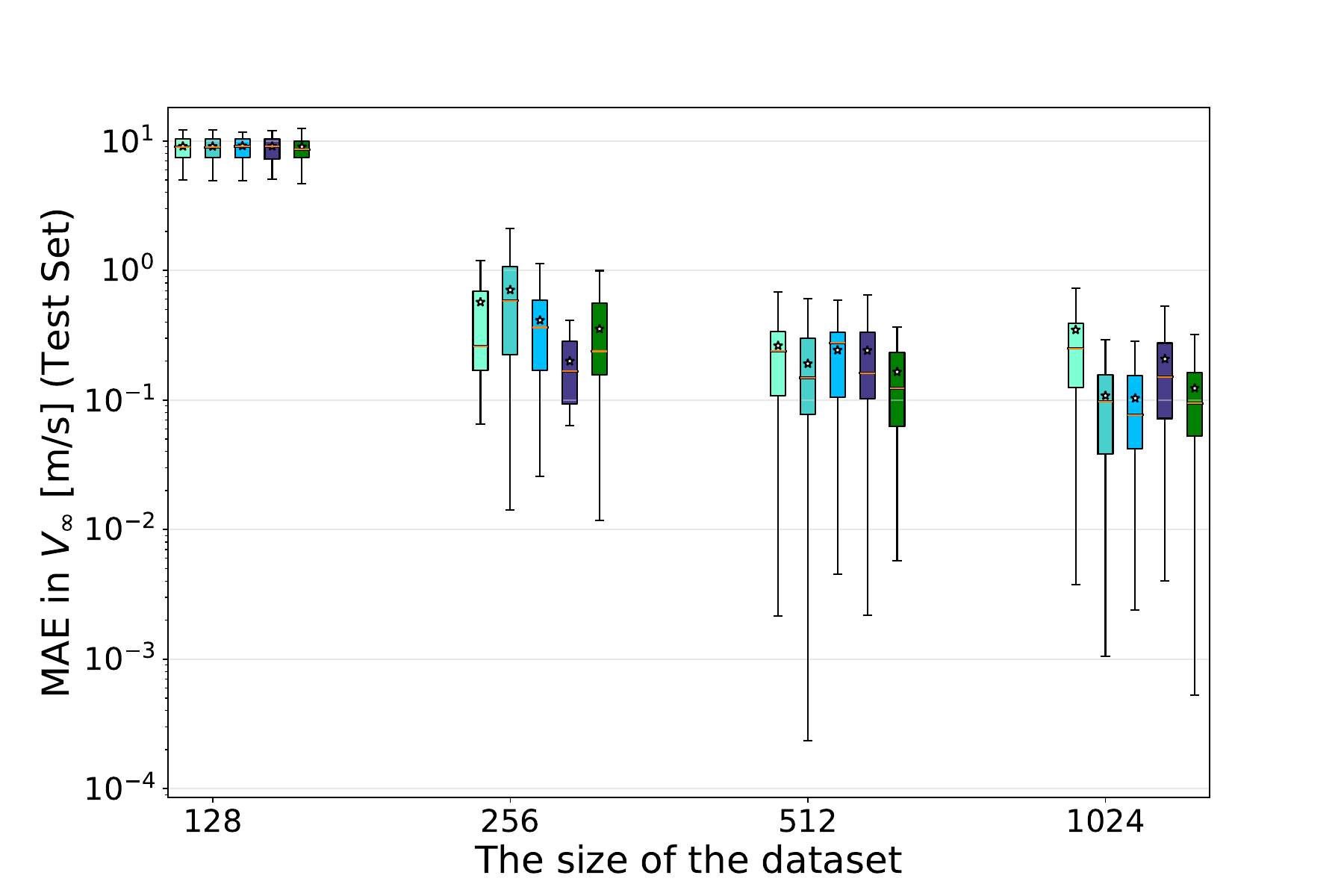} 	
\caption{ConvNet-D} \label{fig_mae_vinf_ol_convD}
\end{subfigure}
\hfill \vspace{4mm} \break 
\begin{subfigure}[b]{0.48\textwidth}
\includegraphics[width=.99\columnwidth]{figs/results1/offline/off_test_error_legend_s.pdf}
\caption*{$n_s$ (the number of surface data points)} \label{fig_mae_vinf_ol_leg}
\end{subfigure}  
\caption{The variation of the test set MAE values with the numbers of surface data points and dataset size (Task: Predicting $V_{\infty}$)}
\label{fig_mae_vinf_ol}  
\end{figure} 

\begin{figure}[!t] \centering
\begin{subfigure}[b]{0.48\textwidth}
\includegraphics[width=.99\columnwidth]{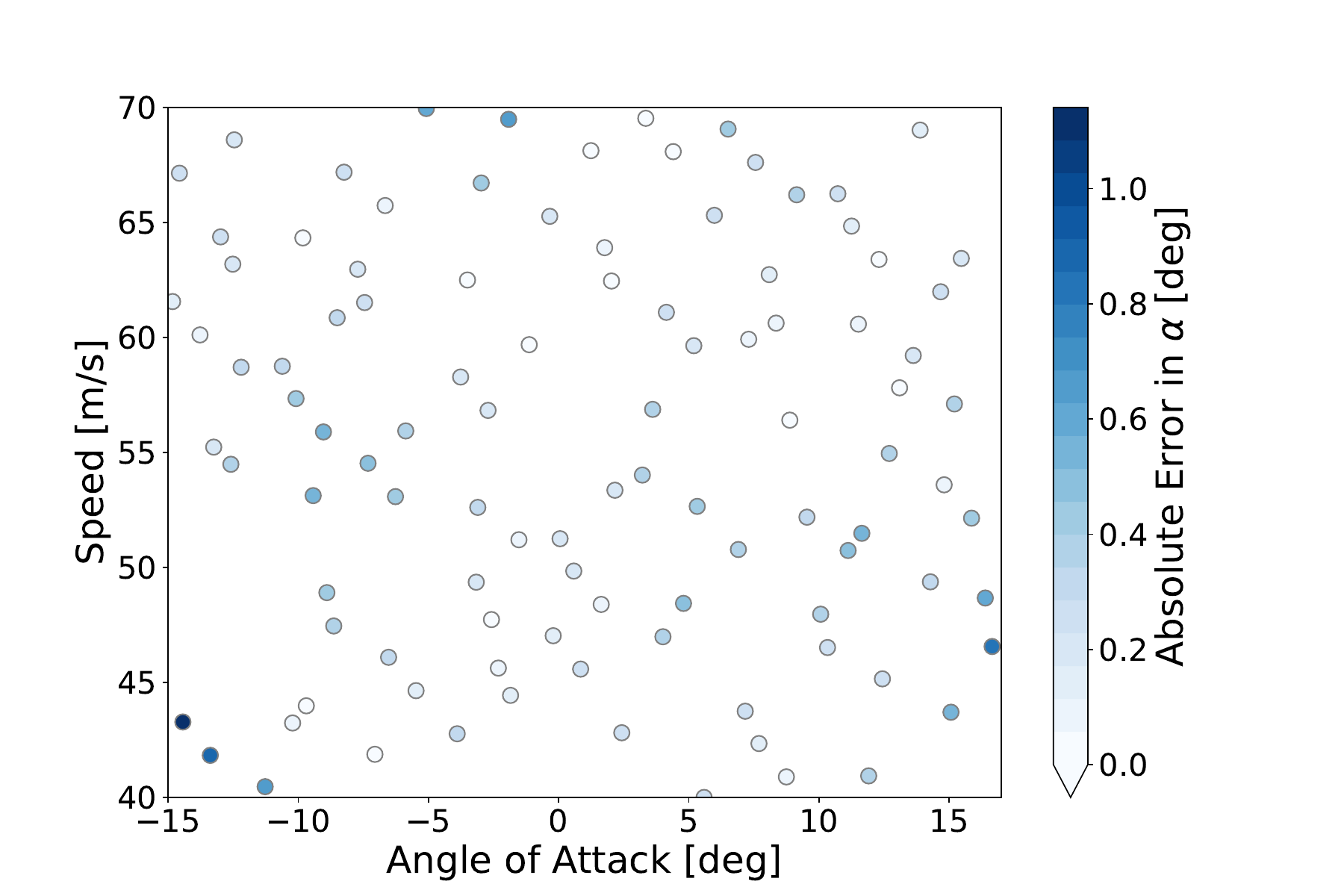}
\caption{Conv-S} \label{fig_mae_aoa_tl1_convS_oono}
\end{subfigure} \hfill
\begin{subfigure}[b]{0.48\textwidth}
\includegraphics[width=.99\columnwidth]{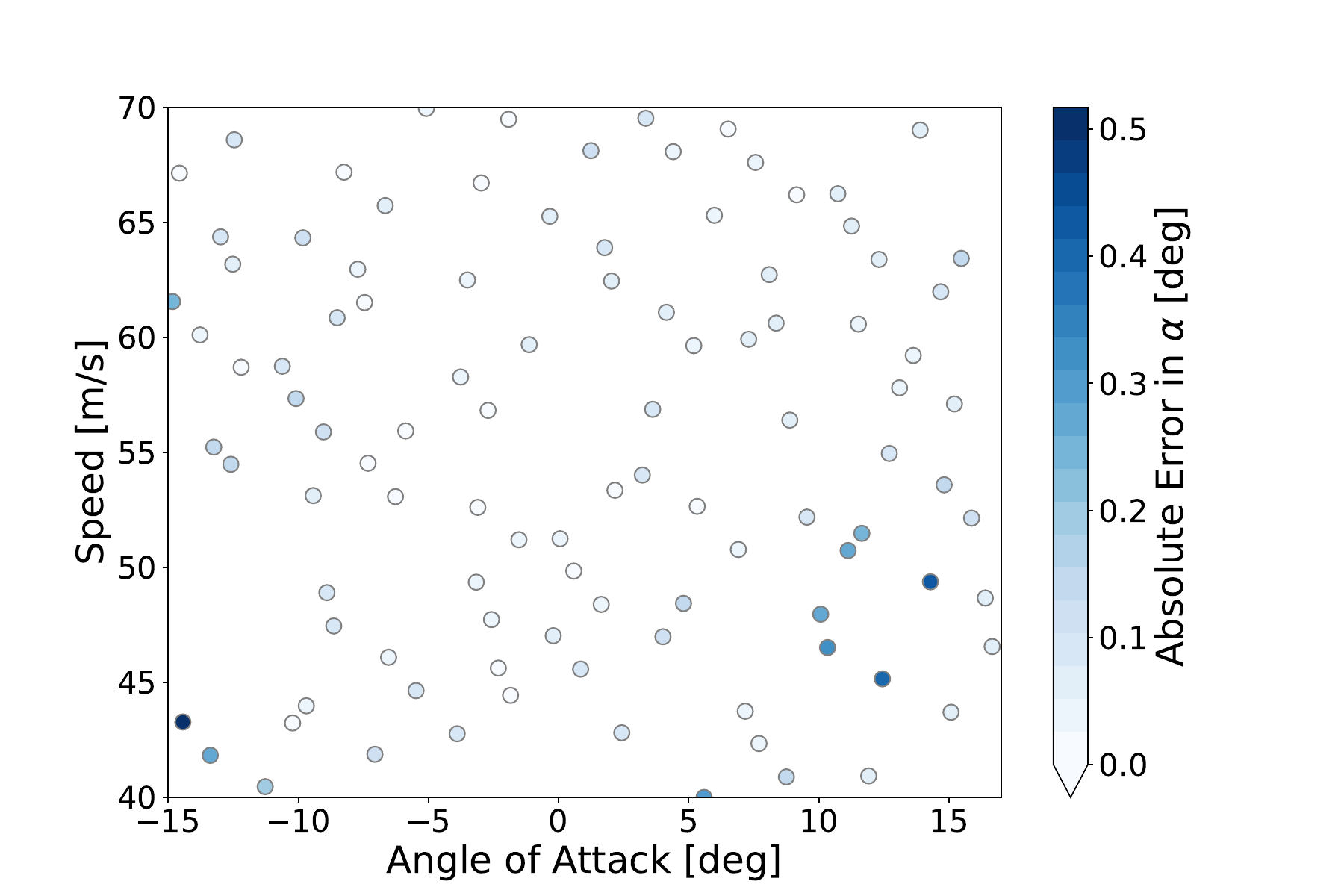} 
\caption{Conv-D} \label{fig_mae_aoa_tl1_convS_tont}
\end{subfigure}
\caption{Test set absolute errors for offline learning and the task of predicting $\alpha$ ($n_s=75$, $n_d=1024$)}
\label{fig_ae_aoa_ol}  
\end{figure}

\begin{figure}[!t] \centering
\begin{subfigure}[b]{0.48\textwidth}
\includegraphics[width=.99\columnwidth]{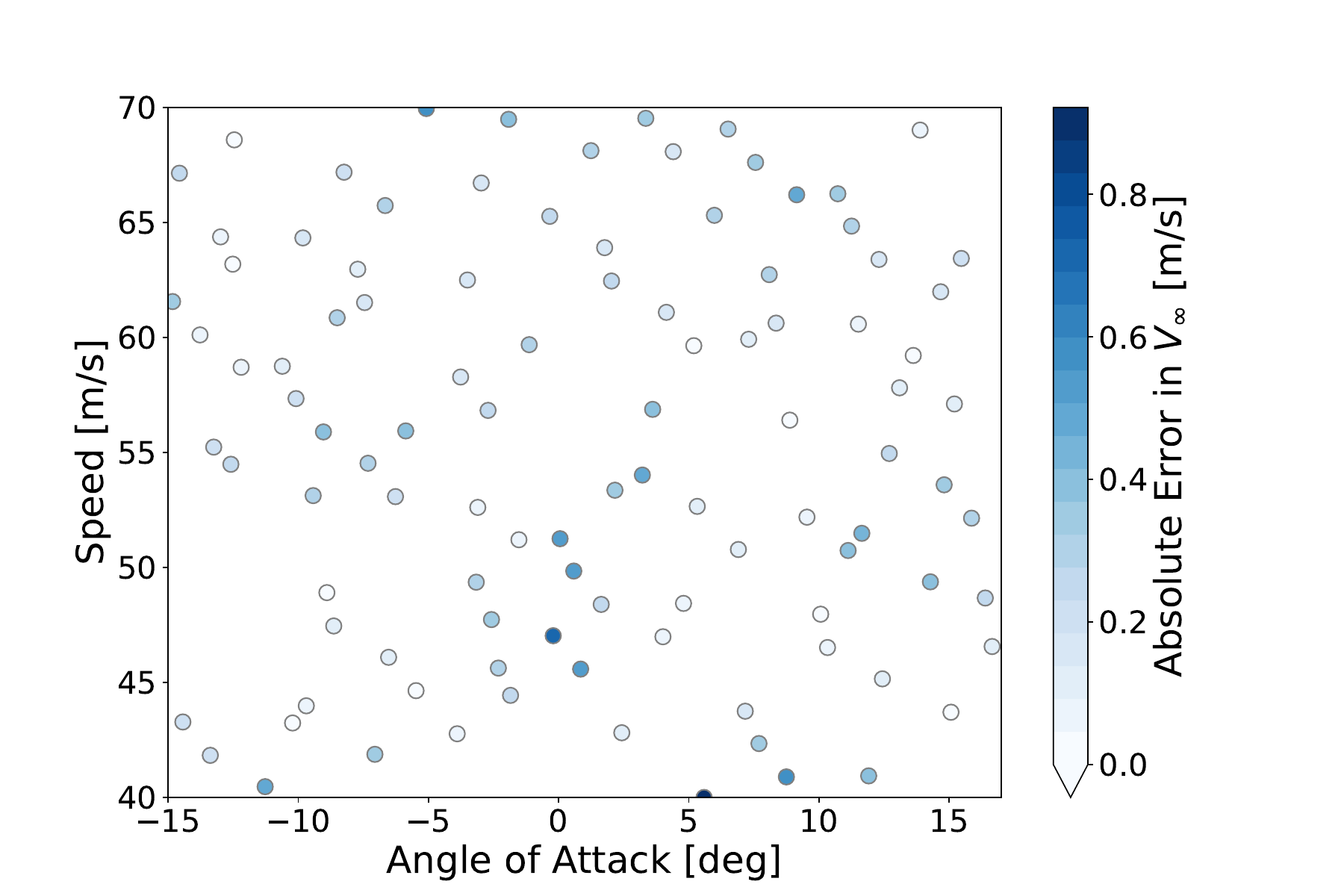}
\caption{Conv-S} \label{fig_mae_vinf_tl1_convS_oono}
\end{subfigure} \hfill
\begin{subfigure}[b]{0.48\textwidth}
\includegraphics[width=.99\columnwidth]{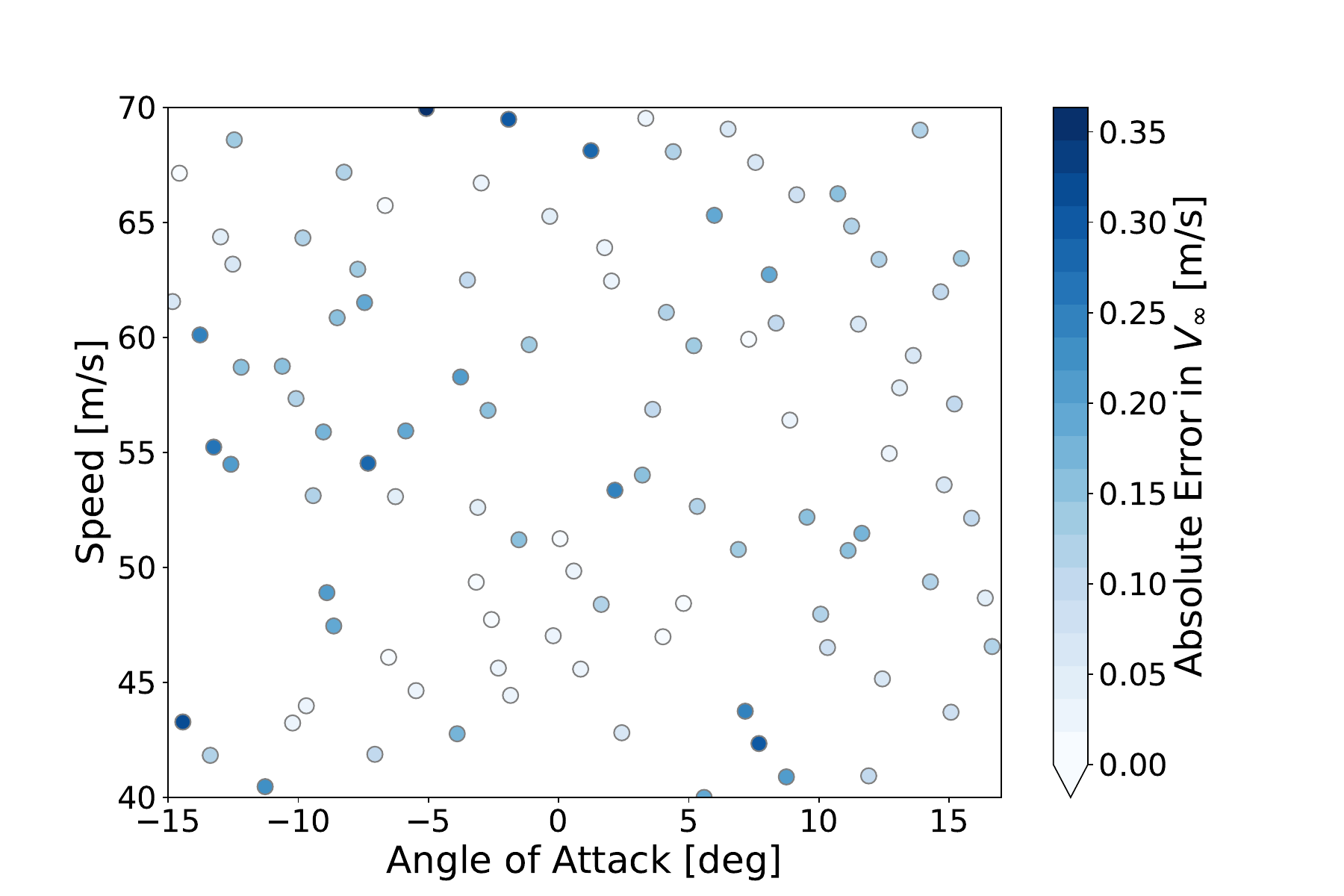}
\caption{Conv-D} \label{fig_mae_vinf_tl1_convS_tont}
\end{subfigure} 
\caption{Test set absolute errors for offline learning and the task of predicting $V_{\infty}$ ($n_s=75$, $n_d=1024$)} \label{fig_ae_vinf_ol}
\end{figure} 

As the first demonstration case, the results obtained during the offline training phase are presented with discussions about how the prediction accuracy varies with several factors. The prediction accuracy is measured by the test set MAEs as well as the variance, median, and quartiles of the test set absolute error values. Statistical values are derived by first computing the absolute errors for each entry in the test set, and then by analyzing the distribution of these errors. For the offline training phase, we investigate the influence of the architecture type, the numbers of points sampled on the airfoil surface, and the dataset size used for learning on the prediction accuracy. As explained above, the dataset size is defined as the sum of the sizes of the training, validation, and test sets for each case. 

We first consider the variations of the test set MAE values with architecture type and dataset size, which are shown in Figures~\ref{fig_mae_aoa_nn_ol} and \ref{fig_mae_vinf_nn_ol} for the tasks of predicting $\alpha$ and $V_{\infty}$, respectively. For these cases, we fix the number of sampled surface data points to 75. It should be noted that the error values in the figures throughout the manuscript are shown in log scale. In comparison with the fully connected architecture, the ConvNet architectures achieve lower MAE values for both prediction tasks except for the minimum dataset size. Considering the practical applications, the MAE values for the minimum dataset size are found to be unacceptably high for all architectures. As the dataset size increases, the prediction accuracy achieved by the fully connected neural network is much worse than the ConvNet based architectures, and the corresponding MAE values cannot be reduced below $10^0$ for both prediction tasks. Using the deep architecture, test set MAE values can be reduced to the level of $10^{-1}$ with a dataset size of 256 and do not improve significantly for the dataset sizes of 512 and 1024. For the shallow architecture, we observe more of a gradual trend where the MAE values decrease with the increase in dataset sizes and can approach the prediction accuracy of the dense architecture when dataset size is set to 1024. Based on these observations, the results for the remaining cases are only provided using the ConvNet based architectures.

Next, we investigate the variation of the test set MAEs with the number of surface data points, where the pressure data is extracted, as well as dataset sizes. The results obtained for the problems of predicting $\alpha$ and $V_{\infty}$ are shown in Figures~\ref{fig_mae_aoa_ol} and \ref{fig_mae_vinf_ol}, respectively. The specific architecture is indicated under each subfigure. In this subsection, each bar plot for a fixed data set size indicates the number of surface data points as described by the corresponding legends. The surface pressure data is first extracted from the CFD data based on the meshed locations on the airfoil surface. Preserving the neighboring locations around the airfoil surface, an input array that contains the full surface information is created. Then, for each downsampling case, a new input array is created by skipping a predetermined number of elements as points are selected equidistantly.

The observations from Figures~\ref{fig_mae_aoa_ol} and \ref{fig_mae_vinf_ol} are summarized as follows. In agreement with the previous observations, the MAE results attained for the smallest dataset size indicate that the number of samples are not adequate to learn the mapping ensuring the level of desired accuracy. As the dataset size increases for both prediction cases, the MAE values have a tendency to decrease as expected. This decrease is much greater for the dense architecture whereas it follows a more gradual trend for the shallow architecture. When the variation of the MAE values with the number of sampled surface data points is inspected, we observe that there is no general trend for different dataset sizes, tasks, and networks. For example, the MAE values decrease as more surface points are sampled for a dataset size of 256 when using the ConvNet-D architecture to predict $\alpha$. In order to extract the angle of attack information from the surface pressure data, local changes can be regarded as important factors, and the trend for this case, where local information is provided to the network at a finer resolution, seems to support this claim. Contrary to this, when employing a shallow network architecture for both tasks with a dataset size of 1024, the MAE values increase with increasing surface data points. In this contradictory case and the remaining cases with no clear trend, the amount of information that is already provided or the chosen pooling operations might be playing a role.

We provide the absolute error values of individual samples in Figures~\ref{fig_ae_aoa_ol} and \ref{fig_ae_vinf_ol} for these parameter values in order to gain a more in-depth insight. For $\alpha$, the greatest absolute error values are obtained close to the lower limits of speed and angle of attack values or at higher angle of attack values below 55 m/s where separated flow is present. For speed, we observe a more random distribution of the absolute error values. For both prediction tasks, the maximum absolute error value obtained using the ConvNet-S architecture is greater compared to ConvNet-D. Based on the discussed results, a dataset size of 1024 is concluded as sufficient to learn the mappings with prediction accuracy approximately resulting in an MAE value around $10^{-1}$ independent of the architecture. Therefore, we set surface data points, $n_s$, to 75 and the dataset size, $n_d$, to 1024 for the demonstration cases in the remaining parts of the manuscript.

\subsection{Adaptation to the changes in data distribution}
\begin{figure*}[!t] \centering
\begin{subfigure}[b]{0.48\textwidth}
\includegraphics[width=.99\columnwidth]{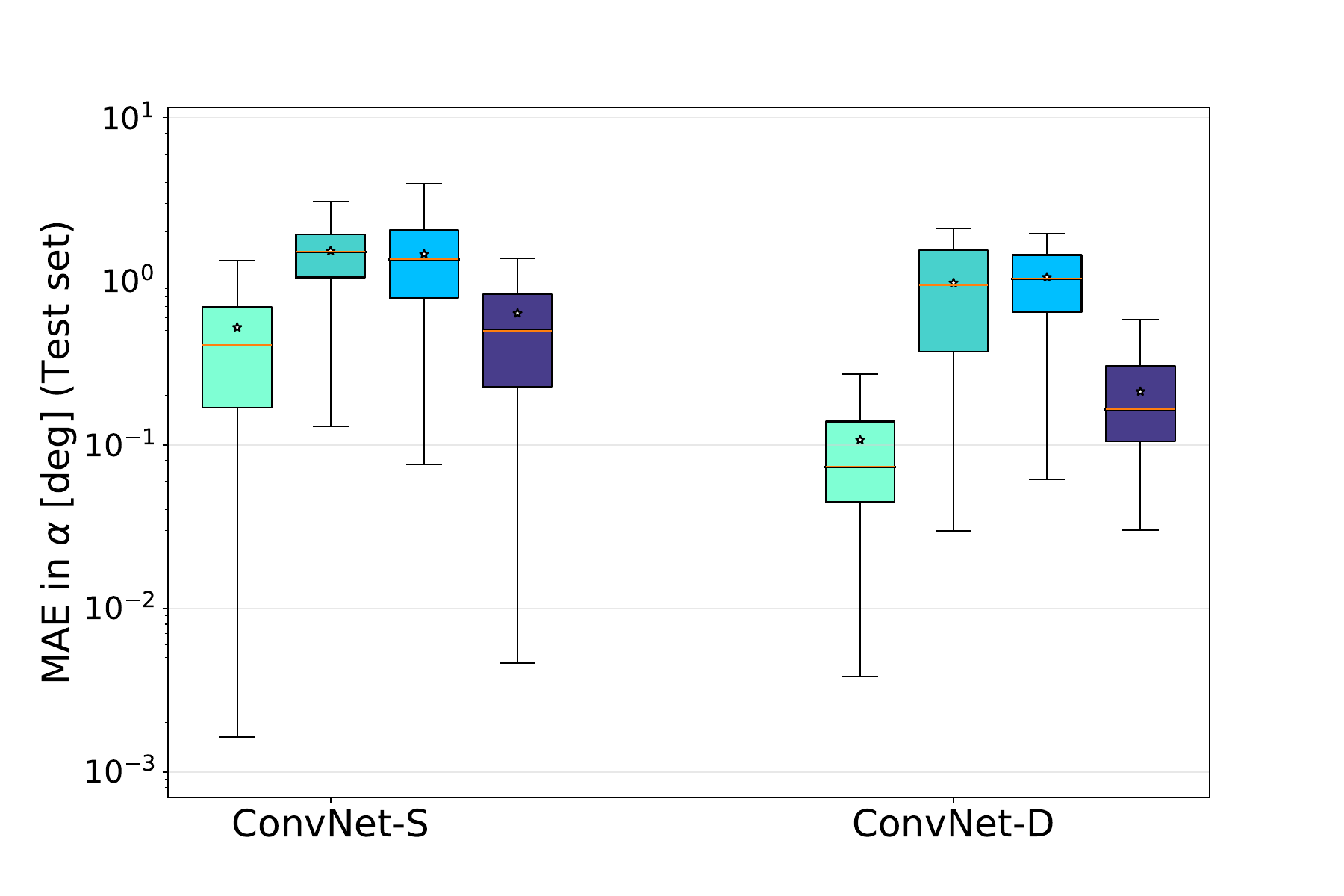} 	
\caption{Task: Predicting $\alpha$} \label{fig_mae_aoa_tl2}
\end{subfigure} \hfill
\begin{subfigure}[b]{0.48\textwidth}
\includegraphics[width=.99\columnwidth]{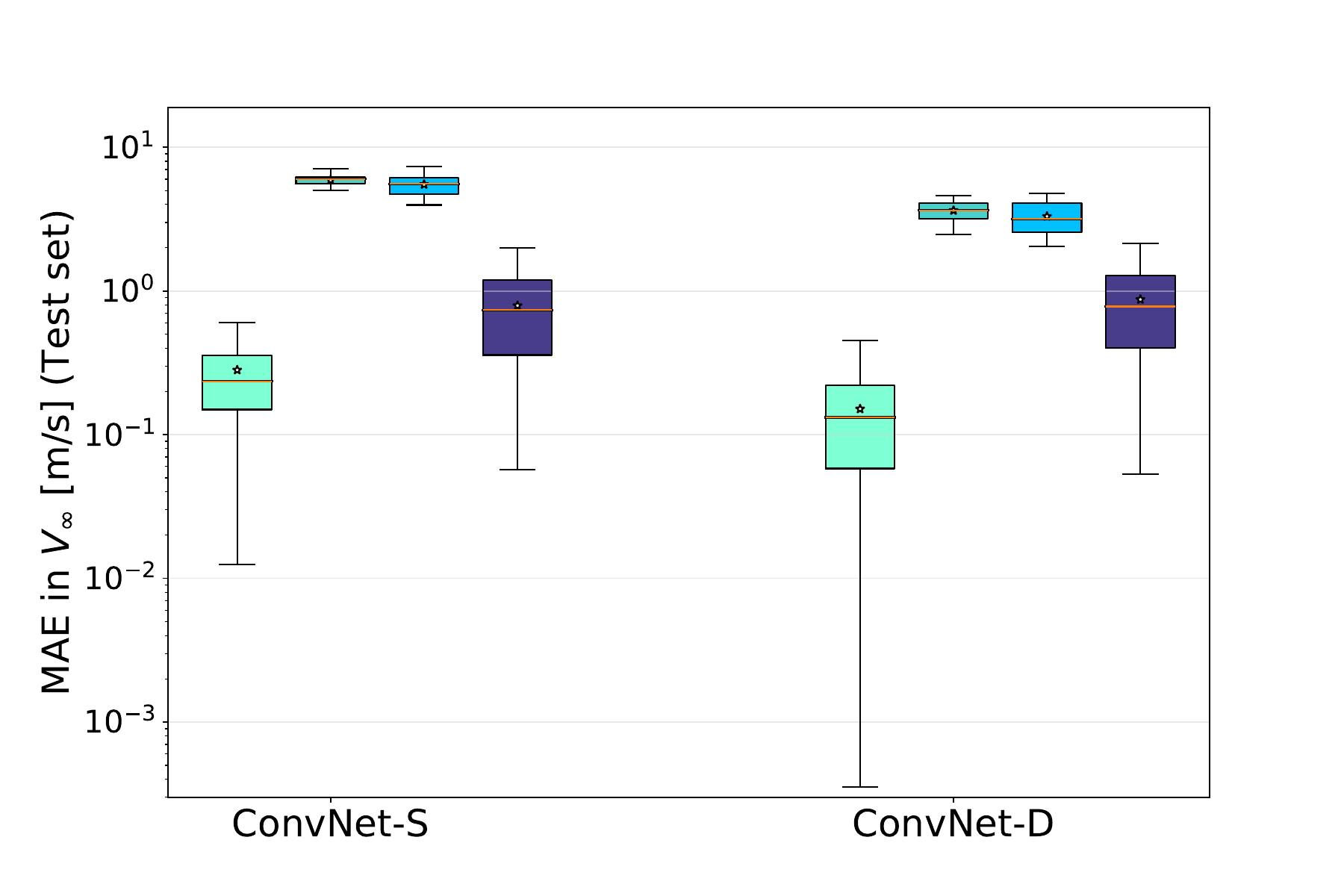} 	
\caption{Task: Predicting $V_{\infty}$} \label{fig_mae_vinf_tl2}
\end{subfigure} \hfill \vspace{4mm} \break 
\begin{subfigure}[b]{0.7\textwidth}
\includegraphics[width=.99\columnwidth]{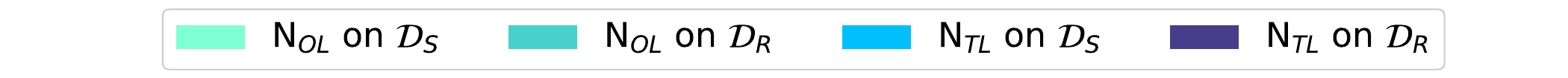}
\end{subfigure}  
\caption{Test set MAE values for transfer learning between the datasets generated using different turbulence models ($n_s=75$, $n_d=1024$)} \label{fig_mae_tl2}
\end{figure*} 

\begin{figure}[!t] \centering
\begin{subfigure}[b]{0.48\textwidth}
\includegraphics[width=.99\columnwidth]{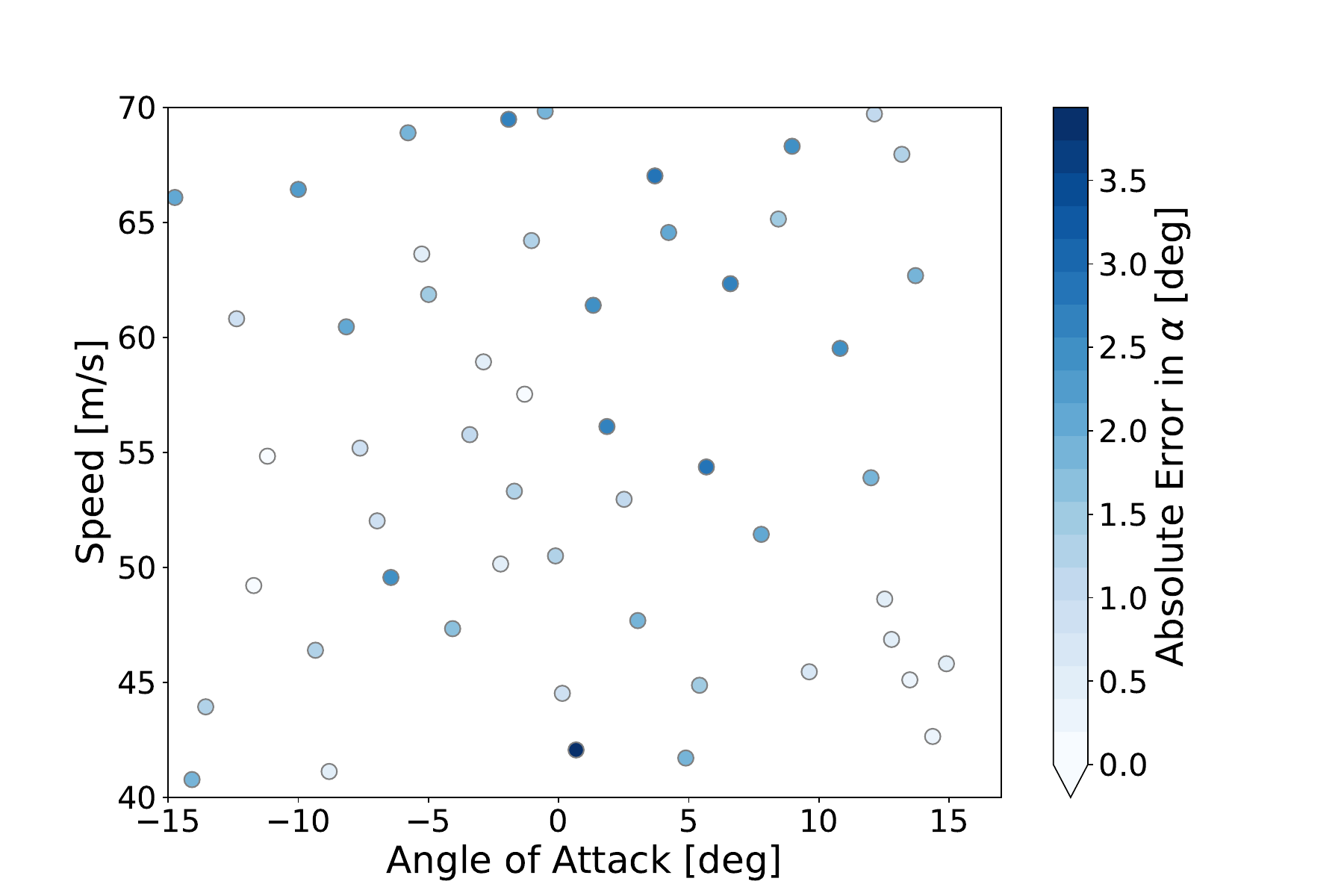} 	
\caption{$N_{TL}$ on $\mathcal{D}_S$ (Conv-S)} \label{fig_ae_aoa_tl1_convS_to}
\end{subfigure} \hfill
\begin{subfigure}[b]{0.48\textwidth}
\includegraphics[width=.99\columnwidth]{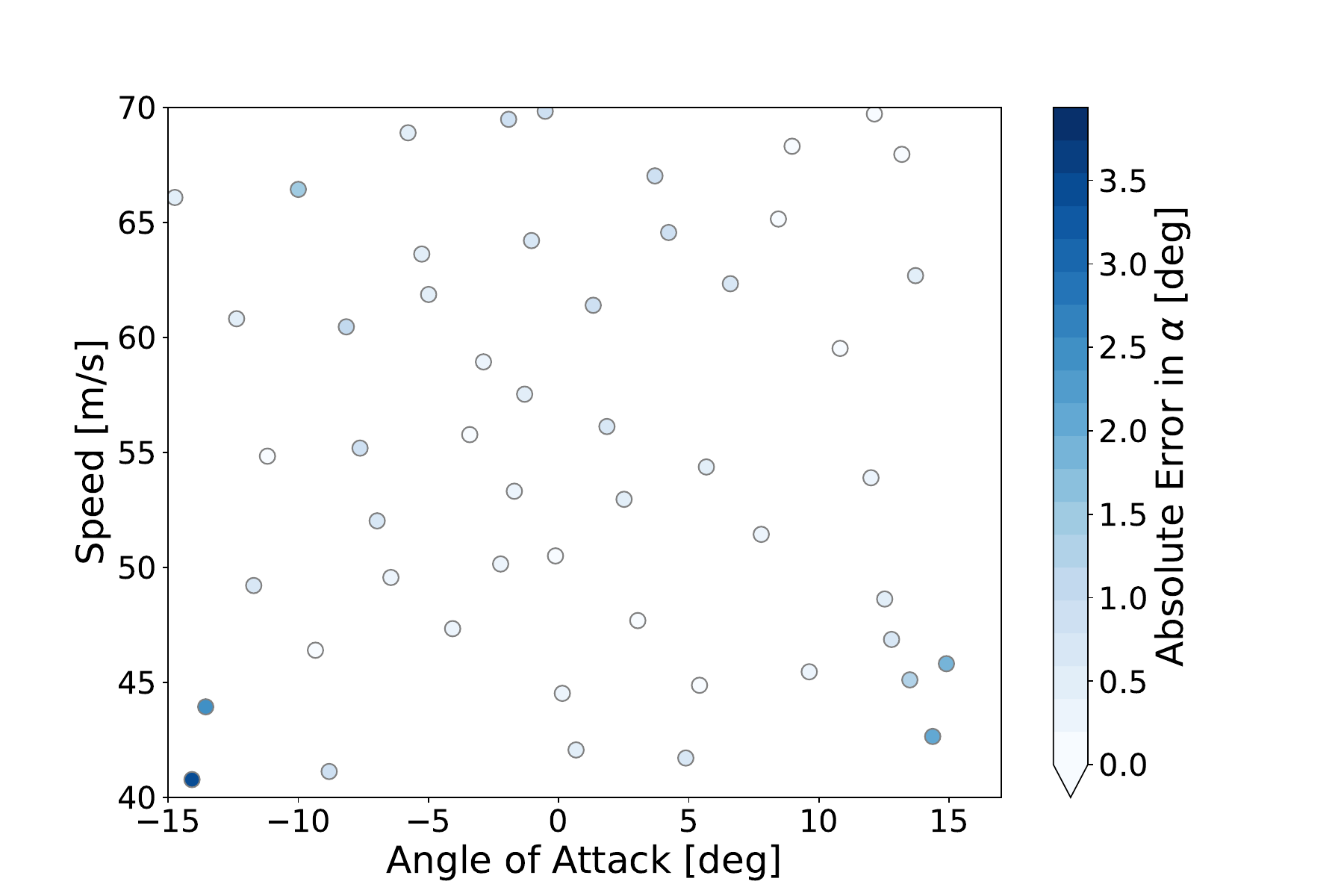} 	
\caption{$N_{TL}$ on $\mathcal{D}_R$ (Conv-S)} \label{fig_ae_aoa_tl2_convS_tt}
\end{subfigure} \hfill \vspace{4mm} \break 
\begin{subfigure}[b]{0.48\textwidth}
\includegraphics[width=.99\columnwidth]{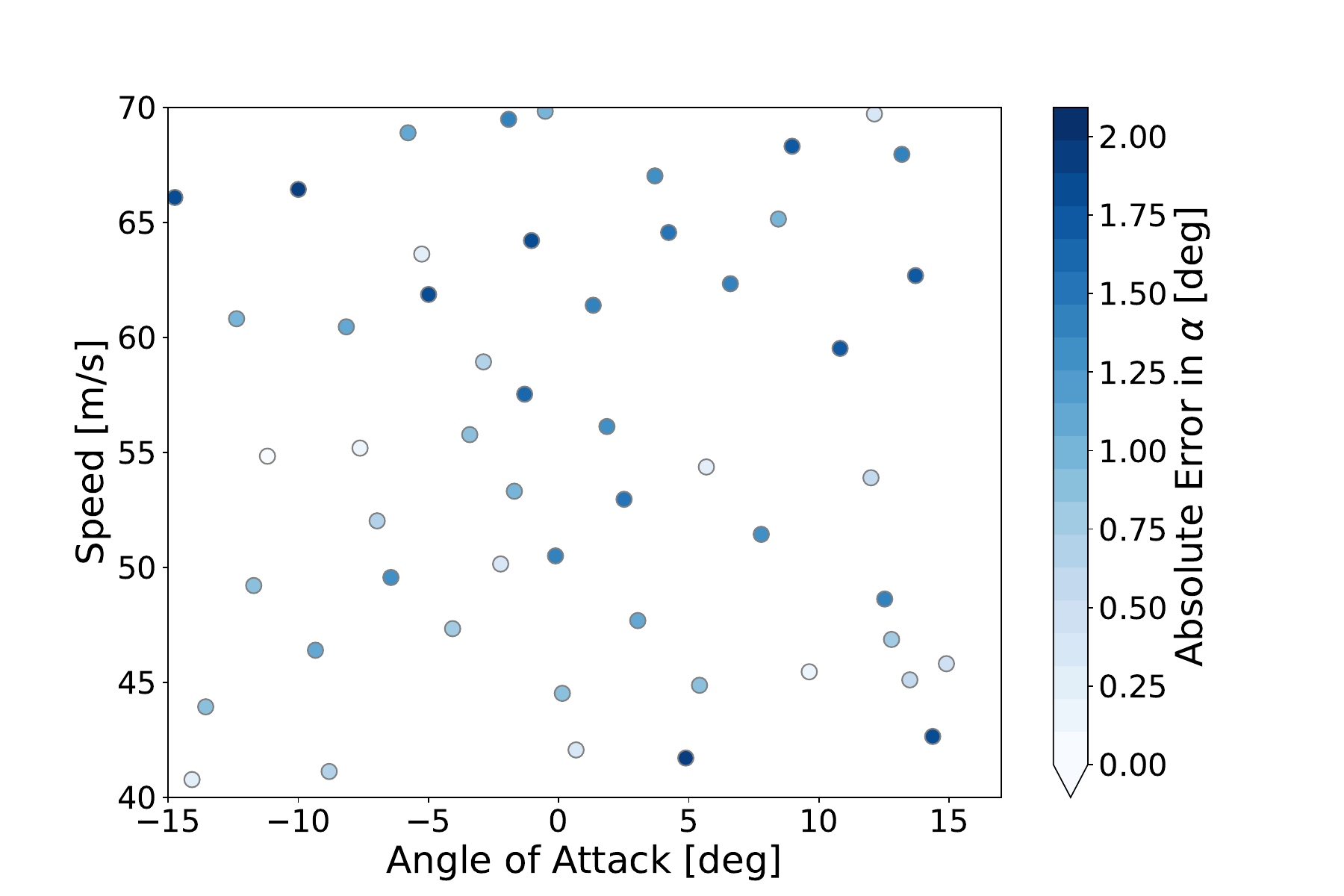} 	
\caption{$N_{TL}$ on $\mathcal{D}_S$ (Conv-D)} \label{fig_ae_aoa_tl2_convD_to}
\end{subfigure} \hfill
\begin{subfigure}[b]{0.48\textwidth}
\includegraphics[width=.99\columnwidth]{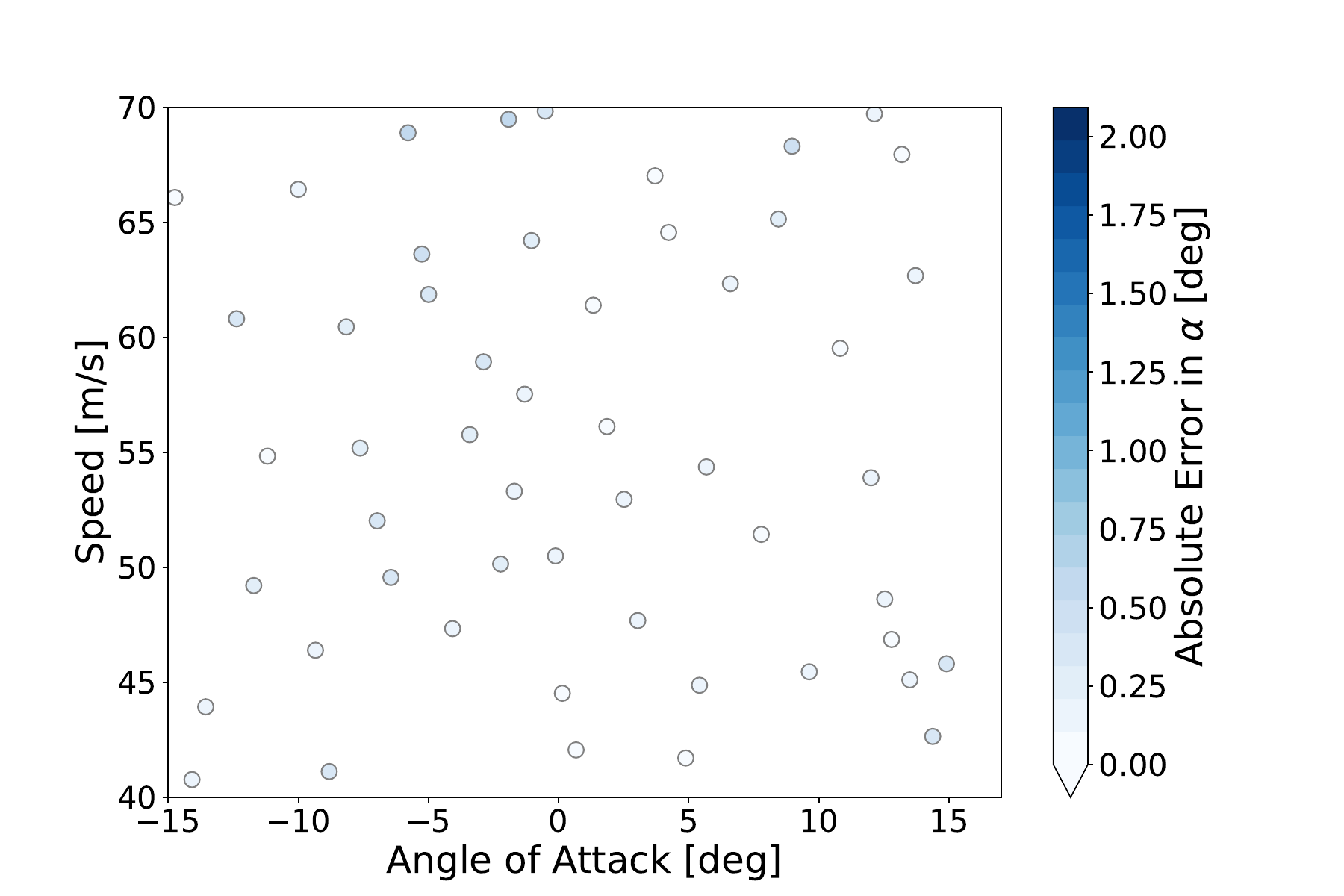} 	
\caption{$N_{TL}$ on $\mathcal{D}_R$ (Conv-D)} \label{fig_ae_aoa_tl2_convD_tt}
\end{subfigure} \hfill \vspace{4mm} \break 
\caption{Test set absolute errors obtained with $N_{TL}$ for transfer learning between the datasets generated using different turbulence models and the task of predicting $\alpha$ ($n_s=75$, $n_d=1024$)}
\label{fig_ae_aoa_tl2_t}  
\end{figure} 

\begin{figure}[!t] \centering
\begin{subfigure}[b]{0.48\textwidth}
\includegraphics[width=.99\columnwidth]{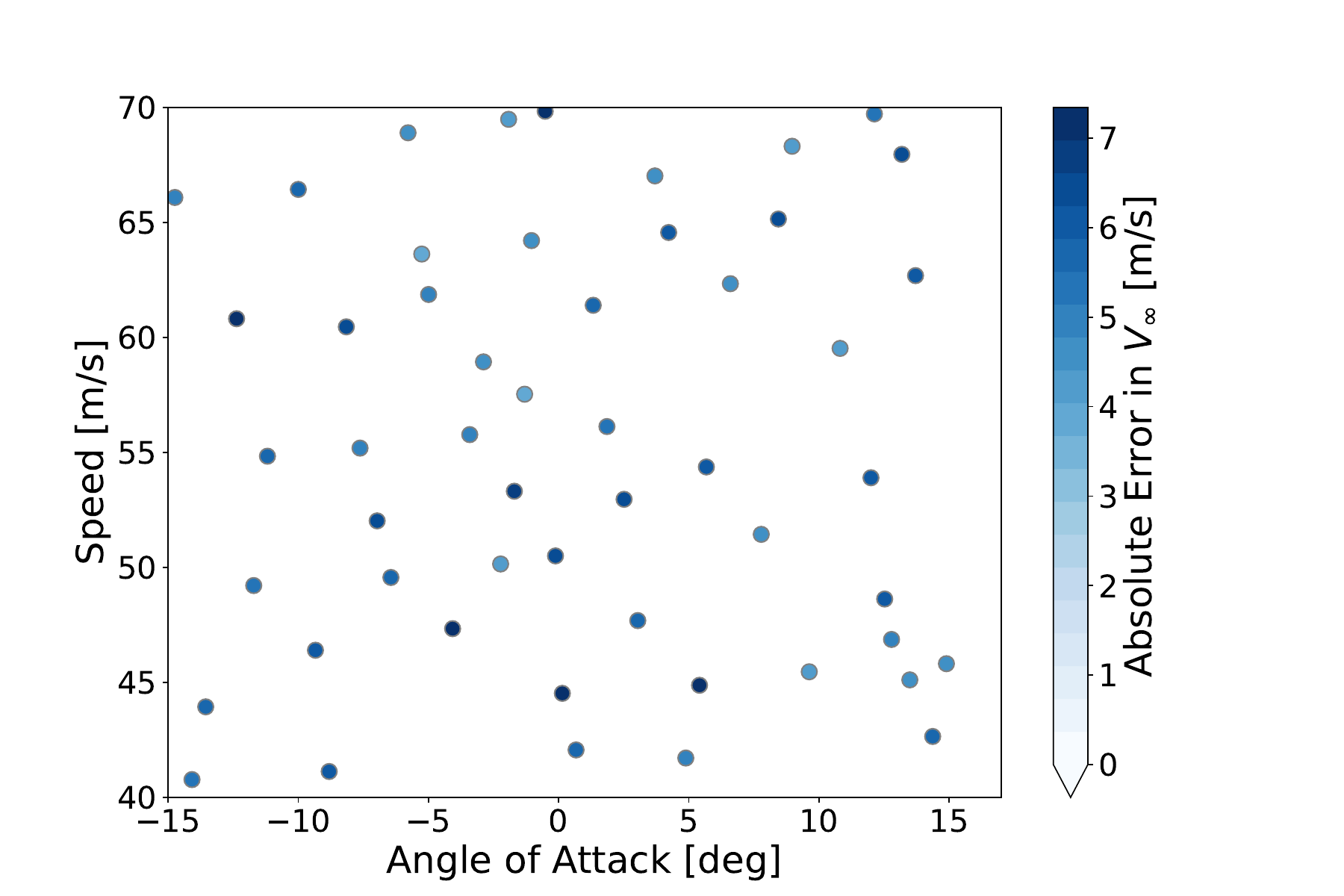} 	
\caption{$N_{TL}$ on $\mathcal{D}_S$ (Conv-S)} \label{fig_ae_spd_tl1_convS_to}
\end{subfigure} \hfill
\begin{subfigure}[b]{0.48\textwidth}
\includegraphics[width=.99\columnwidth]{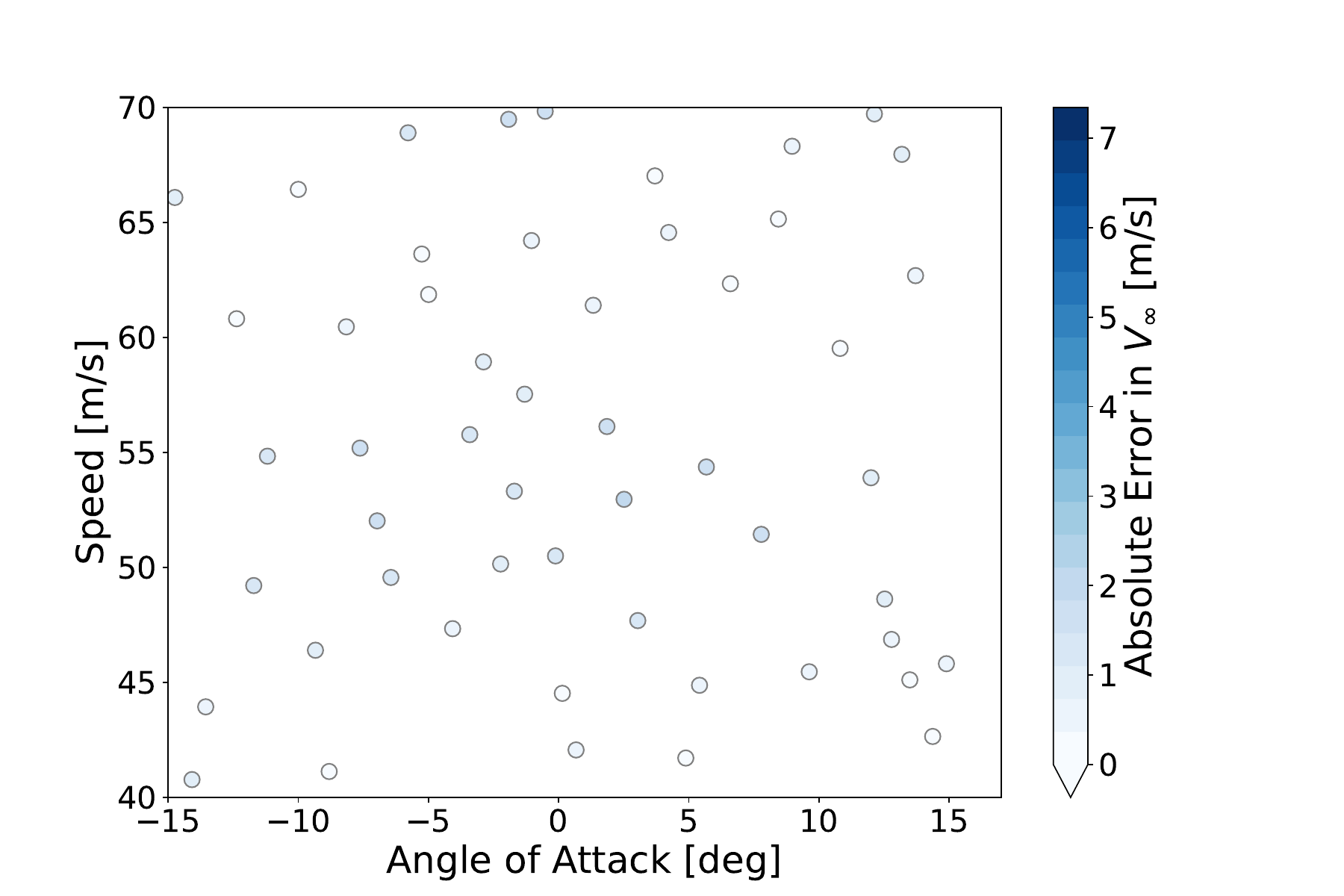} 	
\caption{$N_{TL}$ on $\mathcal{D}_R$ (Conv-S)} \label{fig_ae_spd_tl1_convS_tt}
\end{subfigure} \hfill \vspace{4mm} \break 
\begin{subfigure}[b]{0.48\textwidth}
\includegraphics[width=.99\columnwidth]{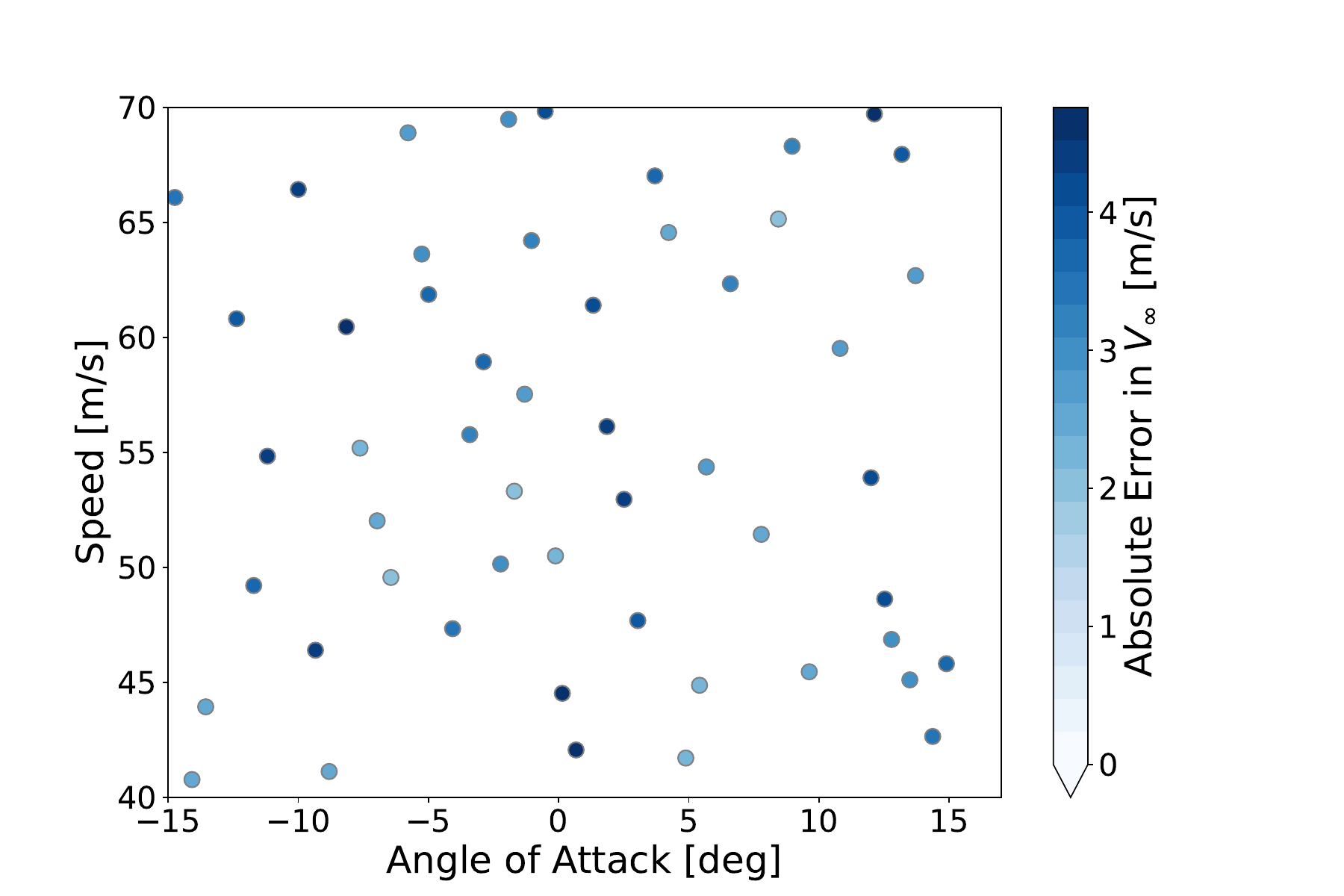} 	
\caption{$N_{TL}$ on $\mathcal{D}_S$ (Conv-D)} \label{fig_ae_spd_tl1_convD_to}
\end{subfigure} \hfill
\begin{subfigure}[b]{0.48\textwidth}
\includegraphics[width=.99\columnwidth]{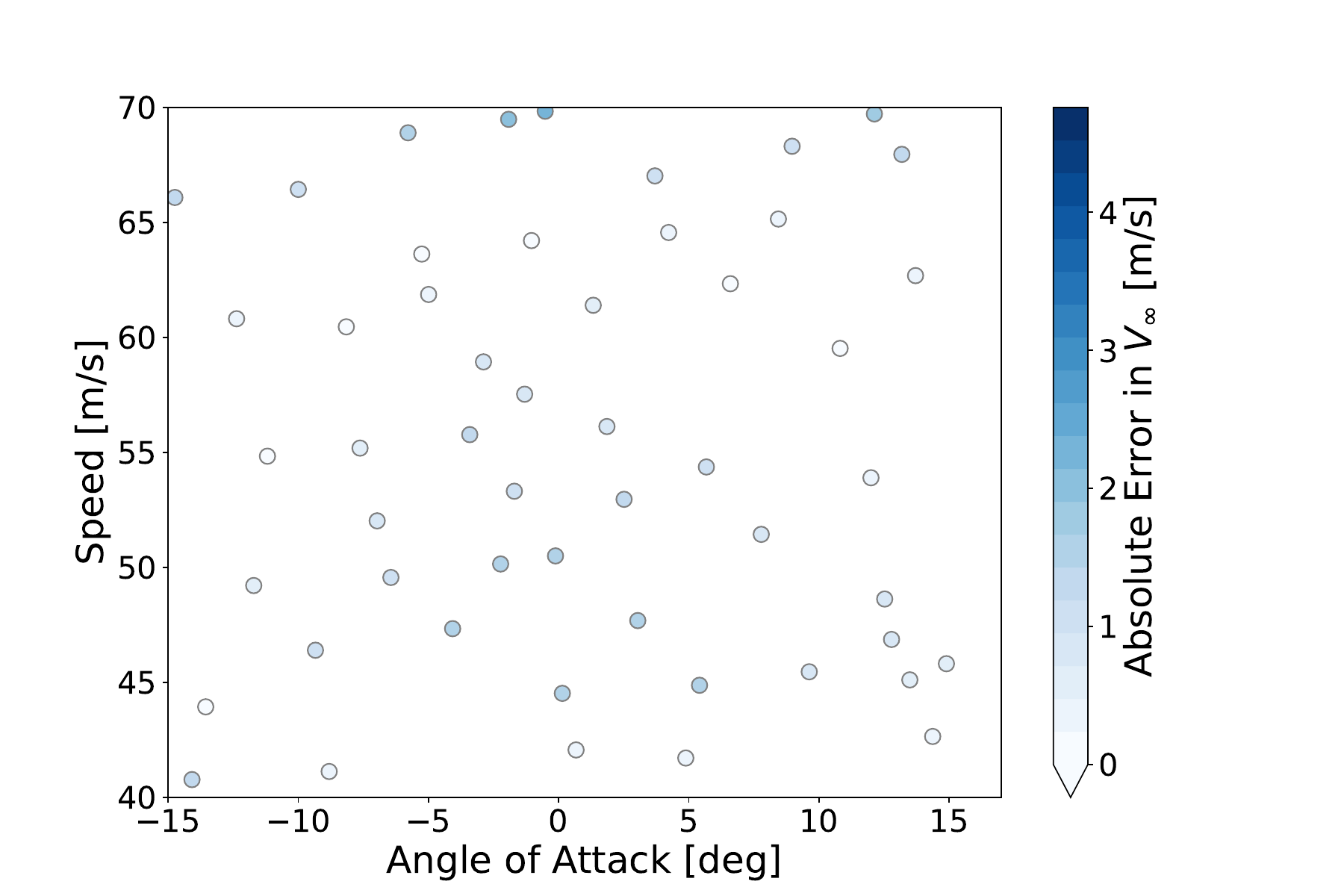} 	
\caption{$N_{TL}$ on $\mathcal{D}_R$ (Conv-D)} \label{fig_ae_spd_tl1_convD_tt}
\end{subfigure} \hfill \vspace{4mm} \break 
\caption{Test set absolute errors obtained with $N_{TL}$ for transfer learning between the datasets generated using different turbulence models and the task of predicting $V_{\infty}$ ($n_s=75$, $n_d=1024$)}
\label{fig_ae_spd_tl2_t}  
\end{figure}

The first domain adaptation case explored in this paper is transferring the learned knowledge from one data domain to another which has a different data distribution. An important prerequisite for successfully employing transfer learning for this case is that the data distributions are related. As explained earlier in Section~\ref{section_data_generation}, the TAU flow solver is employed to compute flow solutions around the airfoil geometry based on the RANS equations. Additionally, in order to generate the datasets with different characteristics in this case, two different turbulence models, namely the Spalart-Allmaras one equation model and a Reynolds stress model, SSG/LRR-ln$\omega$, are selected. The source domain, $\mathcal{D}_S$, on which the offline learning is executed, is generated based on the RANS equations and the SA turbulence model. The target domain, $\mathcal{D}_R$, to which the learning experience on $\mathcal{D}_S$ is transferred, is generated based on the RANS equations and using the Reynolds stress model, SSG/LRR-ln$\omega$, for turbulence modeling. This results in two different data distributions within the same bounds which are nearly identical in some parts of the domain including the cases at low angles of attack without separation. More severe differences exist in some other parts of the domain (e.g., larger angles of attack causing flow separation). A similar problem setup is most likely present when employing a pretrained model during operations whether it be wind tunnel or flight testing.

We define the representations for the neural network architectures trained during the offline and transfer learning phases as $\text{N}_{OL}$ and $\text{N}_{TL}$, respectively. For the remaining sections of the manuscript, these definitions are used consistently. As the first step in the transfer learning phase, the weights of $\text{N}_{TL}$ are initialized with the weights of $\text{N}_{OL}$. Then, the weights of the selected frozen layers are kept fixed, while the remaining weights are trained using the data from the target domain, $\mathcal{D}_R$ as explained in more detail in Section~\ref{section_methodology}. For the demonstration case presented in this subsection, the variations of the test set MAE values with network architectures are shown in Figure~\ref{fig_mae_tl2}. In these figures, the legend labels, ``$N_{OL}$ on $\mathcal{D}_S$'' and ``$N_{OL}$ on $\mathcal{D}_R$'', correspond to the MAE results achieved by the offline trained network, $N_{OL}$, on the test sets belonging to the source domain, $\mathcal{D}_S$, and target domain, $\mathcal{D}_R$, respectively. Similarly, the labels, ``$N_{TL}$ on $\mathcal{D}_S$'' and ``$N_{TL}$ on $\mathcal{D}_R$'', indicate the MAE results achieved by $N_{TL}$ on the test sets belonging to the aforementioned domains, $\mathcal{D}_S$ and $\mathcal{D}_R$. To investigate the influence of architecture type on individual samples, the test set absolute error values obtained with $N_{TL}$ are provided in Fig.~\ref{fig_ae_aoa_tl2_t} for $\alpha$ and in Fig.~\ref{fig_ae_spd_tl2_t} for $V_{\infty}$ for the results presented in Fig.~\ref{fig_mae_tl2}. In addition, the test set absolute error values obtained with $N_{OL}$ for this case are provided in Figures~\ref{fig_ae_aoa_tl2_o} and \ref{fig_ae_spd_tl2_o} in the Appendices. 

The test set MAE values obtained with $N_{OL}$ and $N_{TL}$ are higher for the domains, $\mathcal{D}_R$ and $\mathcal{D}_S$, respectively, on which they are not trained to predict, compared to the results obtained on their intended domains of prediction. The differences between the corresponding error values can increase by as much as one order of magnitude. Besides, it is not possible to make accurate predictions on the source domain using the networks retrained on the target domain, and there is no specific pattern for the individual test set samples as shown in Figures~\ref{fig_ae_aoa_tl2_t} and \ref{fig_ae_spd_tl2_t}. For the individual absolute error values obtained on the target domain using the offline trained networks, the predictions are not accurate for the task of predicting $V_{\infty}$ as shown in Fig.s~\ref{fig_ae_spd_tl2_t} and \ref{fig_ae_spd_tl2_o}. For the prediction task regarding $\alpha$, as shown in Fig. \ref{fig_ae_aoa_tl2_o}, $N_{OL}$ can still achieve a desirable level of accuracy and deviates from the ground truth for the higher $\alpha$ values of the domain. 

These observations imply that the domains are sufficiently different to investigate the influence of transfer learning. In general, we expect that flow field solutions obtained using different turbulence methods, as demonstrated here, result in similar qualitative aerodynamic behavior as well as shifts and deviation of performance variables over the investigated angle of attack and speed ranges. For example, flow separation occurs as a nonlinear phenomenon in the flow field solutions regardless of the used turbulence model, and the angle of attack at which this phenomenon occurs can differ for different turbulence models. For such cases, transfer learning is a viable approach as demonstrated in this subsection.

\subsection{Adaptation to a domain extension}
\begin{figure}[!t] \centering
\begin{subfigure}[b]{0.48\textwidth}
\includegraphics[width=.99\columnwidth]{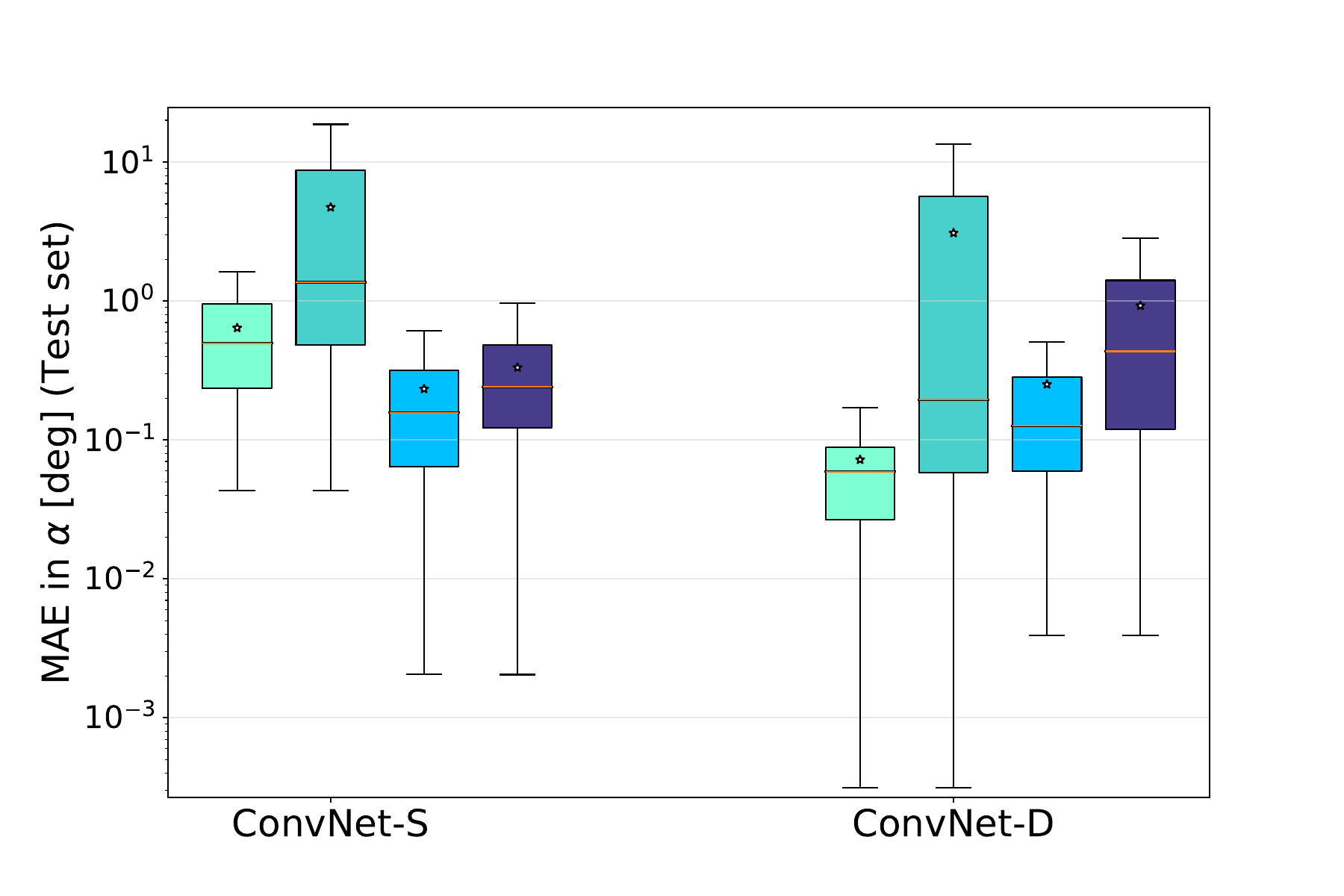} 	
\caption{Task: Predicting $\alpha$} \label{fig_mae_aoa_tl1}
\end{subfigure} \hfill
\begin{subfigure}[b]{0.48\textwidth}
\includegraphics[width=.99\columnwidth]{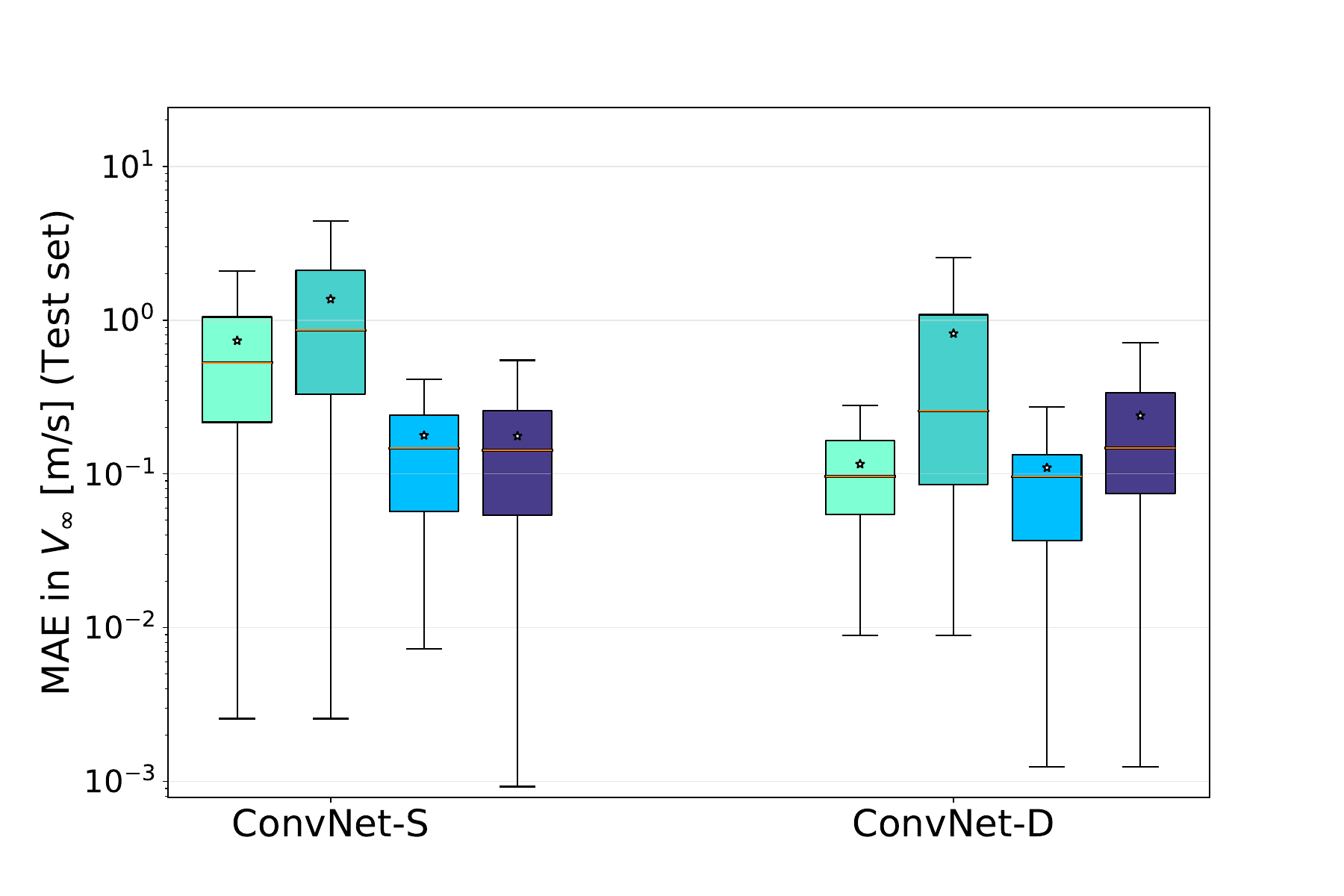} 	
\caption{Task: Predicting $V_{\infty}$} \label{fig_mae_spd_tl1}
\end{subfigure} \hfill \vspace{4mm} \break 
\begin{subfigure}[b]{0.7\textwidth}
\includegraphics[width=.99\columnwidth]{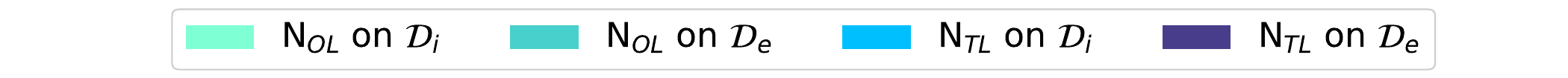}
\end{subfigure}  
\caption{Test set MAE values for transfer learning beyond the boundaries of the source domain ($n_s=75$, $n_d=1024$)} \label{fig_mae_tl1}  
\end{figure} 

\begin{figure}[!t] \centering
\begin{subfigure}[b]{0.48\textwidth}
\includegraphics[width=.99\columnwidth]{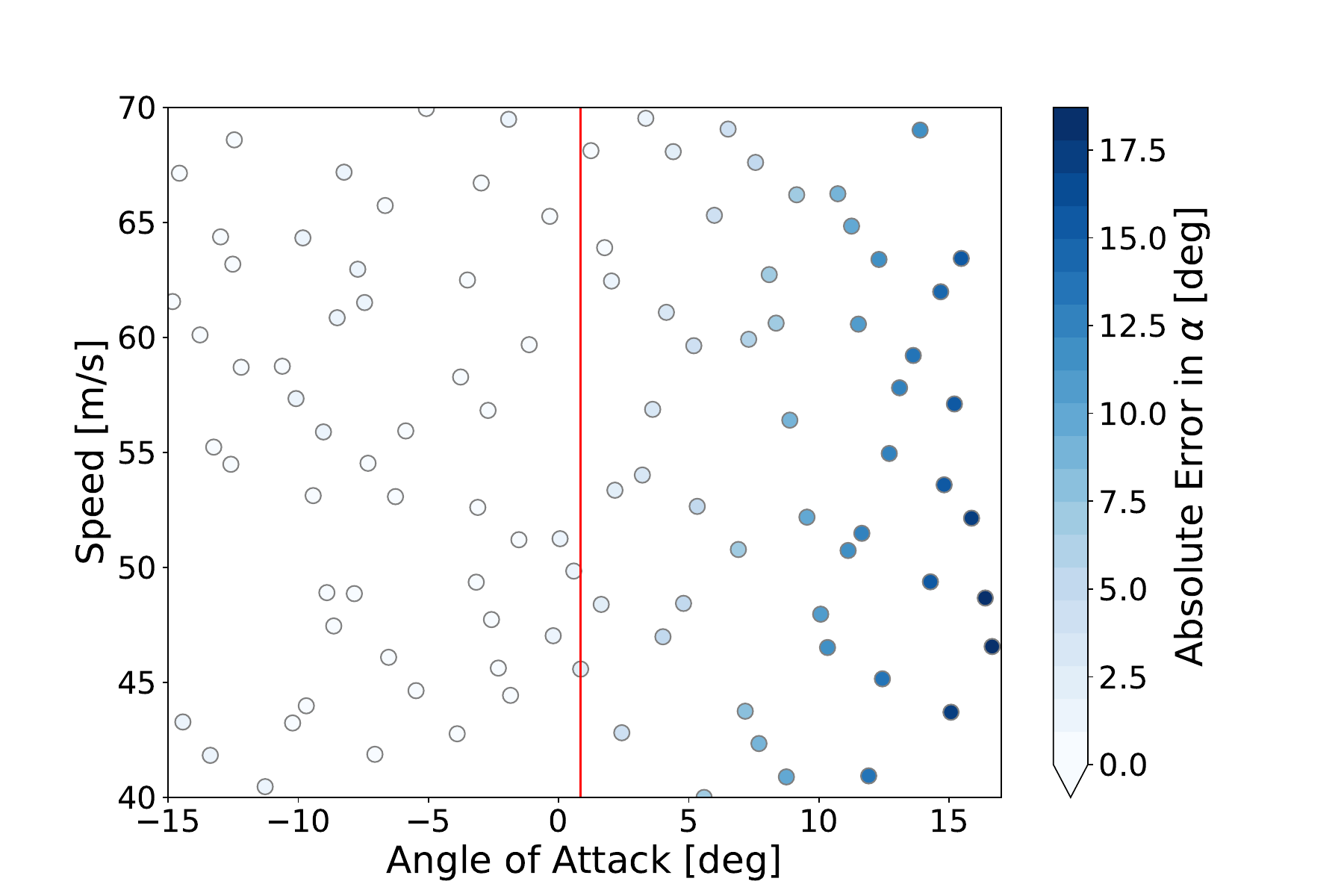}
\caption{Offline Training (Conv-S)} \label{fig_ae_aoa_tl1_convS_o}
\end{subfigure} \hfill
\begin{subfigure}[b]{0.48\textwidth}
\includegraphics[width=.99\columnwidth]{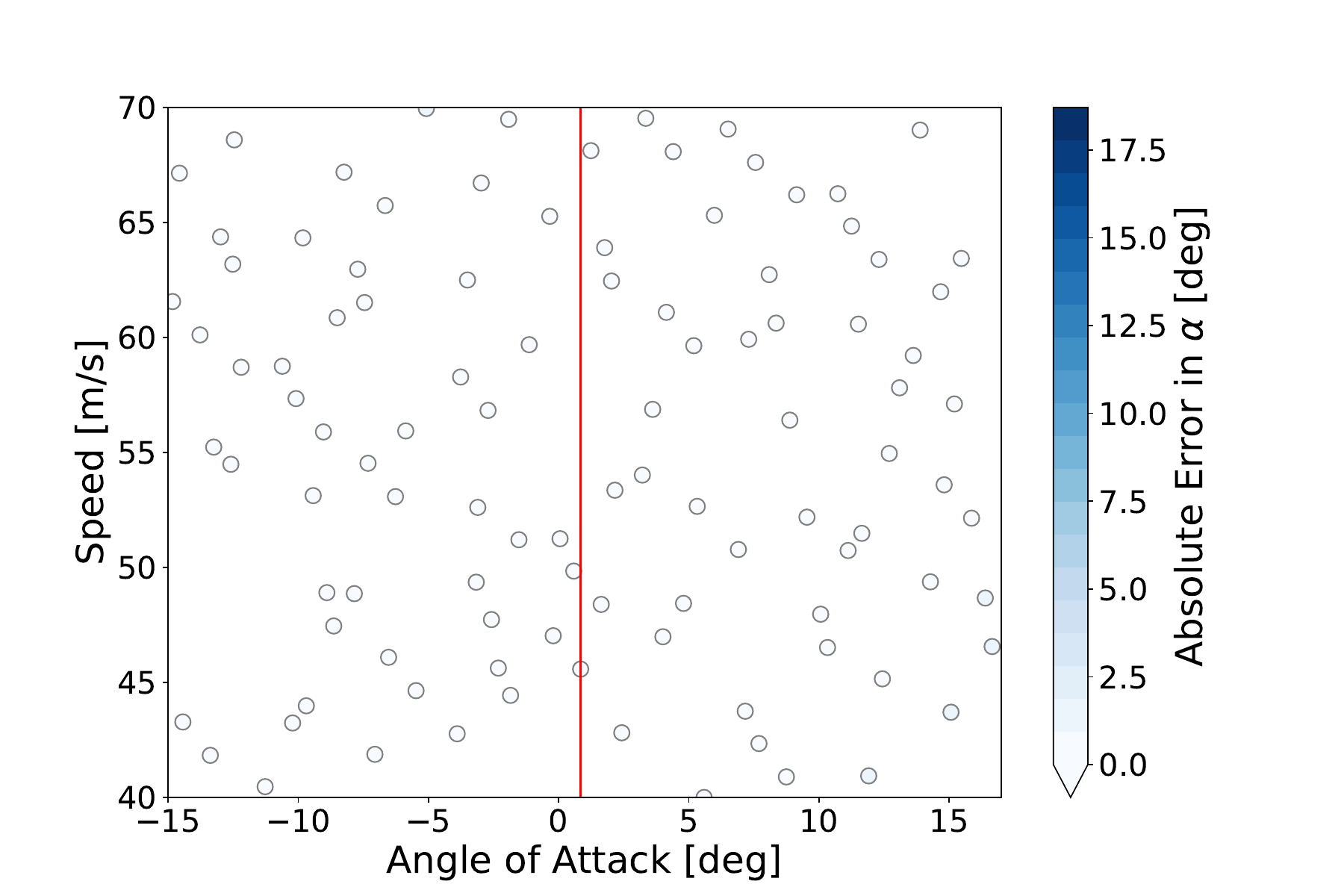} 	
\caption{Transfer Learning (Conv-S)} \label{fig_ae_aoa_tl1_convS_t}
\end{subfigure} \hfill \vspace{4mm} \break 
\begin{subfigure}[b]{0.48\textwidth}
\includegraphics[width=.99\columnwidth]{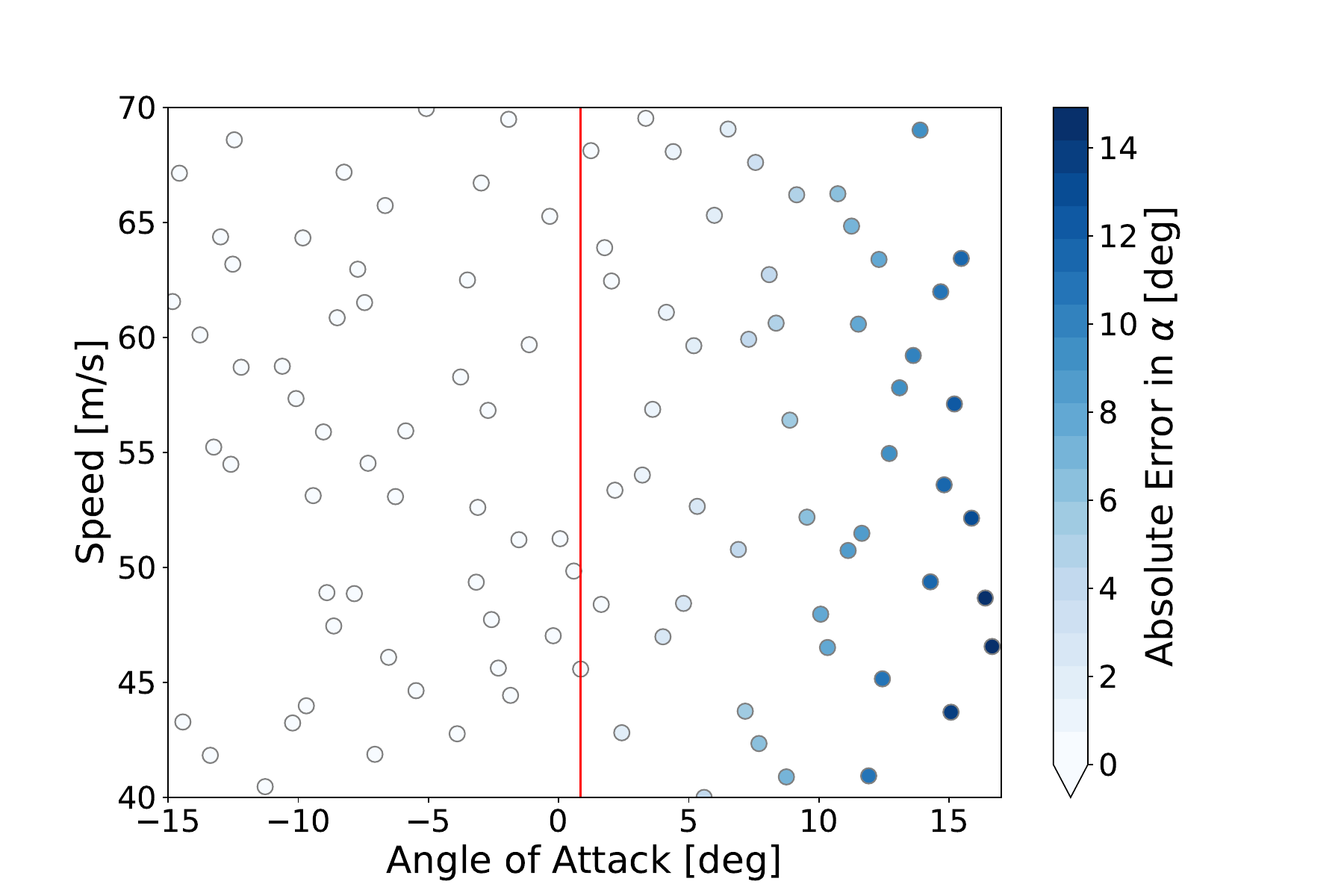} 	
\caption{Offline Training (Conv-D)} \label{fig_ae_aoa_tl1_convD_o}
\end{subfigure} \hfill
\begin{subfigure}[b]{0.48\textwidth}
\includegraphics[width=.99\columnwidth]{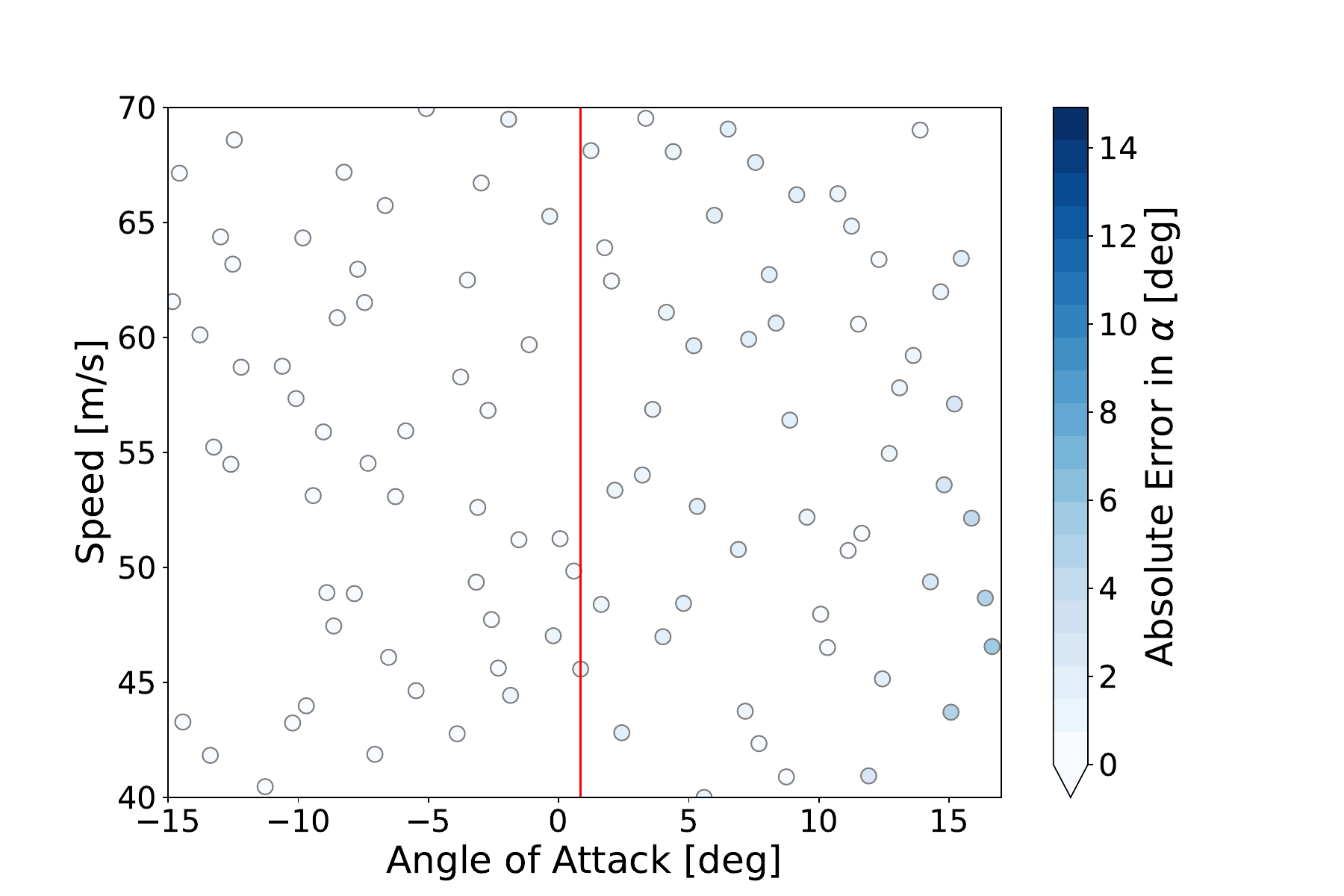} 	
\caption{Transfer Learning (Conv-D)} \label{fig_ae_aoa_tl1_convD_t}
\end{subfigure} \hfill \vspace{4mm} \break 
\caption{Test set absolute errors for transfer learning beyond the boundaries of the source domain for the task of predicting $\alpha$ ($n_s=75$, $n_d=1024$)} \label{fig_ae_aoa_tl1}  
\end{figure} 

\begin{figure}[!t] \centering
\begin{subfigure}[b]{0.48\textwidth}
\includegraphics[width=.99\columnwidth]{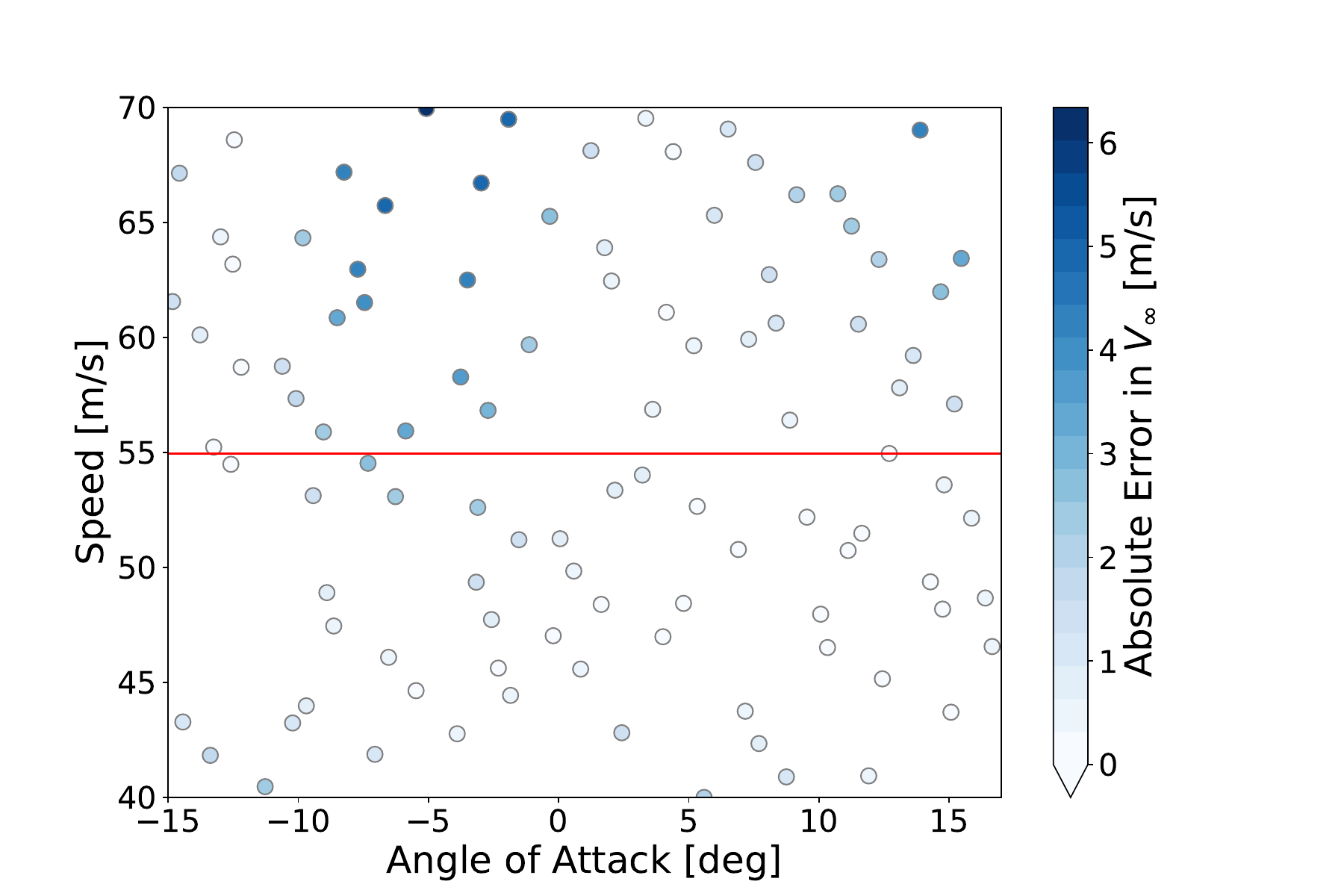}
\caption{Offline Training (Conv-S)} \label{fig_ae_spd_tl1_convS_o}
\end{subfigure} \hfill
\begin{subfigure}[b]{0.48\textwidth}
\includegraphics[width=.99\columnwidth]{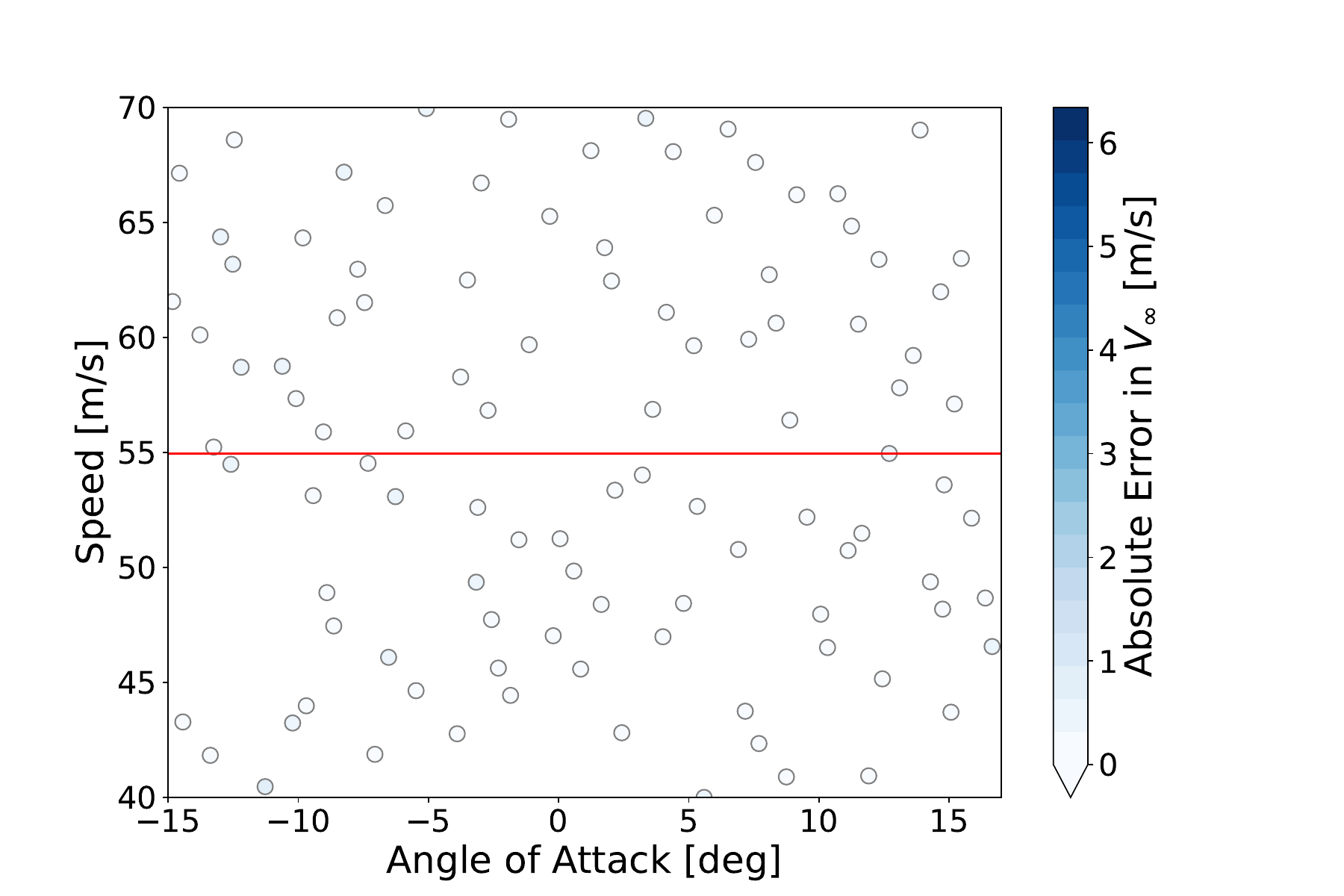} 	
\caption{Transfer Learning (Conv-S)} \label{fig_ae_spd_tl1_convS_t}
\end{subfigure} \hfill \vspace{4mm} \break 
\begin{subfigure}[b]{0.48\textwidth}
\includegraphics[width=.99\columnwidth]{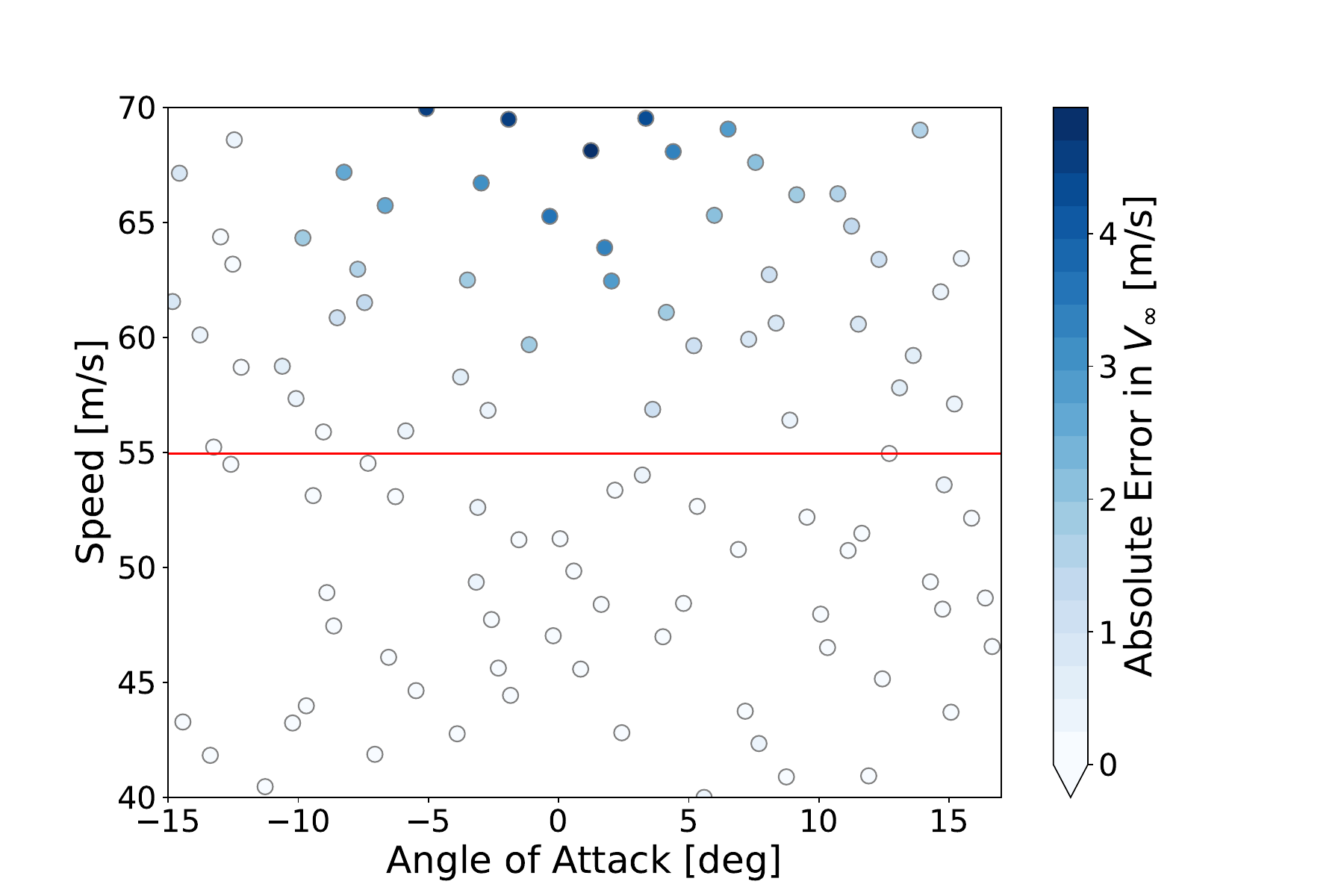}
\caption{Offline Training (Conv-D)} \label{fig_ae_spd_tl1_convD_o}
\end{subfigure} \hfill
\begin{subfigure}[b]{0.48\textwidth}
\includegraphics[width=.99\columnwidth]{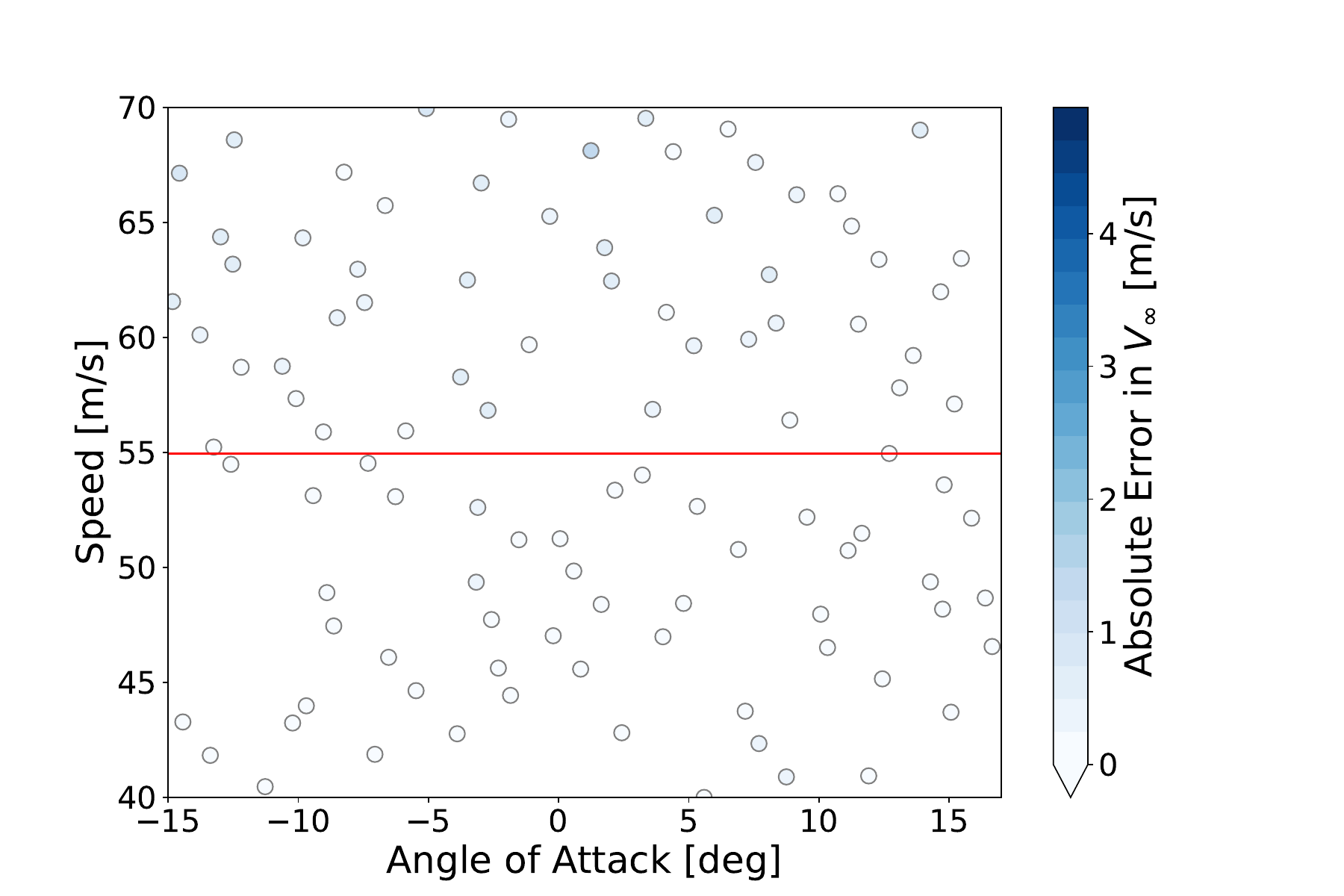} 	
\caption{Transfer Learning (Conv-D)} \label{fig_ae_spd_tl1_convD_t}
\end{subfigure} \hfill \vspace{4mm} \break 
\caption{Test set absolute errors for transfer learning beyond the boundaries of the source domain for the task of predicting $V_{\infty}$ ($n_s=75$, $n_d=1024$)} \label{fig_ae_spd_tl1}  
\end{figure}

In this section, the results for transfer learning beyond the boundaries of a data domain are presented. The initial domain used for the offline learning step, $\mathcal{D}_i$, and the extended domain, $\mathcal{D}_e$ included for the transfer learning phase are created based on the output values of their elements. Further details about how to do the sorting and splitting operations for this case are explained in Sec.\ref{subsec_data_preprocess}. Once the training, validation, and test sets on these domains are created, the offline training is first performed on $\mathcal{D}_i$, and then the knowledge is transferred to the entire domain. From an engineering point of view, this corresponds to training an initial model based on numerical simulations that did not yield any results beyond a certain point, e.g. due to instabilities or lacking convergence, while the actual model is exposed to such conditions during (physical) testing or operations.

For this demonstration case, the test set MAE results are shown in Fig.~\ref{fig_mae_tl1} for the tasks of predicting $\alpha$ and $V_{\infty}$. The corresponding network architecture is included in the horizontal axis of the subfigures. The test set MAEs obtained by using the offline trained network on the initial and extended domains are represented by ``$N_{OL}$ on $\mathcal{D}_i$'' and ``$N_{OL}$ on $\mathcal{D}_e$'', respectively. Similarly, the test set results obtained using the network trained during the transfer learning phase are labeled as ``$N_{TL}$ on $\mathcal{D}_i$'' for the initial domain and ``$N_{TL}$ on $\mathcal{D}_e$'' for the extended domain. Besides, we present the individual absolute error values for each element in the test set on a grid of $\alpha$ and $V_{\infty}$ values. The corresponding results for the tasks of predicting $\alpha$ and $V_{\infty}$ are shown in Figures~\ref{fig_ae_aoa_tl1} and \ref{fig_ae_spd_tl1}, respectively. 

Since the extended domain is not included during the offline training phase, the test set MAE results obtained on $\mathcal{D}_i$ by employing $N_{OL}$ are lower than the results obtained on $\mathcal{D}_e$. The offline training results in Figures~\ref{fig_ae_aoa_tl1} and \ref{fig_ae_spd_tl1} also clearly indicate how the offline trained network models fail to predict for individual test set samples. Specifically, for the case of predicting $\alpha$, the absolute error values increase for greater values of $\alpha$ on the extended domain, where flow separation and unsteady effects take over. While this pattern is common for both architectures when predicting $\alpha$, the regions of the extended domain with greater test set absolute error values do not overlap for the task of predicting $V_{\infty}$. Overall, these outcomes suggest that the extrapolation performances of the offline learned models do not suffice for the prediction tasks. Although the initial domain is still included in the training and validation sets during the transfer learning phase, the capability of the network to predict on the initial domain is only improved for the shallow architecture. For the dense architecture, transfer learning results in either a similar performance as in the case of predicting $V_{\infty}$ or degradation as for predicting $\alpha$. For both architectures, the MAE results obtained on the extended domain improve significantly after transfer learning, and the MAE values are even reduced one order of magnitude for the shallow architecture. We assume that these observations stem from the percentage of weights that we enable to retrain during transfer learning, which is lower for the dense architecture, and hence, this architecture does not have as much capability as the shallow one to adjust to changes. On the contrary, the dense network is a better approximator than the shallow one when trained offline with data from the initial domain because it has more weights that can be tuned accordingly. Furthermore, changing the data distribution as done in the previous demonstration case yields a more common task for the models to adapt to during transfer learning while this seems to be more complicated for adapting to a domain extension.

\subsection{Adaptation to a noisy data domain}
\begin{figure}[!t] \centering
\begin{subfigure}[b]{0.48\textwidth}
  \includegraphics[width=.99\columnwidth]{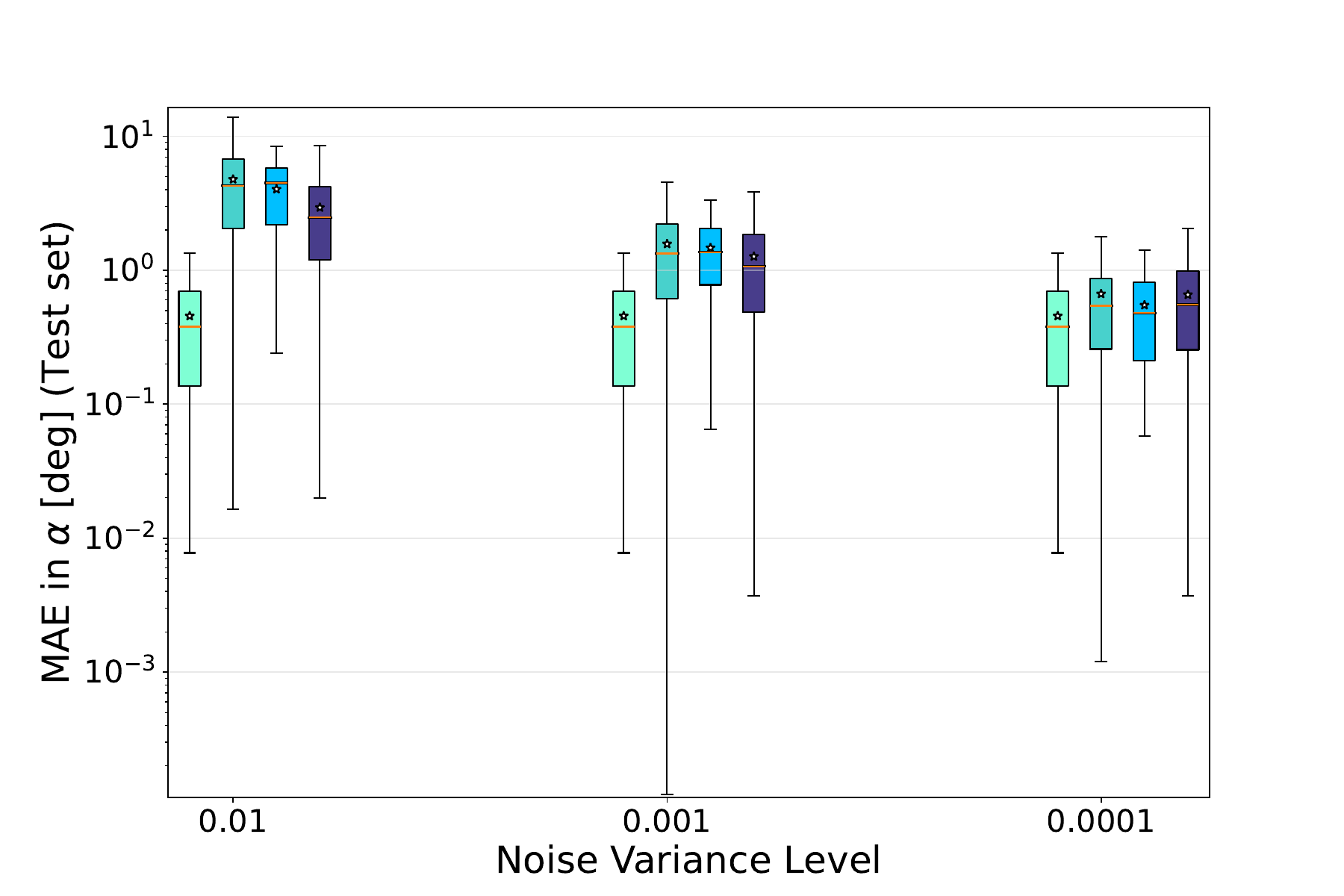} 	
  \caption{ConvNet-S} \label{fig_mae_aoa_noise_convS}
\end{subfigure} \hfill
\begin{subfigure}[b]{0.48\textwidth}
  \includegraphics[width=.99\columnwidth]{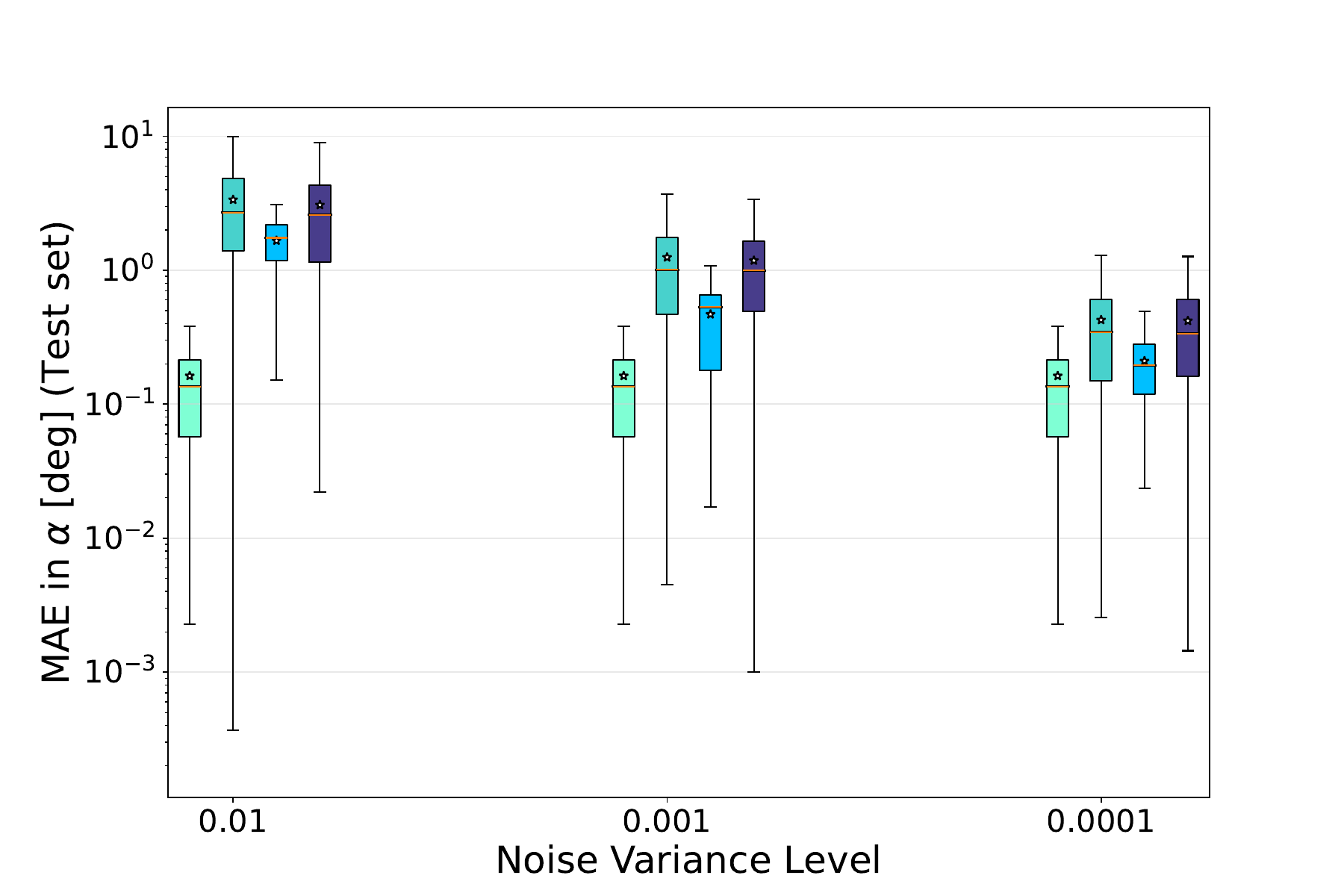} 	
  \caption{ConvNet-D} \label{fig_mae_aoa_noise_convD}
\end{subfigure} 
\begin{subfigure}[b]{0.74\textwidth}
  \includegraphics[width=.99\columnwidth]{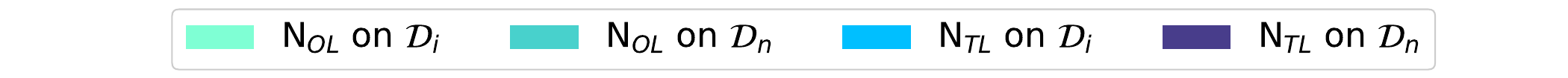}
\end{subfigure}  
\caption{The variation of the test set MAEs with noise levels for the task of predicting $\alpha$ ($n_s=75$, $n_d=1024$)} \label{fig_mae_aoa_noise}  
\end{figure} 

\begin{figure}[!t] \centering
\begin{subfigure}[b]{0.48\textwidth}
\includegraphics[width=.99\columnwidth]{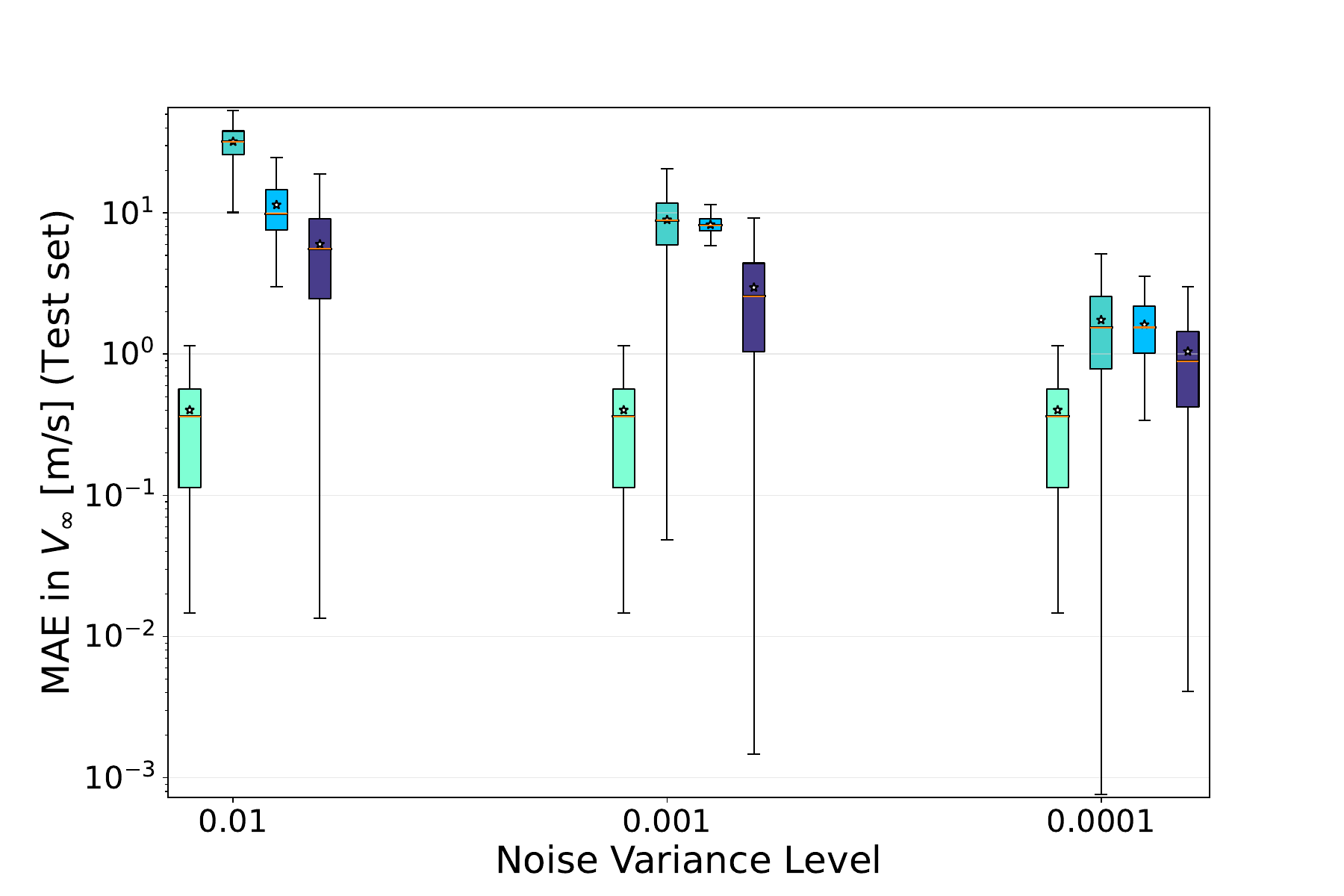} 	
\caption{ConvNet-S} \label{fig_mae_vinf_noise_convS}
\end{subfigure} \hfill
\begin{subfigure}[b]{0.48\textwidth}
\includegraphics[width=.99\columnwidth]{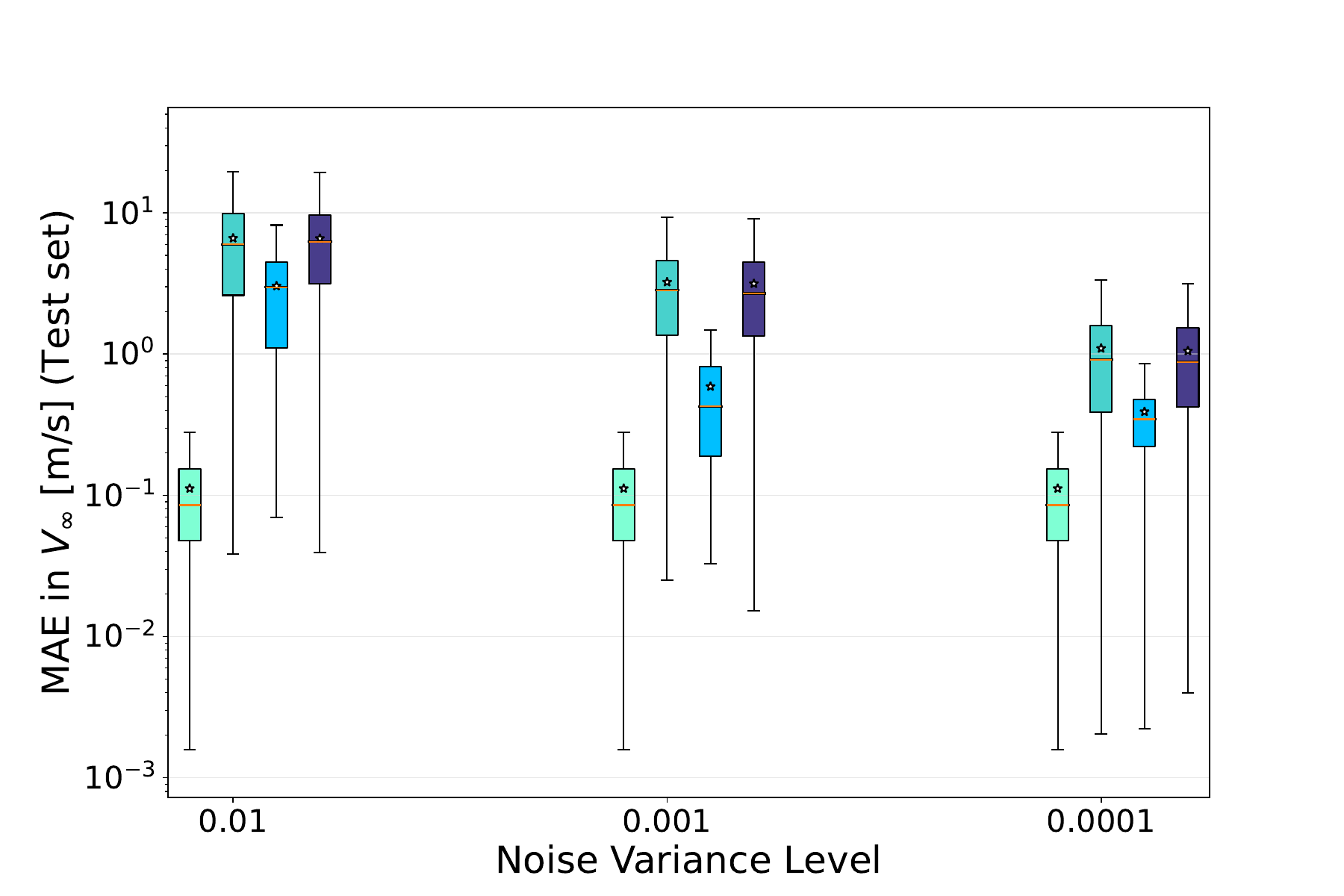} 	
\caption{ConvNet-D} \label{fig_mae_vinf_noise_convD}
\end{subfigure} \vspace{1mm} 
\begin{subfigure}[b]{0.74\textwidth}
\includegraphics[width=.99\columnwidth]{figs/results1/noise/no_test_error_legend.pdf}
\end{subfigure}
\caption{The variation of the test set MAEs with noise levels for the task of predicting $V_{\infty}$ ($n_s=75$, $n_d=1024$)} \label{fig_mae_vinf_noise}
\end{figure} 

As the final demonstration case for domain adaptation, we investigate the possibility of first learning without noise and then leveraging this offline trained model to learn on a noisy domain via transfer learning. For this purpose, a zero-mean, independent, and identically distributed Gaussian noise profile is assumed for each surface data point and added to the input data. This corresponds to receiving noisy readings from pressure sensors in an experiment or during an operational scenario. For each sample in the dataset, multiple noisy input data are considered to include the influence of the noise variance levels. For this specific study, 10 noisy inputs are generated for each sample, and therefore, these inputs correspond to the same output value. The network model is initially trained offline using the data from the noise-free domain labeled as $\mathcal{D}_i$. Next, the weights of the selected frozen layers in the network model are fixed, and the remaining weights are retrained using the data from the noisy domain, which is represented by $\mathcal{D}_n$, during the transfer learning phase. For the presented results, a set of variance values are considered. For the same prediction task and network architecture, each transfer learning phase with a different variance level is initiated using the same offline trained model for the corresponding task and architecture.

The results for this case are shown in Figures~\ref{fig_mae_aoa_noise} and \ref{fig_mae_vinf_noise} for the task of predicting $\alpha$ and $V_{\infty}$, respectively. For both prediction tasks, the MAE values obtained with the noisy data after the transfer learning phase are much higher than the values corresponding to the offline learning phase in this demonstration case. Furthermore, these values are also higher than the values obtained during the the transfer learning phases of the other demonstration cases. Since the weights of the unfrozen layers are updated during the transfer learning phase, the prediction accuracy obtained with $N_{TL}$ on the noise-free domain degrades as well. The degradation increases for both shallow and dense networks as the noise variance levels increase. When using the ConvNet-D architecture during the transfer learning phase for both prediction tasks, the MAE values obtained on the noisy domain, $\mathcal{D}_n$, are still higher than the values obtained on the noise-free domain, $\mathcal{D}_i$. The corresponding values obtained on $\mathcal{D}_n$ for the shallow architecture are, however, lower than the MAE values obtained on $\mathcal{D}_i$.

\subsection{Adaptation to task change}
\begin{figure}[!t] \centering
\begin{subfigure}[b]{0.48\textwidth}
\includegraphics[width=.99\columnwidth]{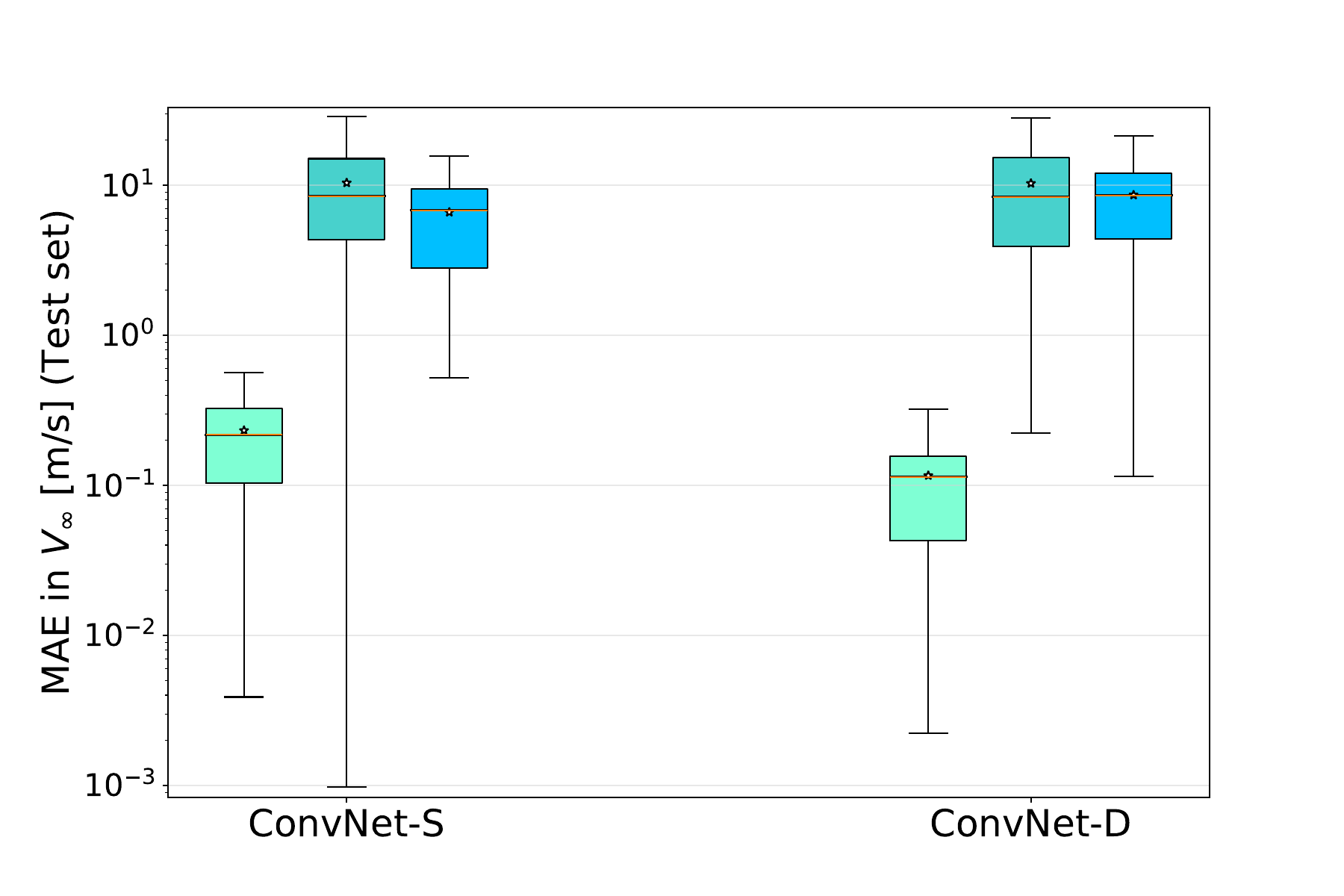}
\caption{Training 1 layer for transfer learning} \label{fig_mae_vinf_tl4_convS}
\end{subfigure} \hfill
\begin{subfigure}[b]{0.48\textwidth}
\includegraphics[width=.99\columnwidth]{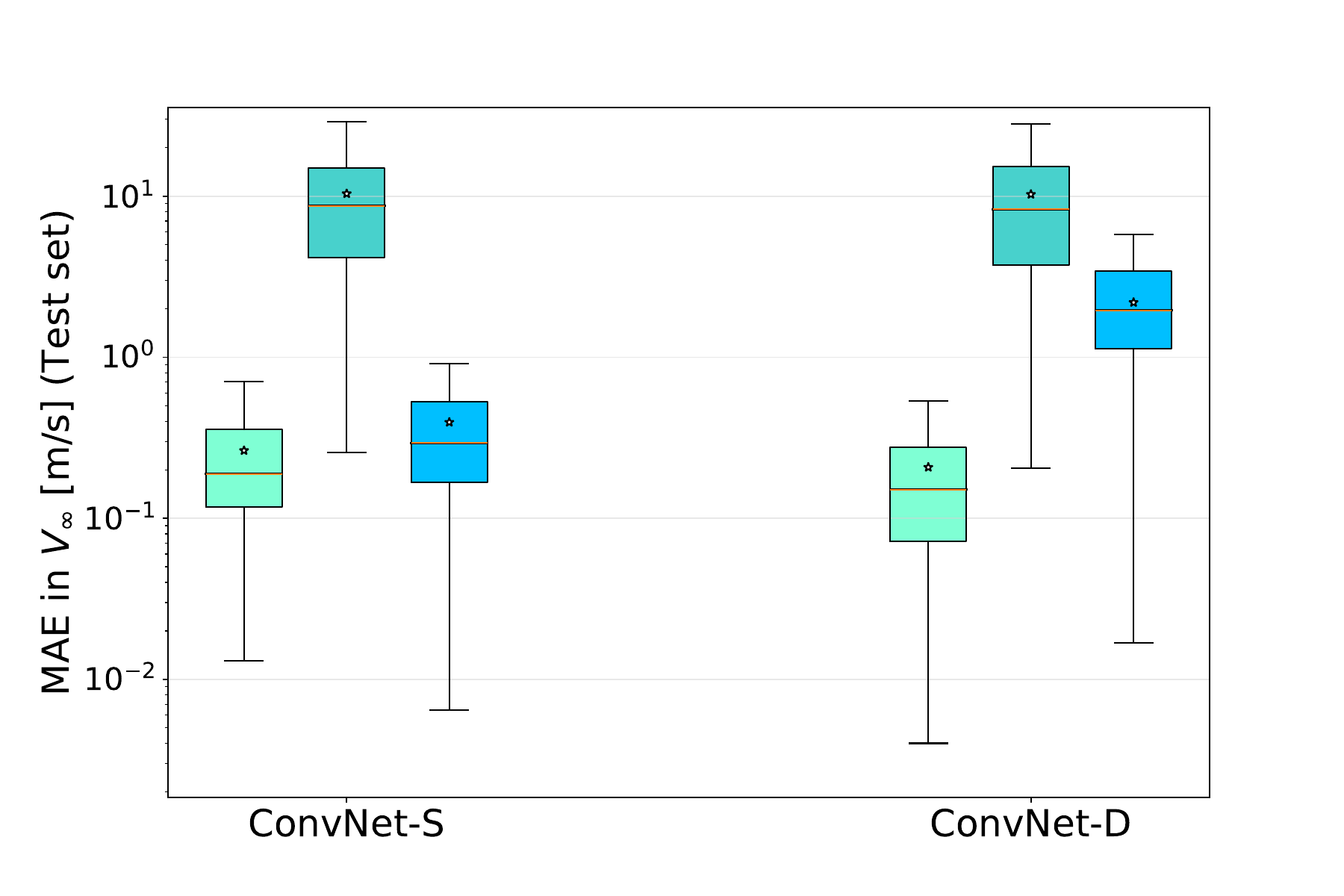}
\caption{Training 2 layers for transfer learning} \label{fig_mae_vinf_tl4_convD}
\end{subfigure} \vspace{2mm} \break 
\includegraphics[width=.7\columnwidth]{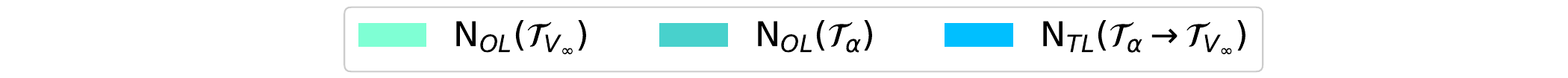}
\caption{Test set MAE values for the case of task adaptation to predicting $V_{\infty}$ ($n_s=75$, $n_d=1024$)} \label{fig_mae_vinf_tl4}
\end{figure} 

In addition to domain adaptation, transfer learning can be used for adapting to task changes. A demonstration case is presented in this subsection where the source task is chosen as predicting $\alpha$ while the target task is set to predicting $V_{\infty}$. In this case, the source and target domains are the same and denoted as $\mathcal{D}$. The network architecture is first trained on the domain, $\mathcal{D}$, to predict $\alpha$, and then transfer learning is applied on the same domain to predict $V_{\infty}$. Tasks are represented by $\tau$ with subscript referring to the predicted variable. For comparison purposes, we also include the results obtained using a network directly trained offline to predict $V_{\infty}$. The MAE values obtained for task adaptation are shown in Figure~\ref{fig_mae_vinf_tl4} for the cases of training either one or two fully-connected layers before the output layer during the transfer learning phase. In this figure, the offline trained networks for the tasks of predicting $\alpha$ and $V_{\infty}$ are represented by $N_{OL}(\tau_{\alpha})$ and $N_{OL}(\tau_{V_{\infty}})$, respectively, whereas the network, which is trained to predict $V_{\infty}$ based on $N_{OL}(\tau_{\alpha})$ during the transfer learning step, is represented by $N_{TL}(\tau_{\alpha} \to \tau_{V_{\infty}})$. 

Investigating the test set prediction accuracy for these networks, the MAE values achieved employing $N_{OL}(\tau_{V_{\infty}})$ can be deemed sufficient for a desired level of prediction accuracy in practical applications. It is also clear that the networks, $N_{OL}(\tau_{\alpha})$, which are trained offline to predict $\alpha$, fail to predict $V_{\infty}$. If one layer is retrained during the transfer learning phase, the prediction accuracy of $N_{TL}(\tau_{\alpha} \to \tau_{V_{\infty}})$ cannot be improved and performs similar to $N_{OL}(\tau_{\alpha})$ in terms of the attained MAE values. When the weights of another layer are also included in the training optimization for transfer learning, the MAE values obtained employing $N_{TL}(\tau_{\alpha} \to \tau_{V_{\infty}})$ are reduced by one order of magnitude for the shallow architecture. This network can even achieve an error performance close to $N_{OL}(\tau_{V_{\infty}})$ for this case. By training two layers in this case, we retrain 98.697\% of the weights in the shallow architecture and 70.359\% of the weights in the dense architecture. However, the weights of the convolutional layers, which are used for feature extraction purposes, are fixed for both architectures. Therefore, transfer learning is not the same as nearly fully retraining the entire network.

Although the MAE values obtained employing $N_{TL}(\tau_{\alpha} \to \tau_{V_{\infty}})$ and the dense architecture are slightly improved, the results are still one order of magnitude higher than the values obtained employing $N_{OL}(\tau_{V_{\infty}})$. This might result from the percentage of the retrained weights, which is greater for the shallow network, and thus provides extra flexibility during the training optimization. Overall, the tasks of learning different onflow parameters based on sparse surface pressure data correspond to learning two different inverse maps. Transferring the learning experience regarding these inverse maps is a more complicated deed and requires that more layers be trained for better prediction accuracy.

\subsection{Training time}
\begin{table}[!t]
\centering  \caption{Training time ($\mu \pm 1\sigma$) for the offline and transfer learning phases when adapting to changing data distributions in the domain}
\begin{tabular}{ccccccc}
\cmidrule(l r ){1-7}
Task: $\tau_{\alpha}$ & \multicolumn{3}{c}{ConvNet-S} & \multicolumn{3}{c}{ConvNet-D} \\
\cmidrule(l r ){1-1} \cmidrule(l r ){2-4} \cmidrule(l r ){5-7}
Size & $t_{OL}$ [s] & $t^{1L}_{TL}$ [s] & $t^{2L}_{TL}$ [s] & $t_{OL}$ [s] &  $t^{1L}_{TL}$ [s] & $t^{2L}_{TL}$ [s] \\
\cmidrule(l r ){1-1} \cmidrule(l r ){2-2} \cmidrule(l r ){3-3} \cmidrule(l r ){4-4} \cmidrule(l r ){5-5} \cmidrule(l r ){6-6} \cmidrule(l r ){7-7} 
 512 & 43.58 $\pm$ 6.65 & 11.03 $\pm$ 1.92 & 27.31 $\pm$ 1.04 & 142.75 $\pm$ 27.99 & 33.02 $\pm$ 4.07 & 23.30 $\pm$ 9.04 \\
1024 & 58.76 $\pm$ 11.76 & 6.24 $\pm$ 1.66 & 21.61 $\pm$ 4.73 & 154.28 $\pm$ 42.35 & 17.62 $\pm$ 6.95 & 35.53 $\pm$ 12.40 \\ \\
\cmidrule(l r ){1-7} 
Task: $\tau_{V_{\infty}}$ & \multicolumn{3}{c}{ConvNet-S} & \multicolumn{3}{c}{ConvNet-D} \\
\cmidrule(l r ){1-1} \cmidrule(l r ){2-4} \cmidrule(l r ){5-7}
Size & $t_{OL}$ [s] & $t^{1L}_{TL}$ [s] & $t^{2L}_{TL}$ [s] & $t_{OL}$ [s] &  $t^{1L}_{TL}$ [s] & $t^{2L}_{TL}$ [s] \\
\cmidrule(l r ){1-1} \cmidrule(l r ){2-2} \cmidrule(l r ){3-3} \cmidrule(l r ){4-4} \cmidrule(l r ){5-5} \cmidrule(l r ){6-6} \cmidrule(l r ){7-7}
512 & 63.21 $\pm$ 10.42 & 19.79 $\pm$ 4.11 & 14.93 $\pm$ 2.16 & 92.71 $\pm$ 22.14 & 53.60 $\pm$ 3.48 & 38.35 $\pm$ 21.01 \\
1024 & 51.10 $\pm$ 7.50 & 8.40 $\pm$ 1.44 & 13.7 $\pm$ 2.88 & 113.03 $\pm$ 40.38 & 25.84 $\pm$ 5.67 & 55.19 $\pm$ 18.90 \\
\end{tabular}  \label{table_training_time}
\end{table}

For online applications such as wind tunnel experiments or wind turbine and flight operations, the required execution times are significant factors that determine the feasibility of real-time implementations. Considering the time limitations in this context, a real-time performance is achievable by using the trained networks in this work for online prediction while it is not possible to train these neural networks in real time. Therefore, we also investigate whether the transfer learning approach can be leveraged to reduce the required training time with the ambition of approaching close to real-time performance while ensuring an acceptable level of prediction accuracy. For this purpose, we include the training time results of the demonstration case for the adaptation to changing data distributions in this section. The training time results for offline and transfer learning phases, $t_{OL}$ and $t_{TL}$, are included in Table~\ref{table_training_time} with additional information about tasks, dataset sizes, and architectures. Moreover, $1L$ and $2L$, which are the superscripts of $t_{TL}$, refer to retraining one layer and two layers during the transfer learning phase, respectively.

In general, the transfer learning methodology leads to reduced training times for both network architectures compared to a fully offline training. As the dataset size increases, the amount of reduction in training time is consistently greater for transfer learning by retraining one layer instead of retraining two layers. Nevertheless, it can be stated that the training times required by the transfer learning phases exceed the upper limits allowed for real-time learning.

\section{Conclusions} \label{section_conclusion} 
In this manuscript, we have proposed a transfer learning framework for the problem of predicting the onflow parameters, namely angle of attack and speed, based on sparse surface pressure data. The data-driven model architecture that is used to learn the mapping from the surface pressure data to the onflow parameters, is selected based on convolutional neural networks due to their efficiency in learning on data while accounting for neighboring relationships. In this framework, a ConvNet based architecture is first trained on the source domain for the source prediction task offline. Then, the weights of the layers in this architecture except the final layer are kept fixed, and the remaining weights are retrained using the data from the target domain. In this context, transfer learning enables to make accurate predictions when there is a change in the flow characteristics or the prediction parameter of interest.

For the core task of predicting the onflow parameters, the ConvNet architectures lead to better prediction performance with increasing datasize whereas fully connected neural networks are insufficient. The number of surface data points result in no consistent change in prediction accuracy. The dense network is concluded as better approximator for offline training because it has more weights that can be tuned accordingly. To investigate the benefits of using transfer learning, we have demonstrated domain and task adaptation cases. For the case of adaptation to the changes in data distribution, the prediction accuracy results achieved on the target domain using the network trained during the transfer learning phase are similar to those achieved on the source domain using the offline trained network. For the case of adaptation to a domain extension, the extrapolation performance of the offline trained network is unsatisfactory as expected. With the addition of more training data during the transfer learning phase, the prediction accuracy results on both the initial and extended domains are improved. The individual test set elements with higher absolute error values are observed to be close to the domain boundaries where nonlinear phenomenon such as flow separation occurs. The transfer learning approach is ineffective in adapting to the domain with noise leading to much higher errors compared to the other demonstration cases. For the task and domain adaptation cases, the MAE values obtained with the retrained shallow ConvNet architecture are lower. This is assumed to result from a greater percentage of unfrozen weights that are retrained during transfer learning. Hence, the shallow architecture provides more flexibility for adaptation. Finally, although the transfer learning approach leads to reduced training times considering greater dataset sizes and denser architectures in particular, real-time learning has not been achieved.

Data-driven prediction models as in this manuscript can also pave the way for potential practical uses. They can serve as redundant backup systems for online monitoring to detect faulty behavior or as a part of the voting mechanisms in flight control system architectures. A future research objective is the investigation of the applicability of the proposed method to transfer the learned experience from CFD runs to wind tunnel experiments and to make accurate predictions online during the experiments. Other future directions include the exploration of alternative ways to improve the adaptation to the domains with noise and to reduce the training times required during the transfer learning phase. Furthermore, the tasks of learning different onflow parameters based on sparse surface pressure data correspond to learning two different inverse maps, and how much these inverse maps are related to each other is a question that requires further investigation. 

\clearpage
\appendix
\section*{Appendices}
\subsection*{A.1. Architecture Parameters} 
In this subsection of the Appendices, we provide the parameters of the ConvNet-D, ConvNet-D, and FCNN architectures used for the demonstration cases. Specifically, Table \ref{table_convnetD} includes the parameter information for the ConvNet-D architecture,  Table \ref{table_convnetS} includes the parameter information for the ConvNet-S architecture, and Table \ref{table_fcnn} includes the parameter information for the FCNN architecture.

\subsection*{A.2. Test Set Prediction Accuracy of the Offline Trained Neural Networks for the Case of Adaptation to Changing Data Distributions} 
\begin{figure}[!t] \centering
\begin{subfigure}[b]{0.48\textwidth}
\includegraphics[width=.99\columnwidth]{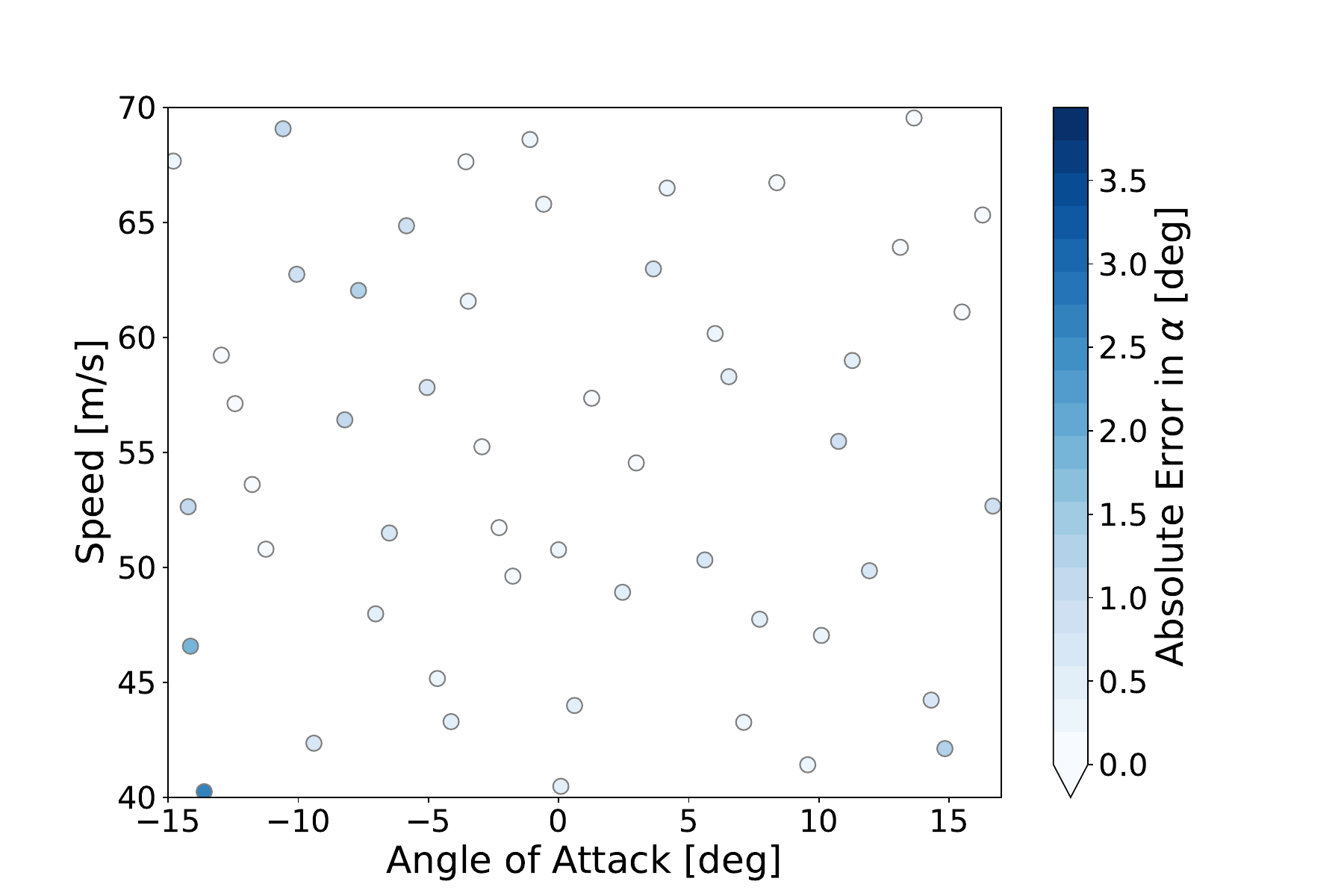} 	
\caption{$N_{OL}$ on $\mathcal{D}_S$ (Conv-S)} \label{fig_mae_aoa_tl1_convS_oo}
\end{subfigure} \hfill
\begin{subfigure}[b]{0.48\textwidth}
\includegraphics[width=.99\columnwidth]{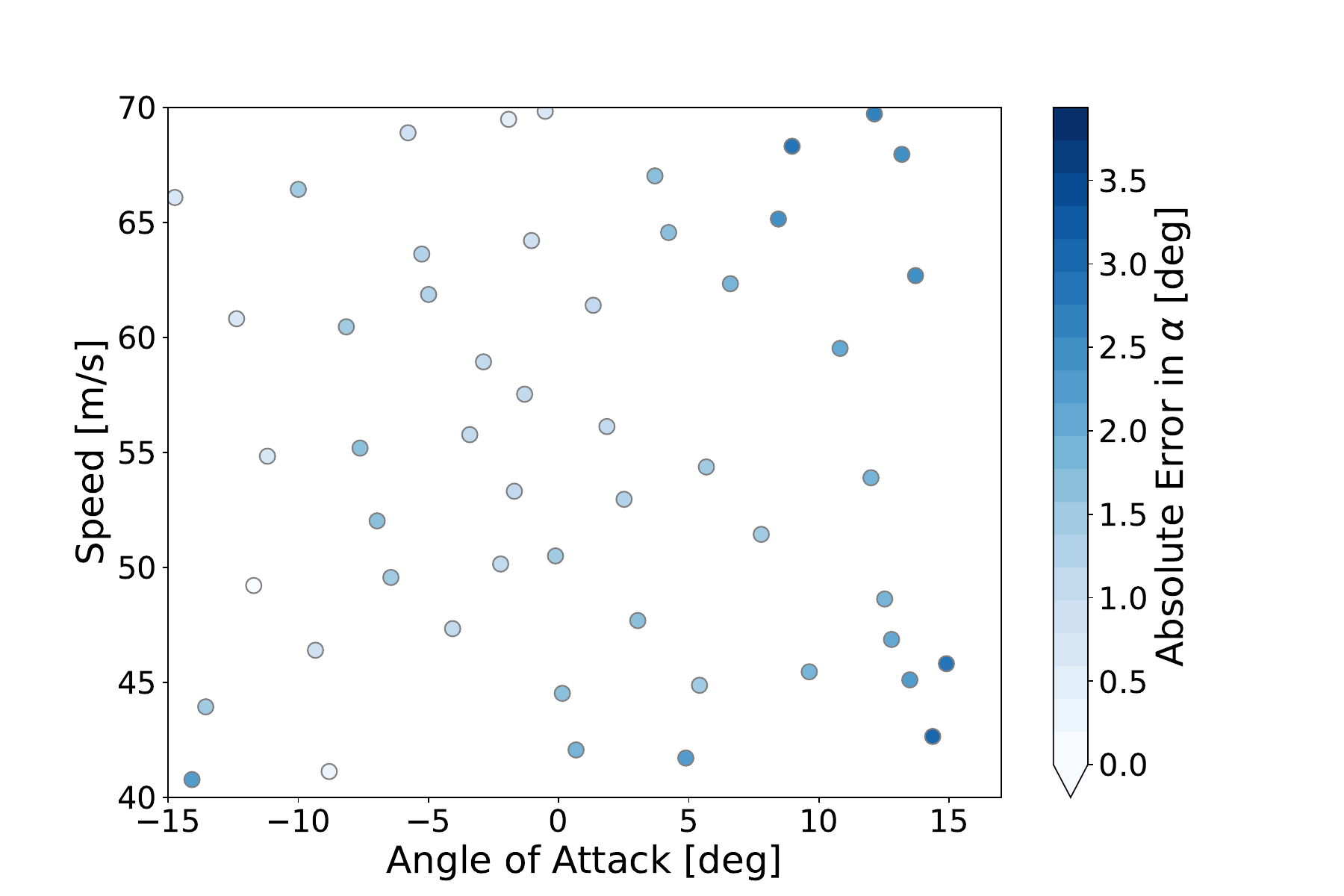} 	
\caption{$N_{OL}$ on $\mathcal{D}_R$ (Conv-S)} \label{fig_ae_aoa_tl2_convS_ot}
\end{subfigure} \hfill \vspace{4mm} \break 
\begin{subfigure}[b]{0.48\textwidth}
\includegraphics[width=.99\columnwidth]{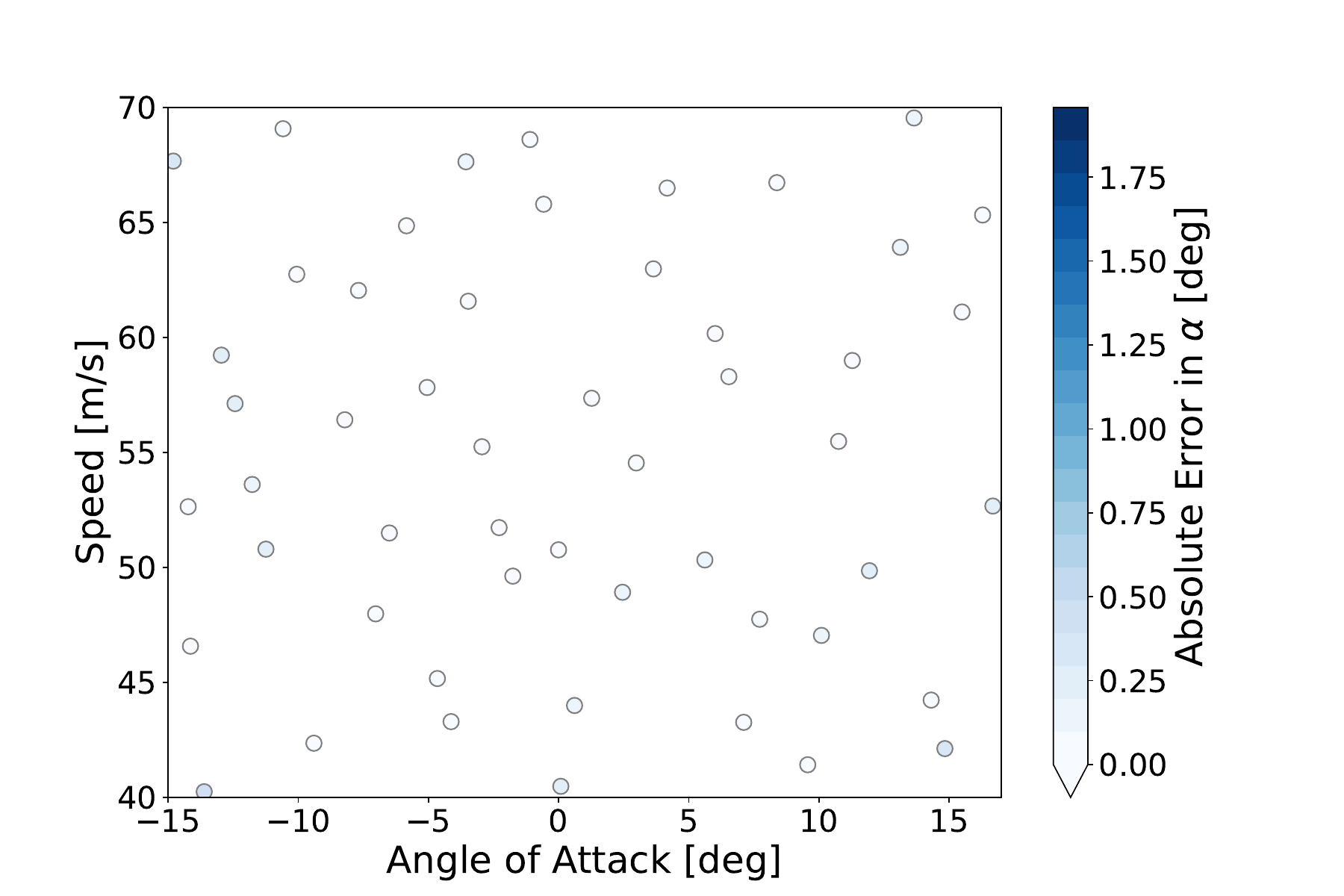} 	
\caption{$N_{OL}$ on $\mathcal{D}_S$ (Conv-D)} \label{fig_ae_aoa_tl2_convD_oo}
\end{subfigure} \hfill
\begin{subfigure}[b]{0.48\textwidth}
\includegraphics[width=.99\columnwidth]{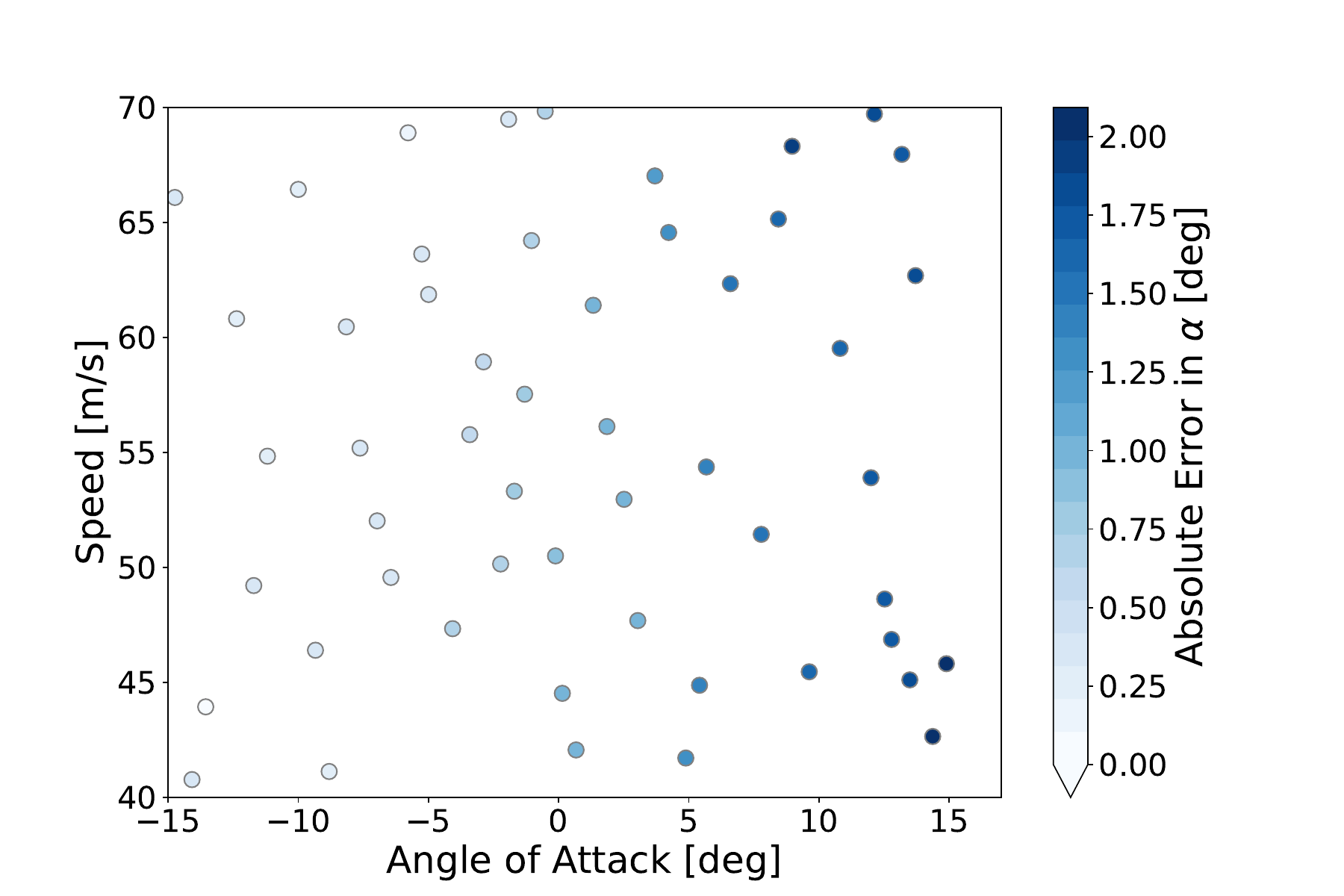} 	
\caption{$N_{OL}$ on $\mathcal{D}_R$ (Conv-D)} \label{fig_ae_aoa_tl2_convD_ot}
\end{subfigure} \hfill \vspace{4mm} \break 
\caption{Test set absolute errors obtained with $N_{OL}$ for transfer learning between the datasets generated using different turbulence models and the task of predicting $\alpha$ ($n_s=75$, $n_d=1024$)}
\label{fig_ae_aoa_tl2_o}
\end{figure}

\begin{figure}[!t] \centering
\begin{subfigure}[b]{0.48\textwidth}
\includegraphics[width=.99\columnwidth]{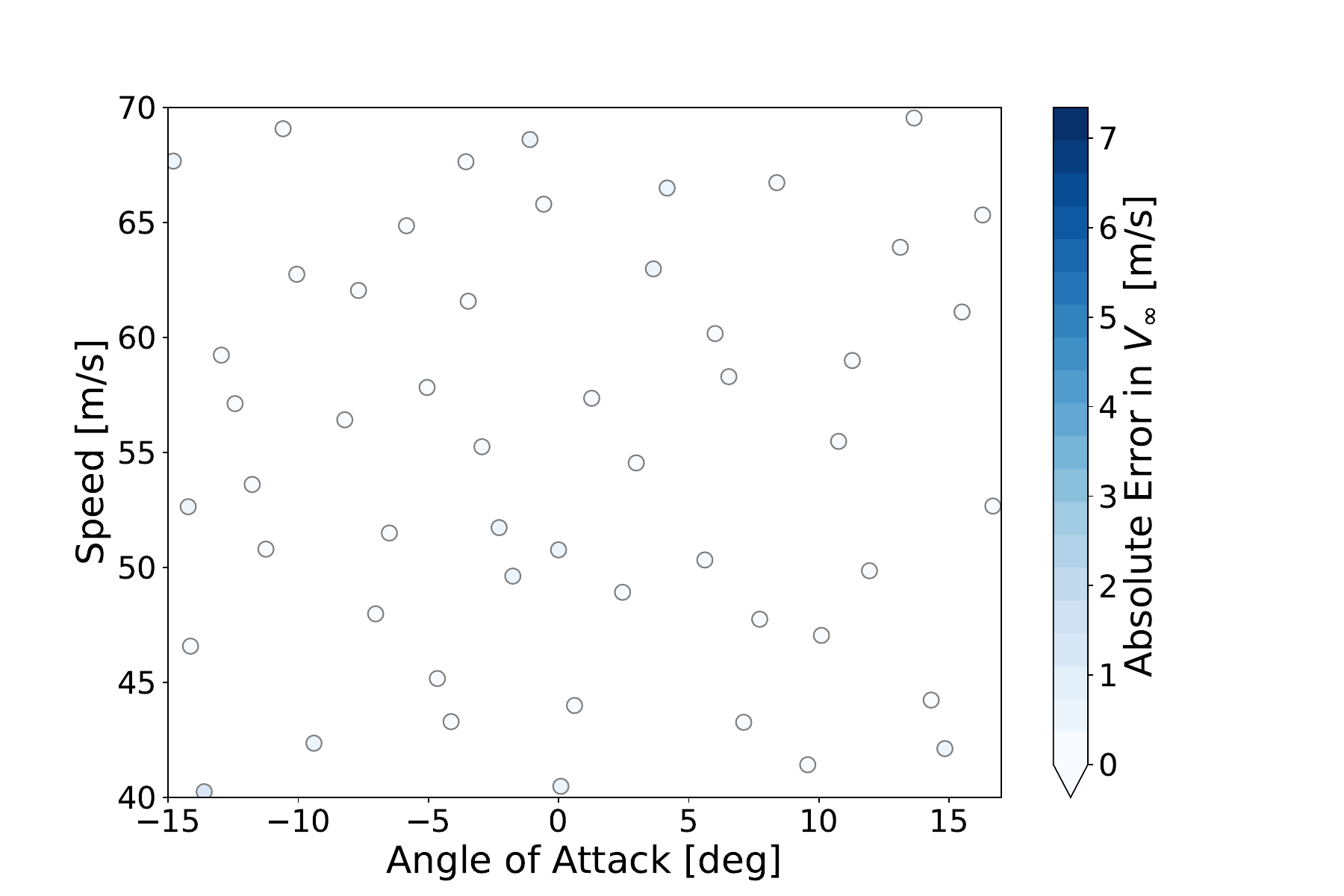} 	
\caption{$N_{OL}$ on $\mathcal{D}_S$ (Conv-S)} \label{fig_mae_spd_tl2_convS_oo}
\end{subfigure} \hfill
\begin{subfigure}[b]{0.48\textwidth}
\includegraphics[width=.99\columnwidth]{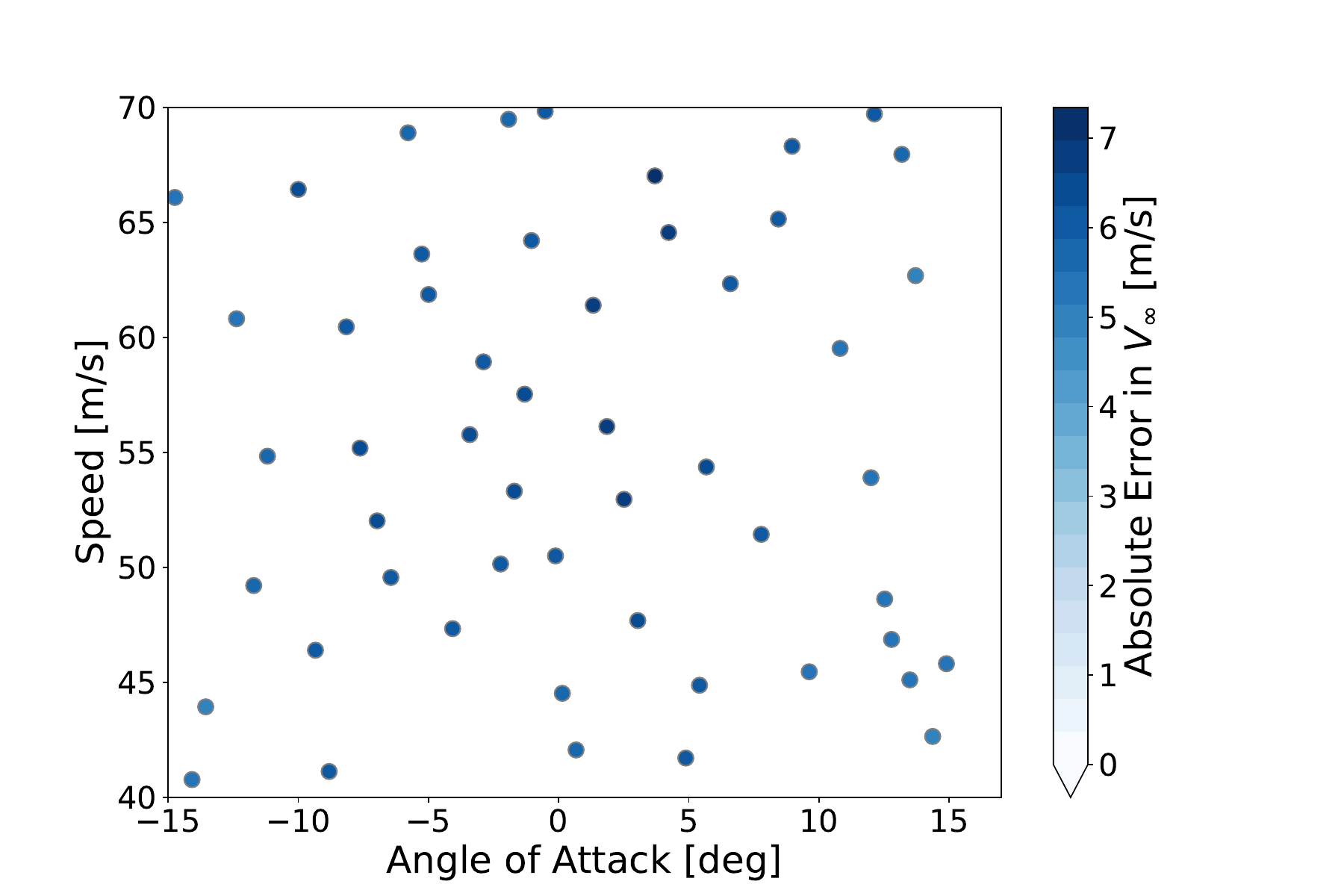} 	
\caption{$N_{OL}$ on $\mathcal{D}_R$ (Conv-S)} \label{fig_mae_spd_tl2_convS_ot}
\end{subfigure} \hfill \vspace{4mm} \break 
\begin{subfigure}[b]{0.48\textwidth}
\includegraphics[width=.99\columnwidth]{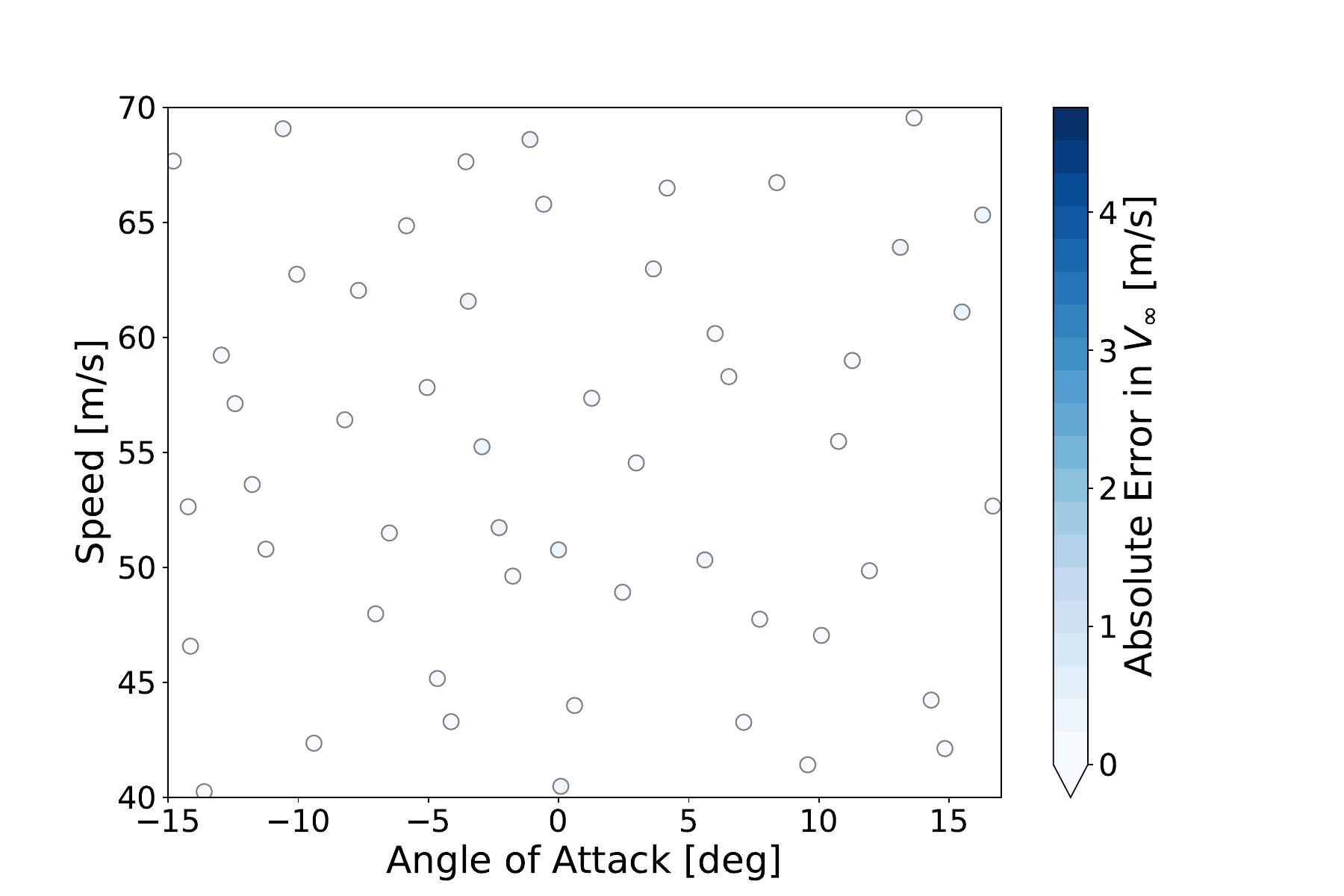} 	
\caption{$N_{OL}$ on $\mathcal{D}_S$ (Conv-D)} \label{fig_ae_spd_tl2_convD_oo}
\end{subfigure} \hfill
\begin{subfigure}[b]{0.48\textwidth}
\includegraphics[width=.99\columnwidth]{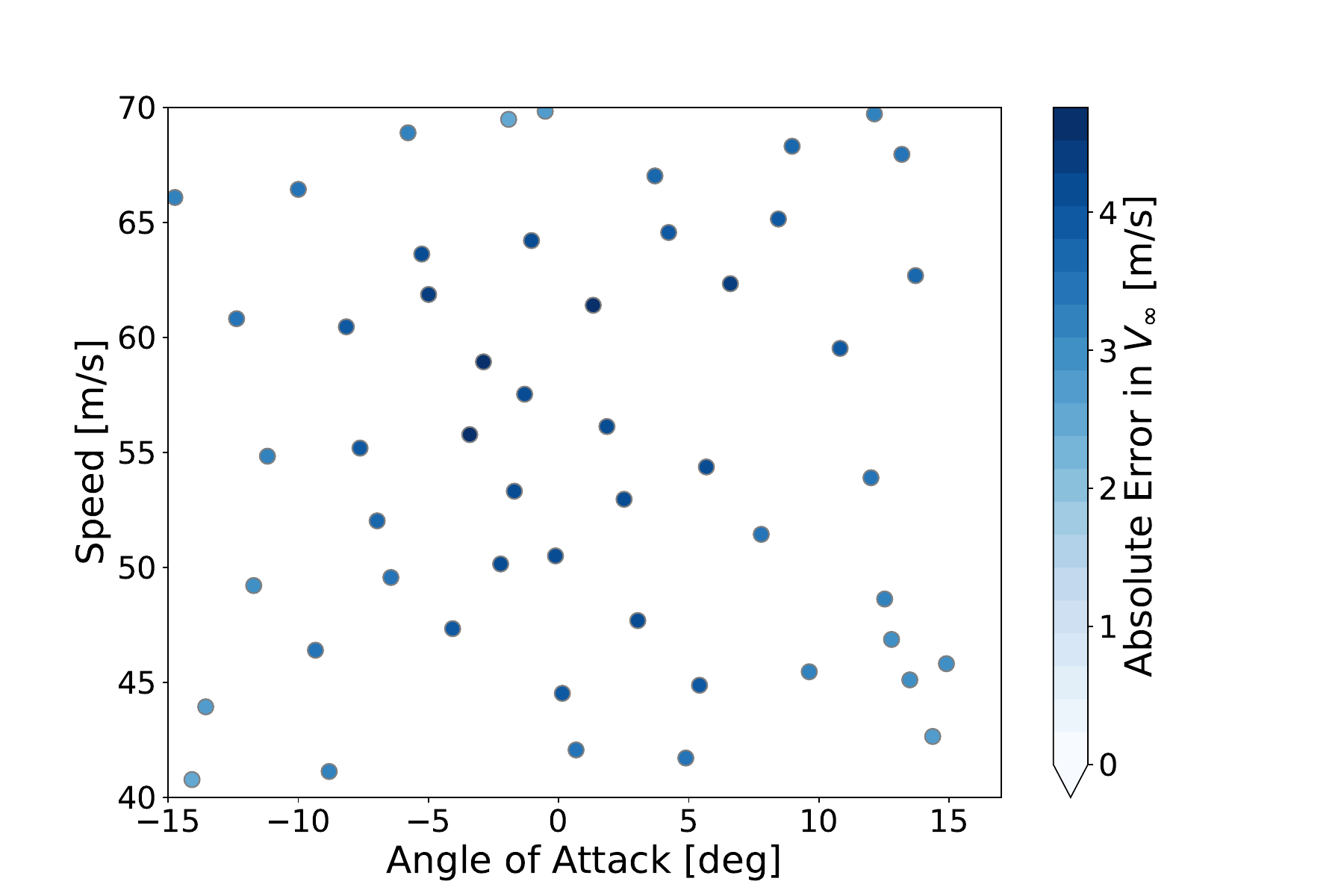} 	
\caption{$N_{OL}$ on $\mathcal{D}_R$ (Conv-D)} \label{fig_ae_spd_tl2_convD_ot}
\end{subfigure} \hfill \vspace{4mm} \break 
\caption{Test set absolute errors obtained with $N_{OL}$ for transfer learning between the datasets generated using different turbulence models and the task of predicting $V_{\infty}$ ($n_s=75$, $n_d=1024$)}
\label{fig_ae_spd_tl2_o}  
\end{figure}

The prediction accuracy results of the individual test set elements obtained using the offline trained neural networks are provided in Figures~\ref{fig_ae_aoa_tl2_o} and \ref{fig_ae_spd_tl2_o} for the case of adaptation to changing data distributions and the tasks of predicting $\alpha$ and $V_{\infty}$, respectively. 

\begin{table}[!h] \centering
\caption{The parameters of the ConvNet-D architecture}
\begin{tabular}{cccccc}
Conv. Layer & size & Conv. Parameters & Linear Layer & size &  Pooling Parameters \\
\cmidrule(l r ){1-1} \cmidrule(l r ){2-2} \cmidrule(l r ){3-3} \cmidrule(l r ){4-4} \cmidrule(l r ){5-5} \cmidrule(l r ){6-6}
Conv1 & (1, 64) &   kernel size:11  & Linear1 & (1536, 4096)  & Max Pooling  \\
Conv2 & (64, 192) & stride:4 &   Linear2 & (4096, 4096)  & kernel size:3 \\
Conv3 & (192, 384) &  padding:2 & Linear3 & (4096, 1) & stride:2 \\
Conv4 & (384, 256) &  & & & \\
Conv4 & (256, 256) &  & & & \\
\end{tabular} \label{table_convnetD}
\end{table}

\begin{table}[!h] \centering 
\caption{The parameters of the ConvNet-S architecture}
\begin{tabular}{cccccc}
Conv. Layer & size & Conv. Parameters &  Linear Layer & size &   Pooling Parameters \\
\cmidrule(l r ){1-1} \cmidrule(l r ){2-2} \cmidrule(l r ){3-3} \cmidrule(l r ){4-4} \cmidrule(l r ){5-5} \cmidrule(l r ){6-6}
Conv1 & (1, 64) & kernel sizes:(11, 5) & Linear1 & (1152, 4096)  & Max Pooling \\
Conv2 & (64, 192) & stride:(4, -) & Linear2 & (4096, 1)  & kernel size:3 \\
&  &  padding:2 & & & stride:2 \\  
\end{tabular}  \label{table_convnetS}
\end{table}

\begin{table}[!ht] \centering 
\caption{The parameters of the FCNN architecture}
\begin{tabular}{cc}
Linear Layer & size\\
\cmidrule(l r ){1-1} \cmidrule(l r ){2-2}  
Linear1 & ($n_{in}$, 4096) \\
Linear2 & (4096, 1) \\ 
\end{tabular}  \label{table_fcnn}
\end{table}

\section*{Acknowledgements}
The authors gratefully acknowledge the scientific support and HPC resources provided by the German Aerospace Center (DLR). The HPC system CARA is partially funded by ``Saxon State Ministry for Economic Affairs, Labor and Transport'' and ``Federal Ministry for Economic Affairs and Climate Action''.

\clearpage
\def\bibsection{\section*{\refname}}
\bibliography{ref}

\end{document}